\documentclass{article}

\usepackage{microtype}
\usepackage{graphicx}
\usepackage{hyperref}
\usepackage{subcaption}

\usepackage[accepted]{latex/icml2024}
\newcommand{\publicifelse}[2]{\ifdefined\isaccepted#1\else#2\fi}

\newcommand*\LET[2]{\STATE #1 $\gets$ #2}
\usepackage{layouts}

\usepackage{mathtools, amssymb}
\usepackage{xcolor,xstring}
\usepackage{float,stfloats}
\usepackage{booktabs,multirow}
\usepackage{graphicx,tikz}
\usepackage{enumitem}
\usepackage[capitalize,noabbrev]{cleveref}

\icmltitlerunning{Faithfulness Measurable Masked Language Models}

\begin{document}

\twocolumn[
\icmltitle{Faithfulness Measurable Masked Language Models}

\begin{icmlauthorlist}
\icmlauthor{Andreas Madsen}{mila,poly}
\icmlauthor{Siva Reddy}{mila,mcgill,facebook_cifar}
\icmlauthor{Sarath Chandar}{mila,poly,canada_cifar}
\end{icmlauthorlist}

\icmlaffiliation{mila}{Mila, Montreal, Canada}
\icmlaffiliation{poly}{Computer Engineering and Software Engineering Department, Polytechnique Montreal, Montreal, Canada}
\icmlaffiliation{mcgill}{Computer Science and Linguistics, McGill University, Montreal, Canada}
\icmlaffiliation{facebook_cifar}{Facebook CIFAR AI Chair}
\icmlaffiliation{canada_cifar}{Canada CIFAR AI Chair}

\icmlcorrespondingauthor{Andreas Madsen}{andreas.madsen@mila.quebec}
\icmlcorrespondingauthor{Siva Reddy}{siva.reddy@mila.quebec}
\icmlcorrespondingauthor{Sarath Chandar}{sarath.chandar@mila.quebec}

\icmlkeywords{Interpretability, Accountability, Transparency, Explainability, Faithfulness, Faithful, Importance Measure, Explanation, Explain, Faithfulness Measurable, Faithfulness Measurable Model, post-hoc, intrinsic, Beam, signed IM, unsigned IM}

\vskip 0.3in
]

\printAffiliationsAndNotice{}

\begin{abstract}

A common approach to explaining NLP models is to use importance measures that express which tokens are important for a prediction. Unfortunately, such explanations are often wrong despite being persuasive. Therefore, it is essential to measure their faithfulness. One such metric is if tokens are truly important, then masking them should result in worse model performance.
However, token masking introduces out-of-distribution issues, and existing solutions that address this are computationally expensive and employ proxy models. Furthermore, other metrics are very limited in scope.
This work proposes an inherently faithfulness measurable model that addresses these challenges. This is achieved using a novel fine-tuning method that incorporates masking, such that masking tokens become in-distribution by design.
This differs from existing approaches, which are completely model-agnostic but are inapplicable in practice.
We demonstrate the generality of our approach by applying it to 16 different datasets and validate it using statistical in-distribution tests. The faithfulness is then measured with 9 different importance measures.
Because masking is in-distribution, importance measures that themselves use masking become consistently more faithful. Additionally, because the model makes faithfulness cheap to measure, we can optimize explanations towards maximal faithfulness; thus, our model becomes indirectly inherently explainable.


\end{abstract}

\section{Introduction}
\label{sec:introduction}

As machine learning models are increasingly being deployed, the demand for interpretability to ensure safe operation increases \citep{Doshi-Velez2017a}. In NLP, importance measures such as attention or integrated gradient are a popular way of explaining which input tokens are important for making a prediction \citep{Bhatt2020}. These explanations are not only used directly to explain models but are also used in other explanations such as contrastive \citep{Kayo2022}, counterfactuals \citep{Ross2020}, and adversarial explanations \citep{Ebrahimi2018}.

Unfortunately, importance measures (IMs) are often found to provide false explanations despite being persuasive \citep{Jain2019, Hooker2019}. For example, a given IMs might not be better at revealing important tokens than pointing at random tokens \citep{Madsen2022}. This presents a risk, as false but persuasive explanations can lead to unsupported confidence in a model. Therefore, it's important to measure faithfulness. \citet{Jacovi2020} defines faithfulness as: ``how accurately it (explanation) reflects the true reasoning process of the model''. In this work, we propose a methodology that enables existing models to support measuring faithfulness by design.

Measuring faithfulness is challenging, as there is generally no known ground-truth for the correct explanation. Instead, faithfulness metrics have to use proxies. One such proxy is the \emph{erasure-metric} by \citet{Samek2017}: if tokens are truly important, then masking them should result in worse model performance compared to masking random tokens.

However, masking tokens can create out-of-distribution issues. This can be solved by retraining the model after allegedly important tokens have been masked \citep{Hooker2019, Madsen2022}. Unfortunately, this is computationally expensive, leaks the gold label, and the measured model is now different from the model of interest.

In general, the cost of proxies has been some combination of incorrect assumptions, expensive computations, or using a proxy-model \citep{Jain2019, Bastings2021, Madsen2022}. Based on previous work, we propose the following desirable, which to the best of our knowledge, no previous faithfulness metric for importance measures satisfies in all aspects, but we satisfy:\\

\begin{figure*}[t!]
    \centering
    \includegraphics[width=\linewidth]{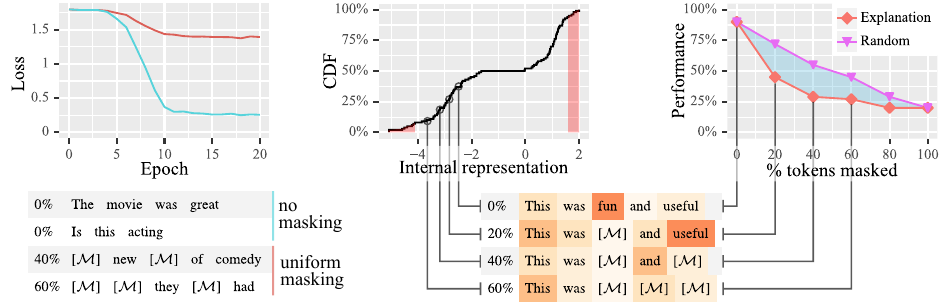}
    \begin{subfigure}[b]{0.31\textwidth}
      \caption{\textbf{Masked fine-tuning.} In-distribution support for masking any permutation of tokens is achieved by uniformly masking half of the mini-batch during fine-tuning. The other half is left unmasked to maintain regular unmasked performance.}
      \label{fig:workflow:masked-fine-tuning}
    \end{subfigure}
    \hfill
    \begin{subfigure}[b]{0.31\textwidth}
      \caption{\textbf{In-distribution validation.} CDFs of the model's embeddings given a masked validation dataset, provide in-distribution p-values and validate that test observations masked according to an explanation are in-distribution.}
      \label{fig:workflow:ood}
    \end{subfigure}
    \hfill
    \begin{subfigure}[b]{0.31\textwidth}
      \caption{\textbf{Faithfulness metric.} Observations are masked according to an explanation. A model performance lower than masking random tokens means the explanation is faithful. A larger area between the curves means more faithful.}
      \label{fig:workflow:metric}
    \end{subfigure}
    \caption{To measure faithfulness, a \emph{faithfulness measurable masked language model} is created (a), then the model is checked for out-of-distribution issues given an explanation (b), and finally, the faithfulness is measured by masking allegedly important tokens (c). -- \emph{[$\mathcal{M}$] is the masking token.}}
    \label{fig:workflow}
\end{figure*}

\begin{enumerate}[label={\alph*)}, itemsep=-1pt, topsep=0pt]
    \item The method does not assume a known true explanation. 
    \item The method measures faithfulness of an explanation w.r.t. a specific model instance and single observation. For example, it is not a proxy-model that is measured.
    \item The method uses only the original dataset, e.g. does not introduce spurious correlations.
    \item The method only uses inputs that are in-distribution w.r.t. the model.
    \item The method is computationally cheap by not training/fine-tuning repeatedly and only computes explanations of the test dataset.
    \item The method can be applied to any classification task.
\end{enumerate}
The key idea is to use the erasure-metric but without solving the out-of-distribution (OOD) issue using retraining, which avoids all the limitations. This is achieved by including masking in the fine-tuning procedure of masked language models, such that masking is always in-distribution (\Cref{fig:workflow:masked-fine-tuning}). This is possible because language models are heavily over-parameterized and can thus support such additional complexity. Although our approach applies to Masked Language Models (MLMs), we suspect future work could apply this idea to any language model with sufficient capacity. For example, \citet{Vafa2021} applied a similar idea to causal language models for top-k explanations.

Our approach is significantly different from previous literature on importance measures, which have taken a completely model-agnostic perspective using for example retraining \citep{Hooker2019,Madsen2022,Amara2023}. Instead, we fine-tune a model such that measuring faithfulness of importance measures is easy by design. We call such designed models: \textbf{inherently faithfulness measurable models} (FFMs). 

To validate that masking is in-distribution, we generalize previous OOD detection work from computer vision \citep{Heller2022}. This serves as a statistically grounded meta-validation of the faithfulness measure itself (\Cref{fig:workflow:ood}), something that previous works have not achieved \citep{Vafa2021,Hase2021,Madsen2022}. Finally, once the model is validated, the erasure-metric \citep{Samek2017} can be applied (\Cref{fig:workflow:metric}).

Note, the concept of an \emph{inherently faithfulness measurable model} (FMM) is significantly different from inherently explainable models, which are interpretable by design \citep{Jacovi2020}. An FMM does not guarantee that an explanation exists, and an inherently explainable model doesn't provide a means of measuring faithfulness.

However, with an FMM, measuring faithfulness is computationally cheap. Therefore, optimizing an IM explanation w.r.t. faithfulness is possible, as proposed by \citet{Zhou2022a}. However, they did not solve the OOD issue caused by masking, as it was ``orthogonal'' to their idea, but our \emph{inherently faithfulness measurable model} fills that gap, making it indirectly inherently explainable.

Finally, for completeness, we compare a large variety of existing importance measure explanations and modify some existing explanations to be able to separate positive from negative contributions. In general, we find that the explanations that take advantage of masking (occlusion-based) are more faithful than gradient-based methods. However, the robustness provided by a \emph{faithfulness measurable model} also makes some gradient-based methods more faithful. 

\paragraph{To summarize, our contributions are:}
\begin{itemize}[noitemsep,topsep=0pt]
    \item Introducing the concept of an \emph{inherently faithfulness measurable model} (FMM).
    \item Proposing \emph{masked fine-tuning} that enables masking to be in-distribution.
    \item Establishing a statistically grounded meta-validation for the faithfulness measurable model, using out-of-distribution detection.
    \item Making existing occlusion-based explanations more faithful, as they no longer cause out-of-distribution issues.
    \item Introducing signed variants of existing importance measures, which can separate between positive and negative contributing tokens.
\end{itemize}

\section{Related Work}
\label{sec:related-work}
Much recent work in NLP has been devoted to investigating the faithfulness of importance measures. In this section, we categorize these faithfulness metrics according to their underlying principle and discuss their limitations. The limitations are annotated as (\textbf{a}) to (\textbf{f}) and refer to the desirables mentioned in the \hyperref[sec:introduction]{Introduction}.

Note, we do not cover the faithfulness metrics that are specific to attention \citep{Moradi2021, Wiegreffe2020, Vashishth2019}, as this paper presents a faithfulness metric for importance measures in general. Additionally, we do not cover other explanations, such as top-k explanations by \citet{Hase2021} and \citet{Vafa2021}.

\subsection{Correlating importance measures}
\label{sec:related-works:correlation}
One early idea was to compare two importance measures. The claim is that a correlation would be a very unlikely coincidence unless both explanations are faithful \citep{Jain2019}. Both \citet{Jain2019} and \citet{Meister2021a} find little correlation between attention, gradient, and leave-one-out; the explanations are therefore not faithful. \citet{Jain2019} do acknowledge the limitations of their approach, as it assumes each importance measure is faithful to begin with (\textbf{a}). 

\subsection{Known explanations in synthetic tasks}
\label{sec:related-works:known}
\citet{Arras2020} construct a purely synthetic task, where the true explanation is known, therefore the correlation can be applied appropriately. Unfortunately, this approach cannot be used on real datasets (\textbf{f}). Instead, \citet{Bastings2021} introduce spurious correlations into real datasets, creating partially synthetic tasks. They then evaluate if importance measures can detect these correlations. It is assumed that if an explanation fails this test, it is generally unfaithful. \citet{Bastings2021} conclude that faithfulness is both model and task-dependent. 

Both methods are valid when measuring faithfulness on models trained on (partially) synthetic data. However, the model and task-dependent conclusion also means that we can't generalize the faithfulness findings to the models (\textbf{b}) and datasets of interest (\textbf{c}), thus limiting the applicability of this approach.

\subsection{Similar inputs, similar explanation}
\label{sec:related-works:similar}
\citet{Jacovi2020} suggest that if similar inputs show similar explanations, then the explanation method is faithful. \citet{Zaman2022} apply this idea using a multilingual dataset (\textbf{f}), where each language example is explained and aligned to the English example using a known alignment mapping. If the correlation between language pairs is high, this indicates faithfulness.

Besides being limited to multilingual datasets, the metric assumes the model behaves similarly among languages. However, languages may have different linguistic properties or spurious correlations. A faithful explanation would then yield different explanations for each language.

\subsection{Removing important information should affect the prediction}
\label{sec:related-works:roar}
\citet{Samek2017} introduce the erasure-metric: if information (input tokens) is truly important, then removing it should result in worse model performance compared to removing random information. However, \citet{Hooker2019} argue that removing tokens introduces an out-of-distribution (OOD) issue (\textbf{d}), and \citep{Amara2023} have later shown this in detail.

\citet{Hooker2019} solve the ODD issue by retraining the model. They point out an important limitation; namely, the method cannot separate an unfaithful explanation from ``there exist dataset redundancies''. However, this was later solved by \citet{Madsen2022}, who conclude, like \citet{Bastings2021}, that faithfulness is both model and task-dependent.

Unfortunately, both methods are computationally expensive (\textbf{e}), due to the model being retrained. Additionally, retraining means that it is no longer possible to comment on the faithfulness of the deployed model (\textbf{b}). Finally, removing tokens and retraining can introduce redundancies which underestimates the faithfulness \citep{Madsen2022}.

\section{Inherently faithfulness measurable models (FMMs)}
\label{sec:faithfulness-measurable-models}

As an alternative to existing faithfulness methods for importance measures, which all aim to work with any models, we propose creating \emph{inherently faithfulness measurable models} (FMMs). These models provide the typical output (e.g., classification) for a given task and, by design, provide the means to measure the faithfulness of an explanation. Importantly, this allows measuring the faithfulness of a specific model, as there is no need for proxy models, an important property in a real deployment setting. This idea is similar to what \citet{Hase2021} and \citet{Vafa2021} proposed for top-k explanations.

An FMM does have the limitation that a specific model is required. However, our proposed method is very general, as it only requires a modified fine-tuning procedure applied to a masked language model.

\subsection{Faithfulness of importance measures}
In this paper, we look at importance measures (IMs), which are explanations that either score or rank how important each input token is for making a prediction. A faithfulness metric measures how much such an explanation reflects the true reasoning process of the model \citep{Jacovi2020}. Importantly, such a metric should work regardless of how the importance measure is calculated.

For importance measures, there are multiple definitions of truth. In this paper, we use the erasure-metric definition: \emph{if information (tokens) is truly important, then masking them should result in worse model performance compared to masking random information (tokens)} \citep{Samek2017, Hooker2019, DeYoung2020}.

The challenge with an erasure-metric is that fine-tuned models do not support masking tokens. Even masked language models are usually only trained with 12\% or 15\% masking \citep{Devlin2019, Liu2019, Wettig2022}, and an erasure-metric use between 0\% and 100\% masking. Furthermore, catastrophic forgetting of the masking token is likely when fine-tuning.

\citet{Hooker2019} and \citet{Madsen2022} solve this by retraining the model with partially masked inputs and call the approach ROAR (\underline{R}em\underline{o}ve {\underline{a}nd  \underline{R}etrain). Unfortunately, as discussed in \Cref{sec:related-work}, retraining has issues. It is computationally expensive, leaks the gold label, and measures a proxy model instead of the true model.

We find that the core issue is the need for retraining. Instead, if the fine-tuned model supports masking any permutation of tokens, then retraining would not be required, eliminating all issues. We propose a new fine-tuning procedure called \emph{masked fine-tuning} to achieve this.

To evaluate faithfulness of an importance measure, we propose a three-step process, as visualized in \Cref{fig:workflow} (also see \Cref{appendix:workflow} for details):
\begin{enumerate}[noitemsep, topsep=0pt]
    \item Create a faithfulness measurable masked language model, using \emph{masked fine-tuning}. -- See \Cref{sec:faithfulness-measurable-models:masked-fine-tuning} and \Cref{fig:workflow:masked-fine-tuning}.
    \item Check for out-of-distribution (OOD) issues, by using a statistical in-distribution test. -- See \Cref{sec:faithfulness-measurable-models:ood} and \Cref{fig:workflow:ood}.
    \item Measure the faithfulness of an explanation. -- See \Cref{sec:faithfulness-measurable-models:metric} and \Cref{fig:workflow:metric}.
\end{enumerate}

\subsection{Masked fine-tuning}
\label{sec:faithfulness-measurable-models:masked-fine-tuning}
To provide masking support in the fine-tuned model, we propose randomly masking the training dataset by uniformly sampling a masking rate between 0\% and 100\% for each observation and then randomly masking that ratio of tokens. However, half of the mini-batch remains unmasked to maintain the regular unmasked performance. This is analogous to multi-task learning, where one task is masking support, and the other is regular performance. 

To include masking support in early stopping, the validation dataset is duplicated, where one copy is unmasked, and one copy is randomly masked.

The high-level idea is similar to \citet{Hase2021} which enabled masking support for a fixed number of tokens. However, to support a variable number of tokens we improve upon this work by sampling a masking-ratio.

\subsection{In-distribution validation}
\label{sec:faithfulness-measurable-models:ood}
Erasure-based metrics are only valid when the input masked according to the importance measure is in-distribution, and previous works did not validate for this \cite{Hooker2019,Madsen2022,Hase2021,Vafa2021}. Additionally, in-distribution is the statistical null-hypothesis and can never be proven. However, we can validate this using an out-of-distribution (OOD) test.

We use the \emph{MaSF} method by \citet{Heller2022} as the OOD test, which provides non-parametric p-values under the in-distribution null-hypothesis. While \emph{MaSF} is developed and tested for computer vision, it is very general and naturally applies to NLP. The method works by developing empirical Cumulative Distribution Functions (CDFs) of the model's intermediate embeddings. In the case of RoBERTa we use the embeddings after the layer-normalization \citep{Ba2016}. 

The validation dataset is used to develop the empirical CDFs. Because the validation dataset should be from the same distribution as the training dataset, the masked fine-tuning transformation is also used on the validation dataset. Once the CDFs are developed, a test observation can be tested against the CDFs which provides in-distribution p-values for each embedding, these are then aggregated using Fisher \citep{Fisher1992} and Simes's \citep{Simes1986} method. Finally, to a p-value for the entire masked test dataset being in-distribution we perform another Simes \citep{Simes1986} aggregation. The details of the entire workflow and procedure are described in \Cref{appendix:workflow:masf}.

\subsection{Faithfulness metric}
\label{sec:faithfulness-measurable-models:metric}

To measure faithfulness on a model trained using \emph{masked fine-tuning}: the importance measure (IM) is computed for a given input, then $x\%$ (e.g., $10\%$) of the most important tokens are masked, then the IM is calculated on this masked input, finally an additional $x\%$ of the most important tokens are masked. This is repeated until $100\%$  of the input is masked. The importance measure is re-calculated because otherwise, dataset redundancies will interfere with the metric, as shown by \citet{Madsen2022}.

At each iteration, the masked input is validated using MaSF and the performance is measured. Faithfulness is shown if and only if the performance is less than when masking random tokens. This procedure is identical to Recursive-ROAR by \citet{Madsen2022}, but without retraining and with in-distribution validation.

\section{Importance measures (IMs)}
\label{sec:importance-measures}

Our proposed \emph{inherently faithfulness measurable model} and the erasure-metric can be applied to a single observation, which is useful in practical settings but not for statistical conclusions. Therefore, we focus on importance measures, which are feasible to compute on the entire test dataset.

The importance measures used in this paper are all from existing literature; hence, the details are in the \Cref{appendix:importance-measures}, except we introduce signed and absolute variants of existing IMs. Additionally, we argue that occlusion-based IMs are only valid in combination with \emph{masked fine-tuning}.

\subsection{Gradient-based vs occlusion-based}
\label{sec:importance-measures:grad-vs-occlusion}
A common idea is that if a small change in the input causes a large change in the output, then that indicates importance.

\paragraph{Gradient-based} The relationship of change can be modeled by the gradient w.r.t. the input. Let $f(\mathbf{x})$ be the model with input\footnote{$V$ is vocabulary-size and $T$ is sequence-length.} $\mathbf{x} \in \mathbb{R}^{T \times V}$, then the gradient explanation (\textbf{grad}) is $\nabla_\mathbf{x} f(\mathbf{x})_y \in \mathbb{R}^{T \times V}$ \citep{Baehrens2010, Li2016}. Later work \citep{Kindermans2016} have proposed using $\textbf{x} \odot \nabla_\mathbf{x} f(\mathbf{x})_y$ instead (i.e. \textbf{x $\odot$ grad}), or to sample multiple gradients (\textbf{IG}) \citep{Sundararajan2017a}. 

\paragraph{Occlusion-based} An alternative to using gradients is to measure the change when removing or masking a token and measuring how it affects the output. One rarely addressed concern, is that removing or masking tokens is likely to cause out-of-distribution issues \citep{Zhou2022a}. This would cause an otherwise sound method to become unfaithful. However, because our proposal \emph{faithfulness measurable masked language model} supports masking, this should not be a concern. Importantly, this exemplifies how a \emph{inherently faithfulness measurable model} may also produce more faithful explanations.

A simple occlusion-based explanation is leave-one-out (\textbf{LOO}) which tests the effect of each token, one-by-one \citep{Li2016a}. However, this does not account for when one token is masked another token may become more important. Instead, \citet{Zhou2022a} propose to optimize for the faithfulness metric itself, rather than using a heuristic. The idea is to use beam-search (\textbf{Beam}), where the generated sequence is the optimal masking order of tokens.

\subsection{Signed and absolute variants}
Most literature does not distinguish between positive and negative contributing tokens\footnote{Unfortunately, the details of this are often omitted in the literature and can often only be learned through reading the code. However, \citet{Meister2021a} do provide the details.}, which we categorize as an \emph{absolute IM}. In the opposite case, when positive and negative are separated, we categorize them as a \emph{signed IM}. In most cases, a signed IM can be transformed into an absolute IM using $abs(\cdot)$.

\section{Experiments}

We use RoBERTa in size \texttt{base} and \texttt{large}, with the default GLUE hyperparameters provided by \citet{Liu2019}. We present results on 16 classification datasets in the appendix but only include BoolQ and MRPC in the main paper. These were chosen as they represent the general trends we observe, although we observe very consistent results across all datasets. The full dataset list is in \Cref{appendix:datasets} and model details are in \Cref{appendix:models}. The code is available at \publicifelse{\url{https://github.com/AndreasMadsen/faithfulness-measurable-models}}{\url{[REDACTED]}}.

For each experiment, we use 5 seeds and present their means with their 95\% confidence interval (error-bars or ribbons). The 95\% confidence interval is computed using the bias-corrected and accelerated bootstrap method \citep{Buckland1998,Michael2011}. When relevant, each seed is presented as a plus ({\small\texttt{+}}).

\subsection{Masked fine-tuning}
\label{sec:experiment:fine-tune}

There are two criteria for learning our proposed \emph{faithfulness measurable masked language model}:
\begin{enumerate}[itemsep=-1pt, topsep=0pt]
    \item The usual performance metric, where no data is masked, should not decrease.
    \item Masking any permutations of tokens should be in-distribution.
\end{enumerate}

In \Cref{sec:faithfulness-measurable-models:masked-fine-tuning}, we propose \emph{masked fine-tuning}, where one half of a mini-batch is uniformly masked between 0\% and 100\% and the other half is unmasked. Additionally, the validation dataset contains a masked copy and an unmasked copy.

\paragraph{Unmasked performance.} To validate the first goal, \Cref{fig:paper:unmasked-performance} presents an ablation study. It compares \emph{masked fine-tuning} with using only unmasked data (\emph{plain fine-tuning}), as is traditionally done, and using only uniformly masked data (\emph{only masking}). The unmasked performance is then measured (the usual benchmark).

\begin{figure}[t]
    \centering
    \includegraphics[trim=0pt 7pt 0pt 7pt, clip, width=\linewidth]{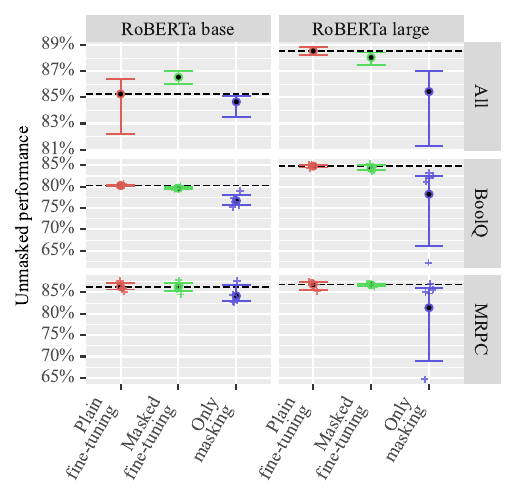}
    \caption{The unmasked performance for each fine-tuning strategy. \emph{Plain fine-tuning} is the baseline (dashed line). We find that our \emph{Masked fine-tuning} does not decrease performance. \emph{All} is computed by taking the average of all datasets. More datasets and a more detailed ablation study can be found in \Cref{appendix:masked-fine-tuning}.}
    \label{fig:paper:unmasked-performance}
\end{figure}

We observe that no performance is loss when using our \emph{masked fine-tuning}, some tasks even perform better likely because masking have a regualizing effect. However, when using \emph{only masking} performance is lost unstable convergence is frequent. For bAbI-2\&3, we also observe unstable convergence using \emph{masked fine-tuning}. However, this is less frequent (worst case: 3/5) and only for RoBERTA-large (see \Cref{appendix:masked-fine-tuning}). Note the default RoBERTa hyperparameters are not meant for synthetic datasets like bAbI. Therefore, optimizing the hyperparameter would likely solve the stability issues with \emph{masked fine-tuning}. Finally, when using \emph{masked fine-tuning}, the models do need to be trained for slightly more epochs (twice more or less); see \Cref{appendix:epochs}. Again, tuning hyperparameter would likely help.

\paragraph{100\% Masked performance.} Measuring in-distribution support for masked data is challenging, as there is generally no known performance baseline. However, for 100\% masked data, only the sequence length is left as information. Therefore, the performance of a model should be at least that of the class-majority baseline, where the most frequent class is ``predicted'' for all observations. We present an ablation study using this baseline in \Cref{fig:paper:masked-100p-performance}. In \Cref{sec:experiment:ood}, we perform a more in-depth validation.

\begin{figure}[t]
    \centering
    \includegraphics[trim=0pt 7pt 0pt 7pt, clip,width=\linewidth]{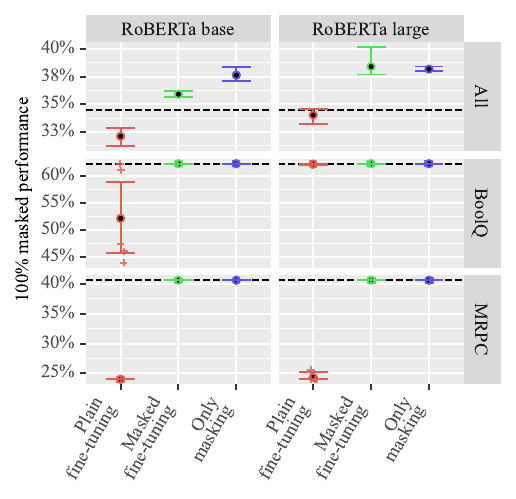}
    \caption{The 100\% masked performance for each fine-tuning strategy. The dashed line represents the class-majority baseline. Results show that masking during training (either our \emph{masked fine-tuning} or \emph{only masking}) is necessary. More datasets and a more detailed ablation study can be found in \Cref{appendix:masked-fine-tuning}.}
    \label{fig:paper:masked-100p-performance}
\end{figure}

From \Cref{fig:paper:masked-100p-performance}, we observe that training with unmasked data (\emph{Plain fine-tuning}) performs worse than the class-majority baseline, clearly showing an out-of-distribution issue. However, when using masked data, either \emph{only masking} or \emph{masked fine-tuning}, both effectively achieve in-distribution results for 100\% masked data.

\paragraph{The best approach used in the following experiments.} \Cref{appendix:masked-fine-tuning} contains a more detailed ablation study separation of the training and validation strategy. However, the conclusion is the same. \emph{Masked fine-tuning} is the only method that achieves good results for both the unmasked and 100\% masked cases. 

For the following experiments in \Cref{sec:experiment:ood} and \Cref{sec:experiment:faithfulness}, the \emph{masked fine-tuning} method is used. Additionally, we will only present results for RoBERTa-base for brevity. RoBERTa-large results are included in the appendix.

\subsection{In-distribution validation}
\label{sec:experiment:ood}

Because the expected performance for masked data is generally unknown, a statistical in-distribution test called \emph{MaSF} \citep{Heller2022} is used instead, as was briefly explained in \Cref{sec:faithfulness-measurable-models:ood}, with details in \Cref{appendix:workflow:masf}.

MaSF provides an in-distribution p-value for each observation. To test if all masked test observations are in-distribution, the p-values are aggregated using Simes's method \citep{Simes1986}. Because in-distribution is the null-hypothesis, we can never confirm in-distribution; we can only validate it. Rejecting the null hypothesis would mean that some observation is out-of-distribution.

\begin{figure}[t!]
    \centering
    \includegraphics[trim=7pt 7pt 6pt 7pt, clip, width=\linewidth]{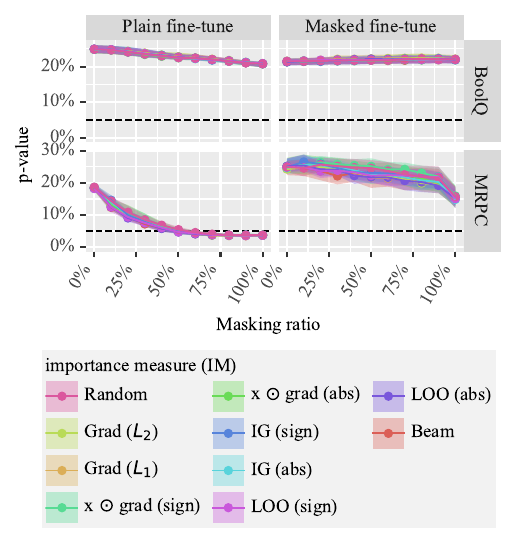}
    \caption{In-distribution p-values using MaSF, for RoBERTa-base with and without masked fine-tuning. The masked tokens are chosen according to an importance measure. P-values below the dashed line show out-of-distribution (OOD) results, given a 5\% risk of a false positive. Results show that only when using \emph{masked fine-tuning} is masking consistently not OOD. Because the results are highly consistent, the overlapping lines do not hide any important details. More datasets and models in \Cref{sec:appendix:ood}. }
    \label{fig:paper:ood}
\end{figure}

Because random uniform masking is not the same as strategically masking tokens, we validate in-distribution for each importance measure, where the masking is done according to the importance measure, identically to how the faithfulness metric is computed (\Cref{sec:faithfulness-measurable-models:metric}).

Additionally, because MaSF does not consider the model's performance, it is necessary to consider these results in combination with regular performance metrics, see \Cref{sec:experiment:fine-tune}.

The results for when using \emph{masked fine-tuning} and \emph{plain fine-tuning} (no masking) are presented in \Cref{fig:paper:ood}. The results show that masked datasets are consistently in-distribution only when using masked fine-tuning.

In the case of BoolQ, we suspect that because the training dataset is fairly small (7542 observations), the model does not completely forget the mask token. Additionally, a few datasets, such as bAbI-2, become out-of-distribution at 100\% masking when using masked fine-tuning (see \Cref{sec:appendix:ood}). This contradicts the performance results for 100\% masked data (\Cref{appendix:masked-fine-tuning}), which clearly show in-distribution performance. This is likely a limitation of MaSF because the empirical CDF in MaSF has very little 100\% masked data, as the masking ratio is uniform between 0\% and 100\%. Fortunately, this is not a concern because \Cref{fig:paper:masked-100p-performance} shows in-distribution results for 100\% masked data.

\begin{figure}[t!]
    \centering
    \includegraphics[trim=0pt 7pt 0pt 7pt, clip, width=\linewidth]{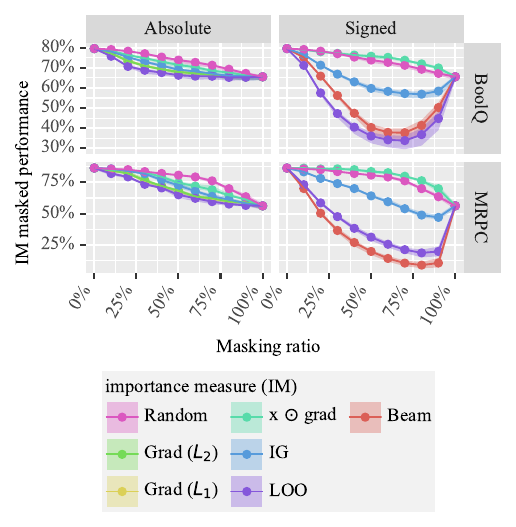}
    \caption{The performance given the masked datasets, where masking is done for the x\% allegedly most important tokens according to the importance measure. If the performance for a given explanation is below the \emph{``Random''} baseline, this shows faithfulness. Although faithfulness is not an absolute concept, so more is better. This plot is for RoBERTa-base and separates importance measures based on their signed and absolute variants. More datasets and models in \Cref{appendix:faithfulness}.}
    \label{fig:paper:faithfulness}
\end{figure}

\subsection{Faithfulness metric}
\label{sec:experiment:faithfulness}

Based on previous experiments, we can conclude that \emph{masked fine-tuning} achieves both objectives: unaffected regular performance and support for masked inputs. Therefore, it is safe to apply the faithfulness metric to these models.

Briefly, the faithfulness metric works by showing that masking using an importance measure (IM) is more effective at removing important tokens than using a known false explanation, such as a random explanation. Therefore, if a curve is below the random baseline, the IM is faithful. Although faithfulness is not an absolute \citep{Jacovi2020}, so further below would indicate more faithful. 

\begin{table}[t!]
    \centering
    \caption{Faithfulness scores using Relative Area Between Curves (RACU) and the non-relative variant (ACU). The less relevant score is grayed out. Higher is better. Negative values indicate not-faithful. The comparison with Recursive-ROAR \citep{Madsen2022} is imperfect because Recursive-ROAR has limitations. See \Cref{tab:appendix:faithfulness:roberta-sb} for all datasets and \Cref{tab:appendix:faithfulness:roberta-sl} for RoBERTa-large. See \Cref{appendix:additional-discussions} for additional result discussion.}
    \resizebox{\linewidth}{!}{\begin{tabular}[t]{llccc}
\toprule
& & \multicolumn{3}{c}{Faithfulness [\%]}  \\
\cmidrule(r){3-5}
Dataset & IM & \multicolumn{2}{c}{Our} & R-ROAR \\
\cmidrule(r){3-4}
& & ACU & RACU & RACU \\
\midrule
\multirow[c]{9}{*}{SST2} & Grad ($L_2$) & {\color{black!30}$12.2_{-0.7}^{+0.6}$} & $40.4_{-1.7}^{+3.0}$ & $26.1_{-2.2}^{+1.6}$ \\
 & Grad ($L_1$) & {\color{black!30}$12.1_{-0.7}^{+0.7}$} & $40.3_{-1.8}^{+3.3}$ & -- \\
 & x $\odot$ grad (sign) & $-3.7_{-1.6}^{+1.5}$ & {\color{black!30}$-12.2_{-6.0}^{+4.5}$} & {\color{black!30}--} \\
 & x $\odot$ grad (abs) & {\color{black!30}$7.1_{-0.2}^{+0.2}$} & $23.5_{-1.1}^{+1.9}$ & $18.6_{-4.6}^{+4.1}$ \\
 & IG (sign) & $31.8_{-2.2}^{+2.8}$ & {\color{black!30}$105.6_{-7.7}^{+7.7}$} & {\color{black!30}--} \\
 & IG (abs) & {\color{black!30}$13.7_{-0.8}^{+0.8}$} & $45.3_{-2.8}^{+4.1}$ & $32.9_{-1.5}^{+1.8}$ \\
 & LOO (sign) & $51.6_{-0.9}^{+1.4}$ & {\color{black!30}$171.3_{-6.2}^{+5.8}$} & {\color{black!30}--} \\
 & LOO (abs) & {\color{black!30}$16.6_{-1.0}^{+1.2}$} & $54.9_{-1.5}^{+2.1}$ & -- \\
 & Beam & $56.4_{-0.7}^{+0.5}$ & {\color{black!30}$187.3_{-7.1}^{+8.1}$} & {\color{black!30}--} \\
\cmidrule{1-5}
\multirow[c]{9}{*}{bAbI-2} & Grad ($L_2$) & {\color{black!30}$28.5_{-0.8}^{+0.8}$} & $96.3_{-2.8}^{+6.8}$ & $57.8_{-2.0}^{+2.0}$ \\
 & Grad ($L_1$) & {\color{black!30}$28.5_{-0.8}^{+0.9}$} & $96.3_{-2.7}^{+6.8}$ & -- \\
 & x $\odot$ grad (sign) & $19.7_{-8.1}^{+6.6}$ & {\color{black!30}$65.7_{-26.3}^{+24.1}$} & {\color{black!30}--} \\
 & x $\odot$ grad (abs) & {\color{black!30}$27.3_{-1.5}^{+1.7}$} & $92.0_{-3.1}^{+2.5}$ & $48.1_{-3.5}^{+3.2}$ \\
 & IG (sign) & $40.3_{-0.8}^{+0.9}$ & {\color{black!30}$136.3_{-6.4}^{+4.4}$} & {\color{black!30}--} \\
 & IG (abs) & {\color{black!30}$29.1_{-1.3}^{+1.0}$} & $98.3_{-3.9}^{+5.5}$ & $42.0_{-4.8}^{+3.8}$ \\
 & LOO (sign) & $40.2_{-0.8}^{+1.2}$ & {\color{black!30}$136.0_{-6.5}^{+4.1}$} & {\color{black!30}--} \\
 & LOO (abs) & {\color{black!30}$28.5_{-1.4}^{+0.9}$} & $96.3_{-3.6}^{+9.2}$ & -- \\
 & Beam & $41.1_{-0.7}^{+1.0}$ & {\color{black!30}$139.2_{-7.3}^{+5.0}$} & {\color{black!30}--} \\
\bottomrule
\end{tabular}
}
    \label{tab:paper:faithfulness}
\end{table}

\Cref{fig:paper:faithfulness} shows that most explanations are faithful, although some are much more faithful than others. In particular, occlusion-based importance measures (LOO, Beam) are the most faithful. This is expected, as they take advantage of the masking support that our \emph{faithfulness measurable masked language model} offers.

Because signed importance measures can differentiate between positive and negative contributing tokens, while absolute tokens are not, it is to be expected that signed importance measures are more faithful. However, comparing them might not be fair because of this difference in capability. We let the reader decide this for themselves.

\paragraph{Relative Area Between Curves (RACU)} \citet{Madsen2022} propose to compute the area between the random curve and an explanation curve (RACU). This is then normalized by the theoretical optimal explanation, which would achieve the performance of 100\% masking immediately. However, the normalization is only theoretically optimal for an absolute importance measure (IM). Signed IMs can trick the model into predicting the opposite label, thus achieving even lower performance. For this reason, we also show the un-normalized metric (ACU) in \Cref{tab:paper:faithfulness}. 

Note that comparing with Recursive ROAR (R-ROAR) \citep{Madsen2022} is troublesome because R-ROAR has issues, such as leaking the gold label. Additionally, while they also use RoBERTa-base it's not the same model because we use masked fine-tuning.

That said, our Faithfulness Measurable Masked Language Model drastically outperforms the R-ROAR approach on faithfulness. In \Cref{appendix:additional-discussions}, we provide additional discussion on some less important observations.

\section{Limitations}
\label{sec:limitations}

In this section, we discuss the most important limitations. Additionally, in the interest of completeness, \Cref{appendix:all-limitations} provides additional limitations.

\paragraph{No faithfulness ablation with regular fine-tuning} We claim \emph{masked fine-tuning} makes importance measures (IMs) more faithful. However, there is no ablation study where we measure faithfulness without \emph{masked fine-tuning}. This is because, without \emph{masked fine-tuning}, masking is out-of-distribution which makes the faithfulness measure invalid.

However, our argument for occlusion-based IMs has a theoretical foundation, as occlusion (i.e., masking) is only in-distribution because of \emph{masked fine-tuning}. We also observe that occlusion-based IMs are consistently more faithful than gradient-based IMs. Finally, for gradient-based IMs, we compare with Recursive ROAR \citep{Madsen2022}, and our approach provides more faithful explanations, although this comparison is imperfect as discussed in \Cref{sec:experiment:faithfulness}.

\paragraph{Uses masked language models (MLMs)}
Masked fine-tuning leverages pre-trained MLMs' partial support for token masking. Therefore, our approach does not immediately generalize to casual language models (CLM). However, despite CLMs' popularity for generative tasks, MLMs are still very relevant for classification tasks \citep{Min2024} and for non-NLP tasks, such as analyzing biological sequences (genomes, proteins, etc.) \citep{Zhang2023a}.

Additionally, it is possible to introduce the mask tokens to CLMs by masking random tokens in the input sequence while keeping the generation objective the same, similar to how unknown-word tokens are used. This approach could also be done in an additional pre-training step using existing pre-trained models.  Regardless, masking support for CLMs is likely a more complex task and is left for future work.

Another direction useful for classification tasks, is to transform CLMs into MLMs, which has been shown to be quite straightforward \citep{Muennighoff2024}. It may also be possible to simply prompt an instruction-tuned CLM, such that it understands what masking means, for example \citet{Madsen2024} prompts with ``The following content may contain redacted information marked with [REDACTED]''.

In terms of supporting sequential outputs rather than just classification outputs, our methodology only requires a performance metric. Using sequential performance metrics such as ROUGE \citep{Lin2004} or BLEU should therefore work perfectly well.

\section{Conclusion}

Using only a simple modified fine-tuning method, called \emph{masked fine-tuning}, we are able to turn a typical general-purpose masked language model (RoBERTa) into an \emph{inherently faithfulness measurable model} (FMM). Meaning that the model, by design, inherently provides a way to measure the faithfulness of importance measure (IM) explanations.

To the best of our knowledge, this is the first work that proposes creating a model designed to be faithfulness measurable. Arguably, previous work in top-k explanations \citep{Hase2021,Vafa2021} and counterfactual explanation \citep{Wu2021,Kaushik2020} have indirectly achieved something similar. However, their motivation was to provide explanations or robustness, not measuring faithfulness.

Importantly, our approach is very general, simple to apply, and stratifies critical desirables that previous faithfulness measures didn't. The \emph{masked fine-tuning} method does not decrease performance on all 16 tested datasets while also adding in-distribution support for token masking, which we are able to verify down to fundamental statistics using an out-of-distribution test.

We find that occlusion-based IMs are consistently the most faithful. This is to be expected, as they take advantage of the masking support. Additionally, Beam uses beam-search to optimize towards faithfulness \citep{Zhou2022a}, which our proposed faithfulness measurable masked language model makes computationally efficient to evaluate.

It is worth considering the significance of this. While our proposed model is not an \emph{inherently explainable model} \citep{Jacovi2020}, it is \emph{indirectly} inherently explainable because it provides a built-in way to measure faithfulness, which can then be optimized for. It does this without sacrificing the generality of the model, as it is still a RoBERTa model. As such, FMMs provide a new direction for interpretability, which bridges the gap between \emph{post-hoc} \citep{Madsen2021} and \emph{inherent} interpretability \citep{Rudin2019}. It does so by prioritizing faithfulness measures first and then the explanation, while previous directions have worked on explanation first and then measure faithfulness \citep{Madsen2024a}.

However, beam-search is just an approximative optimizer, which only archives perfect explanations at infinite beam-width, and Leave-one-out does occasionally outperform Beam for that reason. Future work could look at better optimization methods to improve the faithfulness of importance measures.

%
%

\section*{Impact Statement}
Interpretability is an essential component when considering the ethical deployment of a model. In this paper, we particularly consider deployment a key motivator in our work, as we specify that for a \emph{faithfulness measurable model} the deployed model and measured model should be the same. Without this criterion, there is a dangerous potential for discrepancies between the analyzed model and the deployed model, as has been the case in previous work.

However, there is still a risk that the faithfulness measure can be wrong. This is problematic, as a false faithfulness measure would create unsupported confidence in the importance measure. A key assumption with our proposed metric is that the model provides in-distribution support for any permutation of masked tokens. We take extra care to validate this assumption using sound methodologies. Note that our work is not the only one with this assumption. In particular, ROAR-based metrics \citep{Hooker2019, Madsen2022} also assume in-distribution behavior but did not test for this, while we do test for it. Additionally, we attempt to provide complete transparency regarding the limitations of our work in \Cref{sec:limitations} and \Cref{appendix:all-limitations}.

\subsection*{Use of anonymized medical data}
The Diabetes and Anemia datasets used in this work are based on the anonymized open-access medical database MIMIC-III \citep{Johnson2016}. These datasets are used as they are common in the faithfulness literature \citep{Jain2019, Madsen2022}. In particular, they contain many redundancies and long-sequence inputs. The specific authors of this paper who have worked with this data have undergone the necessary HIPAA certification in order to access this data and comply with HIPAA regulations.

\ifdefined\isaccepted
\section*{Acknowledgements}
Sarath Chandar is supported by the Canada CIFAR AI Chairs program, the Canada Research Chair in Lifelong Machine Learning, and the NSERC Discovery Grant.

Siva Reddy is supported by the Facebook CIFAR AI Chairs program and NSERC Discovery Grant.

Computing resources were provided by the Digital Research Alliance of Canada.
\fi

\bibliography{references}
\bibliographystyle{latex/icml2024}

\newpage
\appendix
\section{Additional discussion}
\label{appendix:additional-discussions}
In this section, we discuss additional observations that can be made from the main paper results. These observations are valuable but are not necessary for the main message in the paper.

\Cref{tab:paper:faithfulness} show that the RACU scores have a lower variance (confidence interval) using our methodology compared to Recursive ROAR. This is likely because Recursive ROAR leaks the gold label \citep{Madsen2022}, which causes oscillation in the faithfulness curve.

We also observe that gradient-based explanations are more faithful when using our model. We suspect this is partially also because there is no leakage issue. However, previous work has also shown that gradient-based methods behave more favorably on robust models in computer vision \citep{Bansal2020}. Using masked fine-tuning can be seen as a robustness objective, as the model becomes robust to missing information.

Surprisingly, input times gradient (x $\odot$ grad) appears to be the worst explanation. This is a curious result because both \emph{gradient} (Grad) and \emph{integrated gradient} (IG) perform much better. Recall that \emph{integrated gradient} is essentially $x \odot \int \text{grad}$ \citep{Sundararajan2017a}. Perhaps a version of \emph{integrated gradient} where the x-hadamard product (x $\odot$) is not used might improve the faithfulness.

Finally, Diabetes and bAbI-2 get 90\% RACU faithfulness (using absolute importance measures). This suggests that there are only a few words that are critical for the prediction of these tasks. This may indicate a problem with their efficacy as benchmarking datasets. However, for Diabetes, which is purely a diagnostic dataset \citep{Jain2019, Johnson2016}, this may be fine if those tokens should be critical to the task.

\section{Additional limitations}
\label{appendix:all-limitations}

\subsection{Not a post-hoc method}
While this work solves existing limitations with previous methods, it introduces the significant limitation that it, by definition, requires a \emph{faithfulness measurable model}. As such, the question of faithfulness needs to be considered ahead of time when developing a model. It can not be an afterthought, which is often how interpretability is approached \citep{Bhatt2020, Madsen2021}. 

While this is a significant limitation, considering explanation ahead of deployment is increasingly becoming a legal requirement \citep{Doshi-Velez2017}. Currently, the European Union provides a ``right to explanation'' regarding automatic decisions, which includes NLP models \citep{Goodman2017}.

\subsection{In-distribution is impossible to prove}
Because in-distribution is always the null hypothesis, it is impossible to statistically show that inputs are truly in-distribution. The typical approach to similar statistical questions\footnote{A similar well-explored statistical question is how to show that the error in a linear model is normally distributed.} is to keep validating in-distribution using various methods. Unfortunately, the literature on this topic in deep learning is extremely limited \citep{Yang2022,Sun2020,Heller2022,Dziedzic2022,Kwon2020}.

Therefore, we would advocate for more work on identifying out-of-distribution inputs using non-parametric methods that primarily consider the model's internal state. Using parametric methods or works that use axillary models is more well-explored but not useful for our purpose.

\subsection{Requires repeated measures on the test dataset}
Because datasets have redundancies, it is necessary to reevaluate the importance measures \citep{Hooker2019, Madsen2022}. This leads to an increased computational cost.

However, unlike previous work \citep{Madsen2022}, our method only requires reevaluation of the test dataset, which is often quite small. Additionally, some IMs, such as the beam-search method \citep{Zhou2022a}, take dataset redundancies into account and therefore do not require reevaluation. Reevaluation could be done if desired but would result in the exact same results.

\subsection{Measures only faithfulness}
There are other important aspects to an importance measure, such as if the explanation is useful to humans \citep{Doshi-Velez2017a}, but our method does not measure this. We consider this a separate topic, as it is an HCI question that requires experimental studies with humans. The topic of faithfulness can not be measured by humans \citep{Jacovi2020}, as neural networks are too complicated for humans to manually evaluate if an explanation is true.

\section{Datasets}
\label{appendix:datasets}

The datasets used in this work are all public and listed below. They are all used for their intended use, which is measuring classification performance. The Diabetes and Anemia datasets are from MIMIC-III which requires a HIPPA certification for data analysis \citep{Johnson2016}. The first author complies with this and has not shared the data with any, including other authors.

Note that when computing the importance measure on paired-sequence tasks, only the first sequence is considered. This is to stay consistent with previous work \citep{Jain2019, Madsen2022} and because for tasks like document-based Q\&A (e.g., bAbI), it does not make sense to mask the question.

The essential statistics for each dataset, and which part is masked and auxiliary, are specified in \Cref{tab:appendix:datasets}.

\begin{table*}[bt!]
    \centering
    \caption{Datasets used, all datasets are either single-sequence or sequence-pair datasets. All datasets are sourced from GLUE \citep{Wang2019}, SuperGLUE \citep{Wang2019c}, MIMIC-III \citep{Johnson2016}, or bAbI \citep{Weston2015}. The decisions regarding which metrics are used are also from these sources. The class-majority baseline is when the most frequent class is always selected.\vspace{0.1in}}
    \resizebox{\linewidth}{!}{{\defcitealias{Dolan2005}{Dolan et al., 2005} 
\begin{tabular}{llcccccccc}
\toprule
Type & Dataset & \multicolumn{3}{c}{Size} & \multicolumn{2}{c}{Inputs} & \multicolumn{2}{c}{Performance} & Citation \\
\cmidrule(r){3-5} \cmidrule(r){6-7} \cmidrule(r){8-9}
& & Train & Validation & Test & masked & auxilary & metric & class-majority \\
\midrule\
\multirow{5}{*}{NLI}        & RTE       & $1992$ & $498$ & $277$ & \texttt{sentence1} & \texttt{sentence2} & Accuracy & $47\%$ & {\footnotesize\citealp{Dagan2006}} \\
                            & SNLI      & $549367$ & $9842$ & $9824$ & \texttt{premise} & \texttt{hypothesis} & Macro F1 & $34\%$ & {\footnotesize\citealp{Bowman2015}} \\
                            & MNLI      & $314162$ & $78540$ & $9815$ & \texttt{premise} & \texttt{hypothesis} & Accuracy & $35\%$ & {\footnotesize\citealp{Williams2018}} \\
                            & QNLI      & $83794$ & $20949$ & $5463$ & \texttt{sentence} & \texttt{question} & Accuracy & $51\%$ & {\footnotesize\citealp{Rajpurkar2016}} \\
                            & CB        & $200$ & $50$ & $56$ & \texttt{premise} & \texttt{hypothesis} & Macro F1 & $22\%$ & {\footnotesize\citealp{Marneffe2019}} \\ \cmidrule{2-10}
\multirow{2}{*}{Paraphrase} & MRPC      & $2934$ & $734$ & $408$ & \texttt{sentence1} & \texttt{sentence2} & Macro F1 & $41\%$ & {\footnotesize\citetalias{Dolan2005}} \\
                            & QQP       & $291077$ & $72769$ & $40430$ & \texttt{question1} & \texttt{question2} & Macro F1 & $39\%$ & {\footnotesize\citealp{Iyer2017}} \\ \cmidrule{2-10}
\multirow{2}{*}{Sentiment}  & SST2      & $53879$ & $13470$ & $872$ & \texttt{sentence} & -- & Accuracy & $51\%$ & {\footnotesize\citealp{Socher2013}} \\
                            & IMDB      & $20000$ & $5000$ & $25000$ & \texttt{text} & -- & Macro F1 & $33\%$ & {\footnotesize\citealp{Maas2011}} \\ \cmidrule{2-10}
\multirow{2}{*}{Diagnosis}  & Anemia    & $4262$ & $729$ & $1243$ & \texttt{text} & -- & Marco F1 & $39\%$ & {\footnotesize\citealp{Johnson2016}} \\ 
                            & Diabetese & $8066$ & $1573$ & $1729$ & \texttt{text} & -- & Marco F1 & $45\%$ & {\footnotesize\citealp{Johnson2016}} \\ \cmidrule{2-10}
Acceptability               & CoLA      & $6841$ & $1710$ & $1043$ & \texttt{sentence} & -- & Matthew & $0\%$ &  {\footnotesize\citealp{Warstadt2019}} \\ \cmidrule{2-10}
\multirow{4}{*}{QA}         & BoolQ     & $7542$ & $1885$ & $3270$ & \texttt{passage} & \texttt{question} & Accuracy & $62\%$ & {\footnotesize\citealp{Clark2019}} \\
                            & bAbI-1    & $8000$ & $2000$ & $1000$ & \texttt{paragraph} & \texttt{question} & Micro F1 & $15\%$ & {\footnotesize\citealp{Weston2015}} \\
                            & bAbI-2    & $8000$ & $2000$ & $1000$ & \texttt{paragraph} & \texttt{question} & Micro F1 & $19\%$ & {\footnotesize\citealp{Weston2015}} \\
                            & bAbI-3    & $8000$ & $2000$ & $1000$ & \texttt{paragraph} & \texttt{question} & Micro F1& $18\%$ & {\footnotesize\citealp{Weston2015}} \\               
\bottomrule
\end{tabular}}}
    \label{tab:appendix:datasets}
\end{table*}

\section{Models}
\label{appendix:models}

In this paper, we use the RoBERTa model \citep{Liu2019}, although any masked language model of similar size or larger is likely to work. We choose RoBERTa model, because converges consistently and reasonable hyperparameters are well established. This should make reproducing the results in this paper easier. We use both the \texttt{base} (125M parameters) and \texttt{large} size (355M parameters). 

The hyperparameters are defined by  \citet[Appendix C, GLUE]{Liu2019}. Although these hyperparameters are for the GLUE tasks, we use them for all tasks. The one exception, is that the maximum number of epoch is higher. This is because when \emph{masked fine-tuning} require more epochs. In  \Cref{tab:appendix:models} we specify the max epoch parameter. However, when using early stopping with the validation dataset, the optimization is not sensitive to the specific number of epochs, lower numbers are only used to reduce the compute time.

\begin{table}[H]
    \centering
    \caption{Dataset statistics. Performance metrics are the mean with a 95\% confidence interval.\vspace{0.1in}}
    \small
    \begin{tabular}{lccc}
\toprule
Dataset & max epoch & \multicolumn{2}{c}{Performance} \\
\cmidrule(r){3-4}
& & RoBERTa- & RoBERTa- \\
& & base & large \\
\midrule
BoolQ & 15 & ${80\%}^{+0.2}_{-0.2}$ & ${85\%}^{+0.3}_{-0.3}$ \\
CB & 50 & ${65\%}^{+17.6}_{-47.9}$ & ${87\%}^{+3.1}_{-8.2}$ \\
CoLA & 15 & ${59\%}^{+1.3}_{-1.1}$ & ${66\%}^{+0.8}_{-0.8}$ \\
IMDB & 10 & ${95\%}^{+0.2}_{-0.2}$ & ${96\%}^{+0.2}_{-0.4}$ \\
Anemia & 20 & ${84\%}^{+0.8}_{-0.7}$ & ${84\%}^{+0.5}_{-0.8}$ \\
Diabetes & 20 & ${76\%}^{+0.9}_{-0.9}$ & ${77\%}^{+0.6}_{-1.6}$ \\
MNLI & 10 & ${87\%}^{+0.4}_{-0.2}$ & ${90\%}^{+0.3}_{-0.2}$ \\
MRPC & 20 & ${86\%}^{+0.8}_{-0.7}$ & ${87\%}^{+0.5}_{-1.2}$ \\
QNLI & 20 & ${92\%}^{+0.1}_{-0.1}$ & ${94\%}^{+0.1}_{-0.2}$ \\
QQP & 10 & ${90\%}^{+0.1}_{-0.1}$ & ${91\%}^{+0.0}_{-0.1}$ \\
RTE & 30 & ${75\%}^{+1.4}_{-2.9}$ & ${83\%}^{+1.3}_{-1.4}$ \\
SNLI & 10 & ${91\%}^{+0.1}_{-0.2}$ & ${92\%}^{+0.1}_{-0.2}$ \\
SST2 & 10 & ${94\%}^{+0.2}_{-0.2}$ & ${96\%}^{+0.2}_{-0.2}$ \\
bAbI-1 & 20 & ${100\%}^{+0.0}_{-0.1}$ & ${100\%}^{+0.0}_{-0.0}$ \\
bAbI-2 & 20 & ${99\%}^{+0.1}_{-0.1}$ & ${100\%}^{+0.1}_{-0.1}$ \\
bAbI-3 & 20 & ${90\%}^{+0.2}_{-0.3}$ & ${90\%}^{+0.5}_{-0.5}$ \\
\bottomrule
\end{tabular}

    \label{tab:appendix:models}
\end{table}

\section{Importance measure details}
\label{appendix:importance-measures}

This section provides additional details on the importance measures, which were only briefly described in \Cref{sec:importance-measures}. In particular, not all importance measures have both signed and absolute variants (\Cref{tab:appendix:im-overview}), this section should clarify why.

\begin{table}[ht]
    \centering
    \caption{Overview of the importance measures used in this paper. Note that not all importance measures exist in both signed and absolute variants due to their mathematical construction.\vspace{0.1in}}
    \resizebox{\linewidth}{!}{\begin{tabular}{llll}
\toprule
Category                   & Full name             & Short name   & Variants \\ \midrule
\multirow{3}{*}{Gradient}  & Gradient w.r.t. input & Grad         & absolute \\ 
                           & Input times gradient  & x $\odot$ grad & both \\ 
                           & Integrated gradient   & IG           & both \\  \cmidrule{2-4}
\multirow{2}{*}{Occlusion} & Leave-on-out          & LOO          & both \\
                           & Beam-search           & Beam         & signed \\
\bottomrule
\end{tabular}}
    \label{tab:appendix:im-overview}
\end{table}

\subsection{Gradient-based}
The common idea is that if a small change in the input causes a large change in the output, then that indicates importance.

\paragraph{Gradient (Grad)} measures the mentioned relationship using the gradient, which is a linear approximation. Let $f(\mathbf{x})$ be the model with input $\mathbf{x} \in \mathbb{R}^{T \times V}$ ($V$ is vocabulary-size and $T$ is sequence-length), then the gradient is $\nabla_\mathbf{x} f(\mathbf{x})_y \in \mathbb{R}^{T \times V}$ \citep{Baehrens2010, Li2016}. Because the desire is an importance measure for each token (i.e., $\mathbb{R}^{T}$), the vocabulary dimension is typically\footnote{The topic of vocabulary-dimension reduction is rarely discussed in papers.} reduced using either $L_1$ or $L_2$ norm. In this paper, we consider both.

\paragraph{Input times Gradient (x $\odot$ grad)} multiplies the gradient with the input, which some argue to be better \citep{Kindermans2016} i.e., $\mathbf{x} \odot \nabla_\mathbf{x} f(\mathbf{x}) \in \mathbb{R}^{T \times V}$. Because $\mathbf{x}$ is a one-hot-encoding, using any norm function is just the absolute value of the non-zero element, which is what is typically used. However, we observe that for NLP, one could also consider the signed version of this, where the non-zero element is simply picked out. In this paper, we consider both the signed and absolute variants.

\paragraph{Integrated Gradient (IG)} by \citet{Sundararajan2017a} is a very popular explanation method. This can be seen as an extension of the  \emph{input times gradient} method, and therefore we also consider a signed and absolute variant for this. The method works by sampling gradients between a baseline $f(\mathbf{b})$ and the input $f(\mathbf{x})$. We use a zero-vector baseline and 20 samples, as is commonly done in NLP literature \citep{Mudrakarta2018}

\begin{equation}
\begin{aligned}    
\operatorname{IG}(\mathbf{x}) &= (\mathbf{x} - \mathbf{b}) \odot \frac{1}{k} \sum_{i=1}^{k} \nabla_{\tilde{\mathbf{x}}_i} f(\tilde{\mathbf{x}}_i)_c \\
\tilde{\mathbf{x}}_i &= \mathbf{b} + \frac{i}{k}(\mathbf{x} - \mathbf{b}).
\end{aligned}
\end{equation}

\subsection{Occlusion-based}
\label{sec:importance-measures:occlusion}

Rather than linear-approximating the relation between input and output using gradients, the relationship can also be approximated by removing or masking each token, one by one, and measuring how it affects the output.

One rarely addressed concern is that removing/masking tokens is likely to be out-of-distribution \citep{Zhou2022a}. This would cause an otherwise sound method to become unfaithful. However, because the proposed \emph{faithfulness measurable masked language model} supports masking, this should not be a concern. Importantly, this exemplifies how a \emph{faithfulness measurable model} may also produce more faithful explanations.

\paragraph{Leave-on-out (LOO)} directly computes the difference between the model output difference between the unmasked input and the input with the $i$'th token masked, i.e. the importance is $\{ f(\mathbf{\hat{x}})_y - f(\mathbf{\tilde{x}}_i)_y \}_{i=1}^T \in \mathbb{R}^{T}$ \citep{Li2016a}. Even though this importance measure does not have a vocabulary dimension, it is common to take its absolute \citep{Meister2021a}. However, we consider both absolute and signed variants.

\paragraph{Optimizing for faithfulness (Beam).} \citet{Zhou2022a} propose that we can optimize for the faithfulness metric itself rather than using heuristics. The central idea is to use beam-search, where the generated sequence is the optimal masking order of tokens. Each iteration of the beam-search masks one additional token, where the token is selected by testing every possibility and maximizing the faithfulness metric. This could be reframed as a recursive version of leave-on-out. They propose several optimization targets, but since our faithfulness metric is analog to comprehensiveness, we use this variation.

The number of forward passes is $\mathcal{O}(B \cdot T^2)$; it is not exactly $B \cdot T^2$ because many of them become redundant. This is quite computationally costly, although one advantage is that this explanation is inherently recursive, hence it is not necessary to reevaluate the importance measure in each iteration of the faithfulness metric. However, for long sequence datasets, such as IMDB, BaBi-3, Anemia, and Diabetes, it is not feasible to apply this explanation. In our experiments, we use a beam-size of $B = 10$.

\section{Workflow algorithms}
\label{appendix:workflow}

This section provides the workflow details and algorithms, which were mentioned in \Cref{sec:faithfulness-measurable-models}.

\subsection{Masked fine-tuning}
\label{appendix:workflow:masked-fine-tuning}

Masked fine-tuning is a multi-task learning method, where one task is the typical unmasked performance and the other task is masking support. We achieve this by uniformly masking half of a mini-batch, at between 0\% and 100\% masking, and do not modify the other half. Note that this is slightly different from some multi-task learning methods, which may sample randomly between the two tasks, where we split deterministically. Other methods may also switch between the two tasks in each step, we don't do this as it can create unstable oscillations. Instead, both tasks are included in the same mini-batch.

There are many approaches to implementing masked fine-tuning with identical results, however \Cref{alg:masked-fine-tune} presents our implementation.

\begin{algorithm}[H]
  \caption{Creates the mini-batches used in masked fine-tuning.}
  \label{alg:masked-fine-tune}
  \begin{algorithmic}
\REQUIRE $B$ is a mini-batch with $N$ randomly sampled observations from the training dataset. [$\mathcal{M}$] is the masking token.

\FUNCTION{MiniBatch($B$)}
  \LET{$M$}{$\varnothing$} \COMMENT{Stores new mini-batch}
  \FOR{$i \gets 1$ {\bfseries to} $N$}
    \IF{$i$ is even}
      \LET{$r$}{$\mathrm{SampleUniform}(0, 1)$}
      \LET{$M_i$}{$\mathrm{MaskTokens}(r, B_i)$} \COMMENT{Masks $r\%$ randomly selected tokens in $B_i$.}
    \ELSE
      \LET{$M_i$}{$B_i$}
    \ENDIF
  \ENDFOR
  \STATE {\bfseries return} $M$
\ENDFUNCTION
\STATE
\FUNCTION{MaskTokens($x$, $r$)}
  \LET{$\tilde{x}$}{x}
  \FOR{$t \gets 1$ {\bfseries to} $T$}
    \LET{$s$}{$\mathrm{SampleUniform}(0, 1)$}
    \IF{$s < r$}
      \LET{$x_t$}{[$\mathcal{M}$]}
      \COMMENT{Masks token $t$.}
    \ENDIF
  \ENDFOR
  \STATE {\bfseries return} $\tilde{x}$
\ENDFUNCTION
\end{algorithmic}

\end{algorithm}

\subsection{In-distribution validation (MaSF)}
\label{appendix:workflow:masf}

\emph{MaSF} is a statistical in-distribution test developed by \citet{Heller2022}. \emph{MaSF} is an acronym that stands for Max-Simes-Fisher, which is the order of aggregation functions it uses. \citet{Heller2022} presented some other combinations and orders of aggregation functions but found this to be the best. However, many combinations had similar performance in their benchmark, so we do not consider this particular choice to be important and assume it generalizes well to our case. In \Cref{sec:experiment:ood}, we verified this with an ablation study.

At its core, \emph{MaSF} is an aggregation of many in-distribution p-values, where each p-value is from an in-distribution test of a latent embedding. That is, given a history of embedding observations, which presents a distribution, what is the probability of observing the new embedding or something more extreme? For example, if that probability is less than 5\%, it could be classified as out-of-distribution at a 5\% risk of a false-positive.

However, a model has many internal embeddings, and thus, there will be many p-values. If each were tested independently, there would be many false positives. This is known as p-hacking. To prevent this, the p-values are aggregated using the Simes and Fisher methods, which are aggregation methods for p-values that prevent this issue. Once the p-values are aggregated, it becomes a ``global null-test''. This means the aggregated statistical test checks if any of the embeddings are out-of-distribution.

\paragraph{Emperical CDF.} Each p-value is computed using an empirical communicative density function (CDF). A nice property of an empirical CDF is that it doesn't assume any distribution, a property called non-paramatic. It is however still a model, only if an infinite amount of data was available would it represent the true distribution.

A CDF measures the probability of observing $z$ or less than $z$, i.e., $\mathbb{P}(Z \le z)$. The empirical version simply counts how many embeddings were historically less than the tested embedding, as shown in \eqref{eq:emperical-cdf}. However, as we are also interested in cases where the embedding is abnormally large, hence we also use $\mathbb{P}(Z > z) = 1 - \mathbb{P}(Z \le z)$. We are then interested in the most unlikely case, which is known as the two-sided p-value, i.e. $\min(\mathbb{P}(Z \le z), 1 - \mathbb{P}(Z \le z))$.
\begin{equation}
    \mathbb{P}(Z \le z) = \frac{1}{|Z_\mathrm{emp}|} \sum_{i=1}^{|Z_\mathrm{emp}|} 1[Z_{\mathrm{emp},i} < z]
    \label{eq:emperical-cdf}
\end{equation}

The historical embeddings are collected by running the model on the validation dataset. Note that for this to be accurate, the validation dataset should be i.i.d. with the training dataset. This can easily be accomplished by randomly splitting the datasets, which is common practice, and transforming the validation dataset the same way as the training dataset.

\paragraph{Algorithm.} In the case of \emph{MaSF}, the embeddings are first aggregated along the sequence dimension using the max operation. \citet{Heller2022} only applied \emph{MaSF} to computer vision, in which case it was the width and height dimensions. However, we generalize this to NLP by swapping width and height with the sequence dimension.

The max-aggregated embeddings from the validation dataset provide the historical data for the empirical CDFs. If a network has $L$ layers, each with $H$ latent dimensions, there will be $L \cdot H$ CDFs. The same max-aggregated embeddings are then transformed into p-values using those CDFs. Next, the p-values are aggregated using Simes's method \citep{Simes1986} along the latent dimension, which provides another set of CDFs and p-values, one for each layer ($L$ CDFs and p-values). Finally, those p-values are aggregated using Fisher's method \citep{Fisher1992}, providing one CDF and one p-value for each observation.

The algorithm for \emph{MaSF} can be found in \Cref{alg:masf}. While this algorithm does work, a practical implementation is in our experience non-trivial, as for an entire test dataset ($\mathcal{D}_T$) are $\mathcal{O}(|\mathcal{D}_T| \cdot H \cdot L)$ CDFs evaluations, each involving $\mathcal{O}(\mathcal{D}_V)$ comparisons. While this is computationally trivial on a GPU, it can require a lot of memory usage when done in parallel. Therefore, we found that a practical implementation must batch over both the test and validation datasets.

\begin{algorithm}[H]
  \caption{MaSF algorithm, which provides p-values under the in-distribution null-hypothesis.}
  \label{alg:masf}
  \begin{algorithmic}
\REQUIRE $x$ is the input. $f_e(x) \in \mathbb{R}^{T \times H  \times L}$ provides the model embeddings, where $T$ is sequence-length, $H$ is the hidden-size, and $L$ is the number of layers. $\mathbb{P}$ are the empirical CDFs; these are collected by running the MaSF algorithm on a validation dataset.

\FUNCTION{MaSF($x$, $\mathbb{P}$)}
  \LET{$e$}{$f_e(x)$} \COMMENT{Get embeddings}
  \FOR{$l \gets 1$ {\bfseries to} $L$}
      \FOR{$h \gets 1$ {\bfseries to} $H$}
        \LET{$z_{l,h}^{(1)}$}{$\max_{t=1}^T e_{l,h,t}$}  \COMMENT{Ma-step}
        \LET{$\tilde{p}_{l,h}^{(1)}$}{$\mathbb{P}^{(1)}_{l,h}(Z < z_{l,h}^{(1)})$}
        \LET{$p_{l,h}^{(1)}$}{$\min(\tilde{p}_{l,h}^{(1)}, 1 - \tilde{p}_{l,h}^{(1)})$}
      \ENDFOR
      \LET{$z_{l}^{(2)}$}{$\mathrm{Simes}(p_{l,:}^{(1)})$} \COMMENT{S-step}
      \LET{$\tilde{p}_{l}^{(2)}$}{$\mathbb{P}^{(2)}_{l}(Z < z_{l}^{(2)})$}
      \LET{$p_{l}^{(2)}$}{$\min(\tilde{p}_{l}^{(2)}, 1 - \tilde{p}_{l}^{(2)})$}
  \ENDFOR
  \LET{$z^{(3)}$}{$\mathrm{Fisher}(p^{(2)})$} \COMMENT{F-step}
  \LET{$\tilde{p}^{(3)}$}{$\mathbb{P}^{(3)}(Z < z^{(3)})$}
  \LET{$p^{(3)}$}{$1 - \tilde{p}^{(3)}$}
  \STATE {\bfseries return} $p^{(3)}$
\ENDFUNCTION
\STATE
\FUNCTION{Simes($p$)}
    \LET{$q$}{$\mathrm{SortAscending}(p)$}
    \STATE {\bfseries return} $\min_{i=1}^N q_i \frac{N}{i}$
\ENDFUNCTION
\STATE
\FUNCTION{Fisher($p$)}
    \STATE {\bfseries return} $-2 \sum_{i=1}^N \log(p_i)$
\ENDFUNCTION
\end{algorithmic}

\end{algorithm}

\subsection{Faithfulness metric}
\label{appendix:workflow:faithfulness-metric}

Faithfulness is measured by masking $10\%$ the tokens according to an importance measure, such that the most important tokens are masked. This new 10\%-masked observation is then used as a new input, and the model prediction is explained. This is then repeated until 100\% masking. This process is identical to the Recursive ROAR approach by \citep{Madsen2022}, except without re-training. For an explanation, we define the evaluation procedure in \Cref{alg:faithfulness-metric}. However, in terms of \Cref{fig:faithfulness-metric-visualization}, \Cref{alg:faithfulness-metric} only provides the ``importance measure'' curve. To get the random baseline, simply apply \Cref{alg:faithfulness-metric} again with a random explanation.

\begin{algorithm}[H]
  \caption{Measures the masked model performance given an explanation.}
  \label{alg:faithfulness-metric}
  \begin{algorithmic}
\REQUIRE $\mathrm{IM}(f, x, y) \in \mathbb{R}^T$ explains the model $f$ for the input $x$ and label $y$, with sequence-length $T$. $\delta$ is the iterations step-size (e.g. 10\%) and [$\mathcal{M}$] is the masking token.
\FUNCTION{RecursiveEval($\mathrm{IM}$, $f$, $x$, $y$, $\delta$)}
  \LET{$\tilde{x}_0$}{$x$}
  \LET{$p_0$}{$\mathrm{PerformanceMetric}(f(x), y)$}
  \FOR{$i \gets 1$ {\bfseries to} $1/\delta$}
    \LET{$e_i$}{$\mathrm{IM}(f, \tilde{x}_{i - 1}, y)$}
    \LET{$\tilde{x}_i$}{$\mathrm{AddMask}(e_i, \tilde{x}_{i - 1}, \delta)$}  \COMMENT{Mask $\delta \cdot T$ more tokens in $\tilde{x}_{i - 1}$ using scores $e_i$.}
    \LET{$p_i$}{$\mathrm{PerformanceMetric}(f(\tilde{x}_i), y)$}
  \ENDFOR
  \STATE {\bfseries return} $p$
\ENDFUNCTION
\end{algorithmic}

\end{algorithm}

\begin{figure}[H]
    \centering
    \includegraphics[width=0.90\linewidth]{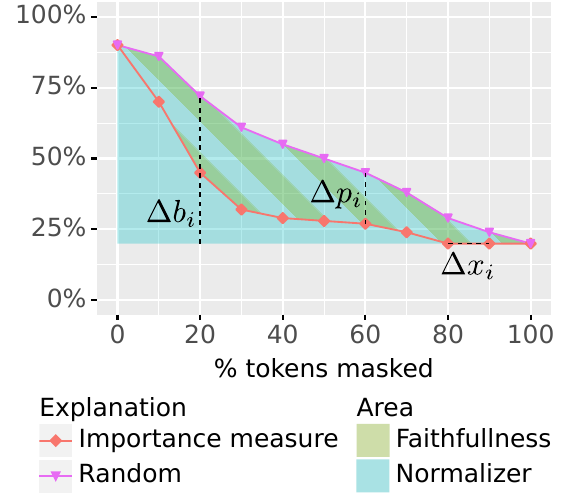}
    \caption{Visualization of the faithfulness calculation done in \eqref{eq:faithfulness-metric}. The \emph{faithfulness} area is the numerator in \eqref{eq:faithfulness-metric}, while the \emph{normalizer} area is the denominator. -- Figure by \citet{Madsen2022}, included with permission.}
    \label{fig:faithfulness-metric-visualization}
\end{figure}

\paragraph{Relative Area between Curves (RACU)}

ACU is the ``faithfulness'' area between the ``importance measure'' curve and the ``random'' curve as shown in \Cref{fig:faithfulness-metric-visualization}. This was not explicitly defined by \citet{Madsen2022}, who defined RACU.

RACU is a normalization of ACU, normalized by the theoretically optimal explanation, which archives the performance of 100\% masking immediately. However, as argued in \Cref{sec:experiment:faithfulness} this normalization is only valid for absolute importance measures. Therefore, we also include the unnormalized score (AUC) in this paper. Both metrics are defined in \Cref{eq:faithfulness-metric}.

Unfortunately, because different models (e.g., RoBERTa-base versus RoBERTa-large) have different performance characteristics, it is not possible to compare unnormalized score (AUC) between different models \citep{Arras2017}. For this reason, RACU may still be useful for signed metrics, even though it can not be interpreted as a 0\%-100\% scale.

\begin{equation}
\begin{aligned}
    \operatorname{ACU} &= \sum_{i=0}^{I-1} \frac{1}{2} \Delta x_i (\Delta p_i + \Delta p_{i+1}) \\
    \operatorname{RACU} &= \frac{
        \operatorname{ACU}
    }{
        \sum_{i=0}^{I-1} \frac{1}{2} \Delta x_i (\Delta b_i + \Delta b_{i+1})
    } \\
    \text{where } \Delta x_i &= x_{i+1} - x_i \quad \textit{step size} \\
    \Delta p_i &= b_i - p_i \quad \textit{performance delta} \\
    \Delta b_i &= b_i - b_I \quad \textit{baseline delta} 
\end{aligned}
\label{eq:faithfulness-metric}
\end{equation}

\section{Compute}

This section reports the compute resources and requirements. The compute hardware specifications are in \Cref{tab:appendix:compute-resources} and were the same for all experiments. All computing was performed using 99\% hydroelectric power.

\begin{table}[h]
    \centering
    \caption{The computing hardware used. Note, that a shared user system were used, only the allocated resources are reported.\vspace{0.1in}}
    \begin{tabular}{lp{5cm}}
        \toprule
        CPU & 12 cores, Intel Silver 4216 Cascade Lake @ 2.1GHz \\
        GPU & 1x NVidia V100 (32G HBM2 memory) \\
        Memory & 24 GB \\
        \bottomrule
    \end{tabular}
    \label{tab:appendix:compute-resources}
\end{table}

Note that the importance measures computed are the same for both the faithfulness results and the out-of-distribution results. Hence, these do not need to be computed twice. Additionally, the beam-search method is in itself recursive, so this was only computed for 0\% masking.

\subsection{Implementation}

We use the HuggingFace implementation of RoBERTa and the TensorFlow framework. The code is available at \publicifelse{\url{https://github.com/AndreasMadsen/faithfulness-measurable-models}}{\url{[REDACTED]}}.

\subsection{Walltimes}

We here include the walltimes for all experiments.
\begin{itemize}[noitemsep,topsep=0pt]
    \item \Cref{tab:appendix:walltime:fine-tine} shows wall-times for the masked fine-tuning.
    \item \Cref{tab:appendix:walltime:odd} shows wall times for the in-distribution validation, not including importance measures.
    \item \Cref{tab:appendix:walltime:faithfulness} shows wall-times for the faithfulness evaluation, not including importance measures.
    \item \Cref{tab:appendix:walltime:im} shows wall-times for the importance measures.
\end{itemize}
\vfill
\begin{table}[H]
    \centering
    \caption{Walltime for fine-tuning. Masked fine-tuning does not affect the total wall time in our setup.\vspace{0.1in}}
    \small
    \begin{tabular}{lcc}
\toprule
Dataset & \multicolumn{2}{c}{Walltime [hh:mm]} \\
\cmidrule(r){2-3}
& RoBERTa- & RoBERTa- \\
& base & large \\
\midrule
BoolQ & 00:51 & 02:03 \\
CB & 00:06 & 00:12 \\
CoLA & 00:17 & 00:33 \\
IMDB & 01:44 & 04:02 \\
Anemia & 00:48 & 02:04 \\
Diabetes & 01:34 & 04:04 \\
MNLI & 06:39 & 14:47 \\
MRPC & 00:12 & 00:27 \\
QNLI & 04:03 & 09:12 \\
QQP & 05:13 & 11:52 \\
RTE & 00:18 & 00:43 \\
SNLI & 04:57 & 10:38 \\
SST2 & 01:19 & 02:44 \\
bAbI-1 & 00:27 & 01:01 \\
bAbI-2 & 00:50 & 02:05 \\
bAbI-3 & 01:43 & 04:28 \\
\midrule
\midrule
sum & 01:43 & 04:28 \\
x5 seeds & 08:37 & 22:21 \\
\bottomrule
\end{tabular}

    \label{tab:appendix:walltime:fine-tine}
\end{table}

\begin{table}[H]
    \centering
    \caption{Walltime for in-distribution validation. This does not include importance measure calculations. See \Cref{tab:appendix:walltime:im}.\vspace{0.1in}}
    \small
    \begin{tabular}[t]{lcc}
\toprule
Dataset & \multicolumn{2}{c}{Walltime [hh:mm]} \\
\cmidrule(r){2-3}
& RoBERTa- & RoBERTa- \\
& base & large \\
\midrule
BoolQ & 00:04 & 00:09 \\
CB & 00:01 & 00:02 \\
CoLA & 00:02 & 00:04 \\
IMDB & 00:15 & 00:44 \\
Anemia & 00:01 & 00:03 \\
Diabetes & 00:01 & 00:04 \\
MNLI & 00:20 & 00:57 \\
MRPC & 00:02 & 00:03 \\
QNLI & 00:05 & 00:12 \\
QQP & 00:47 & 02:13 \\
RTE & 00:02 & 00:04 \\
SNLI & 00:04 & 00:09 \\
SST2 & 00:09 & 00:25 \\
bAbI-1 & 00:01 & 00:03 \\
bAbI-2 & 00:02 & 00:05 \\
bAbI-3 & 00:01 & 00:03 \\
\midrule
\midrule
sum & 02:06 & 05:25 \\
x5 seeds & 10:30 & 27:09 \\
\bottomrule
\end{tabular}

    \label{tab:appendix:walltime:odd}
\end{table}

\begin{table}[H]
    \centering
    \caption{Walltime for faithfulness evaluation. This does not include importance measure calculations. See \Cref{tab:appendix:walltime:im}.\vspace{0.1in}}
    \small
    \begin{tabular}[t]{lcc}
\toprule
Dataset & \multicolumn{2}{c}{Walltime [hh:mm]} \\
\cmidrule(r){2-3}
& RoBERTa- & RoBERTa- \\
& base & large \\
\midrule
BoolQ & 00:02 & 00:05 \\
CB & 00:00 & 00:01 \\
CoLA & 00:01 & 00:02 \\
IMDB & 00:13 & 00:39 \\
Anemia & 00:01 & 00:02 \\
Diabetes & 00:01 & 00:03 \\
MNLI & 00:02 & 00:05 \\
MRPC & 00:01 & 00:02 \\
QNLI & 00:01 & 00:04 \\
QQP & 00:05 & 00:13 \\
RTE & 00:01 & 00:02 \\
SNLI & 00:01 & 00:03 \\
SST2 & 00:01 & 00:01 \\
bAbI-1 & 00:00 & 00:01 \\
bAbI-2 & 00:01 & 00:02 \\
bAbI-3 & 00:00 & 00:02 \\
\midrule
\midrule
sum & 00:38 & 01:34 \\
x5 seeds & 03:13 & 07:51 \\
\bottomrule
\end{tabular}

    \label{tab:appendix:walltime:faithfulness}
\end{table}

\begin{table*}[p]
    \centering
    \caption{Walltime for importance measures. Note that because the beam-search method (Beam) scales quadratic with the sequence-length, it is not feasible to compute for all datasets.\vspace{0.1in}}
    \begin{subtable}[t]{0.275\textwidth}
    \resizebox{0.95\linewidth}{!}{\begin{tabular}[t]{p{1.1cm}ccc}
\toprule
Dataset & IM & \multicolumn{2}{c}{Walltime [hh:mm]} \\
\cmidrule(r){3-4}
& & RoBERTa- & RoBERTa- \\
& & base & large \\
\midrule
\multirow[c]{10}{*}{bAbI-1} & Beam & 00:54 & 02:24 \\
 & Grad ($L_1$) & 00:01 & 00:04 \\
 & Grad ($L_2$) & 00:02 & 00:04 \\
 & x $\odot$ grad (abs) & 00:01 & 00:04 \\
 & x $\odot$ grad (sign) & 00:01 & 00:04 \\
 & IG (abs) & 00:04 & 00:12 \\
 & IG (sign) & 00:04 & 00:11 \\
 & LOO (abs) & 00:24 & 00:49 \\
 & LOO (sign) & 00:24 & 00:49 \\
 & Random & 00:00 & 00:00 \\
\cmidrule{1-4}
\multirow[c]{10}{*}{bAbI-2} & Beam & 20:56 & 61:24 \\
 & Grad ($L_1$) & 00:02 & 00:06 \\
 & Grad ($L_2$) & 00:02 & 00:05 \\
 & x $\odot$ grad (abs) & 00:02 & 00:05 \\
 & x $\odot$ grad (sign) & 00:02 & 00:05 \\
 & IG (abs) & 00:10 & 00:29 \\
 & IG (sign) & 00:10 & 00:29 \\
 & LOO (abs) & 00:39 & 01:26 \\
 & LOO (sign) & 00:39 & 01:26 \\
 & Random & 00:00 & 00:00 \\
\cmidrule{1-4}
\multirow[c]{10}{*}{bAbI-3} & Beam & -- & -- \\
 & Grad ($L_1$) & 00:02 & 00:05 \\
 & Grad ($L_2$) & 00:02 & 00:05 \\
 & x $\odot$ grad (abs) & 00:01 & 00:04 \\
 & x $\odot$ grad (sign) & 00:01 & 00:04 \\
 & IG (abs) & 00:18 & 00:54 \\
 & IG (sign) & 00:18 & 00:53 \\
 & LOO (abs) & 01:09 & 03:18 \\
 & LOO (sign) & 01:09 & 03:18 \\
 & Random & 00:00 & 00:00 \\
\cmidrule{1-4}
\multirow[c]{10}{*}{BoolQ} & Beam & 00:33 & 01:22 \\
 & Grad ($L_1$) & 00:05 & 00:11 \\
 & Grad ($L_2$) & 00:05 & 00:11 \\
 & x $\odot$ grad (abs) & 00:04 & 00:10 \\
 & x $\odot$ grad (sign) & 00:04 & 00:10 \\
 & IG (abs) & 00:39 & 01:48 \\
 & IG (sign) & 00:39 & 01:49 \\
 & LOO (abs) & 00:16 & 00:38 \\
 & LOO (sign) & 00:16 & 00:38 \\
 & Random & 00:00 & 00:00 \\
\cmidrule{1-4}
\multirow[c]{10}{*}{CB} & Beam & 00:45 & 02:09 \\
 & Grad ($L_1$) & 00:01 & 00:03 \\
 & Grad ($L_2$) & 00:01 & 00:03 \\
 & x $\odot$ grad (abs) & 00:01 & 00:03 \\
 & x $\odot$ grad (sign) & 00:01 & 00:03 \\
 & IG (abs) & 00:01 & 00:04 \\
 & IG (sign) & 00:01 & 00:04 \\
 & LOO (abs) & 00:09 & 00:19 \\
 & LOO (sign) & 00:09 & 00:19 \\
 & Random & 00:00 & 00:00 \\
\cmidrule{1-4}
\multirow[c]{10}{*}{CoLA} & Beam & 00:11 & 00:19 \\
 & Grad ($L_1$) & 00:02 & 00:05 \\
 & Grad ($L_2$) & 00:02 & 00:04 \\
 & x $\odot$ grad (abs) & 00:02 & 00:04 \\
 & x $\odot$ grad (sign) & 00:02 & 00:04 \\
 & IG (abs) & 00:04 & 00:09 \\
 & IG (sign) & 00:04 & 00:09 \\
 & LOO (abs) & 00:09 & 00:18 \\
 & LOO (sign) & 00:09 & 00:18 \\
 & Random & 00:00 & 00:00 \\
\cmidrule{1-4}
\multirow[c]{10}{*}{Anemia} & Beam & -- & -- \\
 & Grad ($L_1$) & 00:02 & 00:06 \\
 & Grad ($L_2$) & 00:02 & 00:06 \\
 & x $\odot$ grad (abs) & 00:01 & 00:05 \\
 & x $\odot$ grad (sign) & 00:01 & 00:05 \\
 & IG (abs) & 00:23 & 01:08 \\
 & IG (sign) & 00:23 & 01:08 \\
 & LOO (abs) & 02:23 & 06:58 \\
 & LOO (sign) & 02:23 & 07:01 \\
 & Random & 00:00 & 00:00 \\
\cmidrule{1-4}
\multirow[c]{10}{*}{Diabetes} & Beam & -- & -- \\
 & Grad ($L_1$) & 00:03 & 00:07 \\
 & Grad ($L_2$) & 00:03 & 00:07 \\
 & x $\odot$ grad (abs) & 00:02 & 00:06 \\
 & x $\odot$ grad (sign) & 00:02 & 00:06 \\
 & IG (abs) & 00:32 & 01:34 \\
 & IG (sign) & 00:32 & 01:34 \\
 & LOO (abs) & 03:19 & 09:44 \\
 & LOO (sign) & 03:17 & 09:45 \\
 & Random & 00:00 & 00:00 \\
\bottomrule
\end{tabular}
}
    \end{subtable}
    \begin{subtable}[t]{0.275\textwidth}
    \resizebox{0.95\linewidth}{!}{\begin{tabular}[t]{p{1.1cm}ccc}
\toprule
Dataset & IM & \multicolumn{2}{c}{Walltime [hh:mm]} \\
\cmidrule(r){3-4}
& & RoBERTa- & RoBERTa- \\
& & base & large \\
\midrule
\multirow[c]{10}{*}{MRPC} & Beam & 00:14 & 00:33 \\
 & Grad ($L_1$) & 00:02 & 00:04 \\
 & Grad ($L_2$) & 00:02 & 00:04 \\
 & x $\odot$ grad (abs) & 00:02 & 00:04 \\
 & x $\odot$ grad (sign) & 00:02 & 00:04 \\
 & IG (abs) & 00:03 & 00:07 \\
 & IG (sign) & 00:03 & 00:07 \\
 & LOO (abs) & 00:08 & 00:17 \\
 & LOO (sign) & 00:08 & 00:17 \\
 & Random & 00:00 & 00:00 \\
\cmidrule{1-4}
\multirow[c]{10}{*}{RTE} & Beam & 01:32 & 04:26 \\
 & Grad ($L_1$) & 00:02 & 00:04 \\
 & Grad ($L_2$) & 00:02 & 00:04 \\
 & x $\odot$ grad (abs) & 00:02 & 00:04 \\
 & x $\odot$ grad (sign) & 00:02 & 00:04 \\
 & IG (abs) & 00:04 & 00:09 \\
 & IG (sign) & 00:04 & 00:09 \\
 & LOO (abs) & 00:10 & 00:22 \\
 & LOO (sign) & 00:10 & 00:22 \\
 & Random & 00:00 & 00:00 \\
\cmidrule{1-4}
\multirow[c]{10}{*}{SST2} & Beam & 00:18 & 00:43 \\
 & Grad ($L_1$) & 00:02 & 00:04 \\
 & Grad ($L_2$) & 00:02 & 00:04 \\
 & x $\odot$ grad (abs) & 00:02 & 00:04 \\
 & x $\odot$ grad (sign) & 00:02 & 00:04 \\
 & IG (abs) & 00:04 & 00:09 \\
 & IG (sign) & 00:04 & 00:09 \\
 & LOO (abs) & 00:09 & 00:19 \\
 & LOO (sign) & 00:10 & 00:19 \\
 & Random & 00:00 & 00:00 \\
\cmidrule{1-4}
\multirow[c]{10}{*}{SNLI} & Beam & 01:10 & 02:38 \\
 & Grad ($L_1$) & 00:05 & 00:07 \\
 & Grad ($L_2$) & 00:06 & 00:07 \\
 & x $\odot$ grad (abs) & 00:05 & 00:06 \\
 & x $\odot$ grad (sign) & 00:04 & 00:06 \\
 & IG (abs) & 00:21 & 00:57 \\
 & IG (sign) & 00:21 & 00:56 \\
 & LOO (abs) & 00:12 & 00:26 \\
 & LOO (sign) & 00:12 & 00:26 \\
 & Random & 00:01 & 00:00 \\
\cmidrule{1-4}
\multirow[c]{10}{*}{IMDB} & Beam & -- & -- \\
 & Grad ($L_1$) & 00:34 & 01:18 \\
 & Grad ($L_2$) & 00:34 & 01:17 \\
 & x $\odot$ grad (abs) & 00:22 & 01:03 \\
 & x $\odot$ grad (sign) & 00:22 & 01:03 \\
 & IG (abs) & 06:49 & 20:08 \\
 & IG (sign) & 06:54 & 20:09 \\
 & LOO (abs) & 25:02 & 73:17 \\
 & LOO (sign) & 24:48 & 72:55 \\
 & Random & 00:01 & 00:01 \\
\cmidrule{1-4}
\multirow[c]{10}{*}{MNLI} & Beam & 05:44 & 15:34 \\
 & Grad ($L_1$) & 00:05 & 00:11 \\
 & Grad ($L_2$) & 00:05 & 00:11 \\
 & x $\odot$ grad (abs) & 00:04 & 00:09 \\
 & x $\odot$ grad (sign) & 00:04 & 00:09 \\
 & IG (abs) & 00:35 & 01:35 \\
 & IG (sign) & 00:35 & 01:34 \\
 & LOO (abs) & 00:19 & 00:46 \\
 & LOO (sign) & 00:19 & 00:46 \\
 & Random & 00:00 & 00:00 \\
\cmidrule{1-4}
\multirow[c]{10}{*}{QNLI} & Beam & 06:39 & 18:51 \\
 & Grad ($L_1$) & 00:04 & 00:08 \\
 & Grad ($L_2$) & 00:04 & 00:08 \\
 & x $\odot$ grad (abs) & 00:03 & 00:07 \\
 & x $\odot$ grad (sign) & 00:03 & 00:08 \\
 & IG (abs) & 00:23 & 01:03 \\
 & IG (sign) & 00:23 & 01:04 \\
 & LOO (abs) & 00:17 & 00:43 \\
 & LOO (sign) & 00:17 & 00:43 \\
 & Random & 00:00 & 00:00 \\
\cmidrule{1-4}
\multirow[c]{10}{*}{QQP} & Beam & 04:44 & 11:12 \\
 & Grad ($L_1$) & 00:12 & 00:26 \\
 & Grad ($L_2$) & 00:12 & 00:26 \\
 & x $\odot$ grad (abs) & 00:10 & 00:22 \\
 & x $\odot$ grad (sign) & 00:10 & 00:22 \\
 & IG (abs) & 01:48 & 04:57 \\
 & IG (sign) & 01:48 & 04:59 \\
 & LOO (abs) & 00:36 & 01:24 \\
 & LOO (sign) & 00:36 & 01:23 \\
 & Random & 00:01 & 00:01 \\
\midrule
\midrule
\multicolumn{2}{r}{sum} & 145:00 & 406:58 \\
\multicolumn{2}{r}{x5 seeds} & 725:02 & 2034:53 \\
\bottomrule
\end{tabular}
}
    \end{subtable}
    \label{tab:appendix:walltime:im}
\end{table*}

\clearpage
\section{Masked fine-tuning}
\label{appendix:masked-fine-tuning}

In \Cref{sec:experiment:fine-tune}, we show selected results for unmasked performance and 100\% masked performance. In this appendix, we extend those results to all 16 datasets. In addition to this, this appendix contains a more detailed ablation study, where the training strategy and validation strategy are considered separate. As such, the results in \Cref{sec:experiment:fine-tune} are a strict subset of these detailed results. In \Cref{tab:appendix:masked-fine-tuning:translation} we show how the terminologies relate.

\begin{table}[H]
    \centering
    \caption{This table relates terminologies between the fine-tuning strategies mentioned in \Cref{sec:experiment:fine-tune} and the training strategy and validation strategy terms.\vspace{0.1in}}
    {\footnotesize
    \begin{tabular}{p{1.8cm}cc}
        \toprule
        \Cref{sec:experiment:fine-tune} & Training strategy & Validation strategy \\
        \midrule
         {Masked\linebreak fine-tuning} & Use 50/50 & Use both \\
         {Plain\linebreak fine-tuning} & No masking & No masking \\
         Only masking & Masking & Masking \\
        \bottomrule
    \end{tabular}}
    \label{tab:appendix:masked-fine-tuning:translation}
\end{table}

\paragraph{Training strategy} The training strategy applies to the training dataset during fine-tuning.

\begin{description}[noitemsep,topsep=0pt]
  \item[No masking] No masking is applied to the training dataset. This is what is ordinarily done in the literature.
  \item[Masking] Masking is applied to every observation. The masking is uniformly sampled, at a masking rate between 0\% and 100\%.
  \item[Use 50/50] Half of the mini-batch using the \emph{No masking} strategy and the other half use the \emph{Masking} strategy.
\end{description}

\paragraph{Validation strategy} The validation dataset is used to select the optional epoch. This is similar to early stopping, but rather than stopping immediately. The training continues, and the best epoch is chosen at the end of the training.

The validation strategy applies to the validation dataset during fine-tuning.

\begin{description}[noitemsep,topsep=0pt]
  \item[No masking] No masking is applied to the validation dataset. This is what is ordinarily done in the literature.
  \item[Masking] Masking is applied to every observation. The masking is uniformly sampled, at a masking rate between 0\% and 100\%.
  \item[Use both] A copy of the validation dataset has the \emph{No masking} strategy applied to it. Another copy of the validation dataset has the \emph{Masking} strategy applied to it. As such, the validation dataset is twice as long, but it does not add additional observations or inforamtion.
\end{description}

\subsection{Findings}

We generally find that the choice of validation strategy when using the \emph{Use 50/50} training strategy is not important. Interestingly, \emph{Masking} for the validation dataset and \emph{No masking} for the training dataset often works too.

However, because \emph{Use 50/50} for training strategy and \emph{Use both} for validation strategy, i.e. masked fine-tuning, work well in all cases and is theoretically sound, this is the approach we recommend and use throughout the paper.

\subsection{All datasets aggregation}

In \Cref{sec:experiment:fine-tune} we also include an \emph{All} ``dataset''. This is a simple arithmetic mean over all the performance of all 16 datasets. This is similar to how the GLUE benchmark \citep{Wang2019} works. To compute the confidence interval, a dataset-aggregation is done for each seed, such that the all-observation are i.i.d..

Because some seeds do not converge for some datasets, such as bAbI-2 and bAbI-3 (as mentioned in \Cref{sec:experiment:fine-tune}), those outliers and not included in the aggregation, also hyperparameter optimization will likely help. For complete transparency, we do include them in the statistics for the individual datasets and show all individual performances with a (+) symbol.

\begin{figure}[h]
    \centering
    \includegraphics[trim=0pt 102pt 0pt 7pt, clip, width=\linewidth]{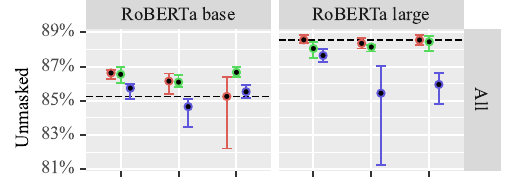}
    \includegraphics[trim=0pt 7pt 0pt 20.5pt, clip, width=\linewidth]{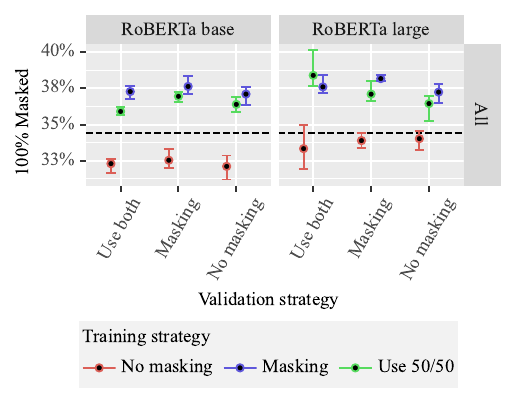}
    \caption{The all aggregation for the 100\% masked performance and unmasked performance. The baseline (dashed line) for 100\% masked performance is the class-majority baseline. Unmasked performance is when using no masking for both validation and training.}
    \label{fig:appendix:masked-fine-tuning:all-aggregate}
\end{figure}

\onecolumn
\subsection{Test dataset}

\begin{figure}[H]
    \centering
    \includegraphics[trim=0pt 7pt 0pt 7pt, clip, width=\linewidth]{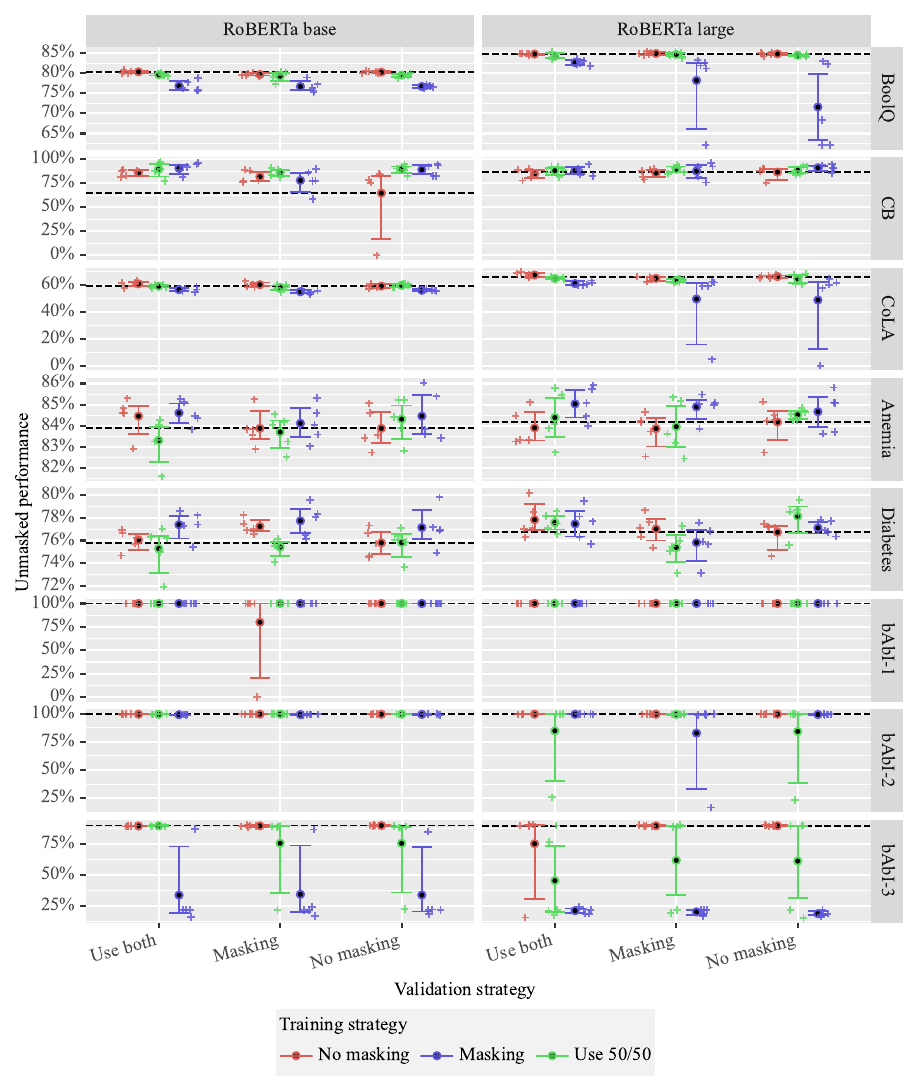}
    \caption{The unmasked performance for each validation and training strategy, using the test dataset. Not that \emph{``No masking''} as a \emph{training strategy} is not a valid option only a baseline, as it creates OOD issues. We find that the multi-task \emph{training strategy} \emph{``Use 50/50''} works best. This plot is \textbf{page-1}. Corresponding main results in \Cref{fig:paper:unmasked-performance}.}
    \label{fig:appendix:unmasked-performance:p1}
\end{figure}

\begin{figure}[H]
    \centering
    \includegraphics[trim=0pt 7pt 0pt 7pt, clip, width=\linewidth]{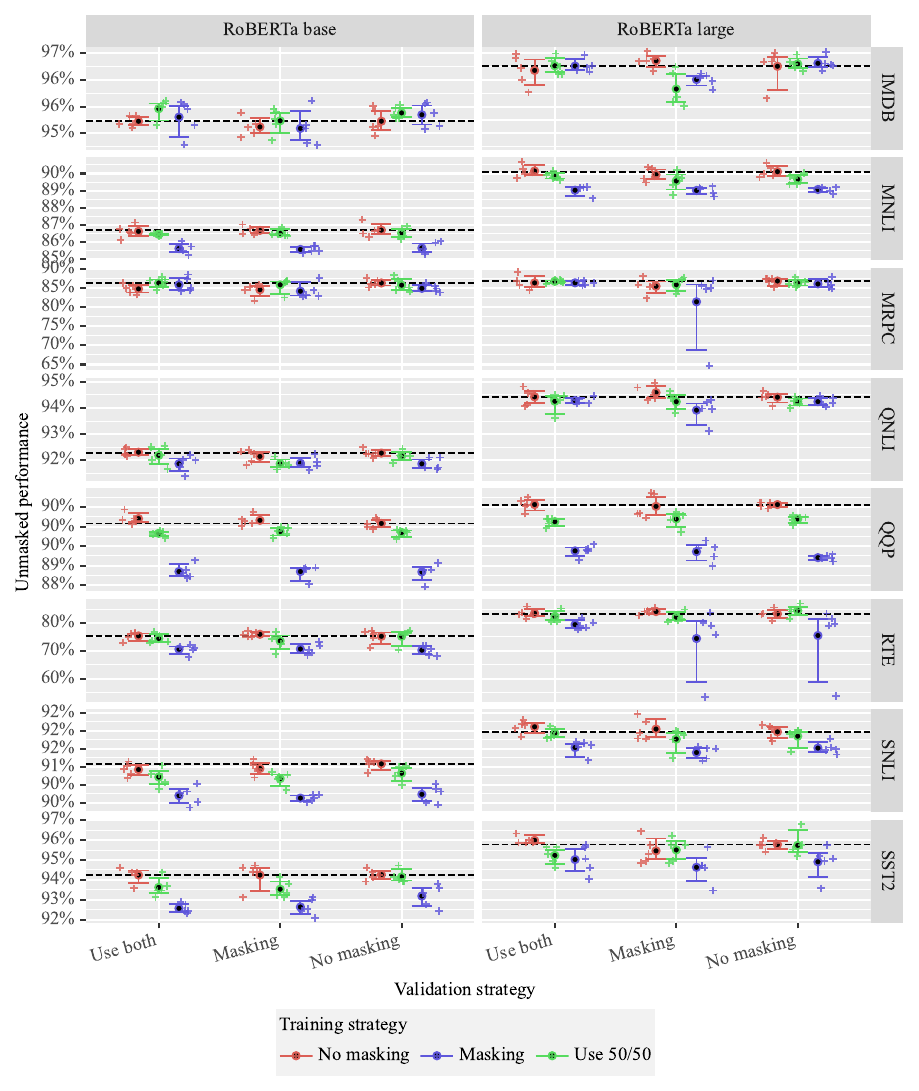}
    \caption{The unmasked performance for each validation and training strategy, using the test dataset. Not that \emph{``No masking''} as a \emph{training strategy} is not a valid option only a baseline, as it creates OOD issues. We find that the multi-task \emph{training strategy} \emph{``Use 50/50''} works best. This plot is \textbf{page-2}. Corresponding main results in \Cref{fig:paper:unmasked-performance}.}
    \label{fig:appendix:unmasked-performance:p2}
\end{figure}

\begin{figure}[H]
    \centering
    \includegraphics[width=\linewidth]{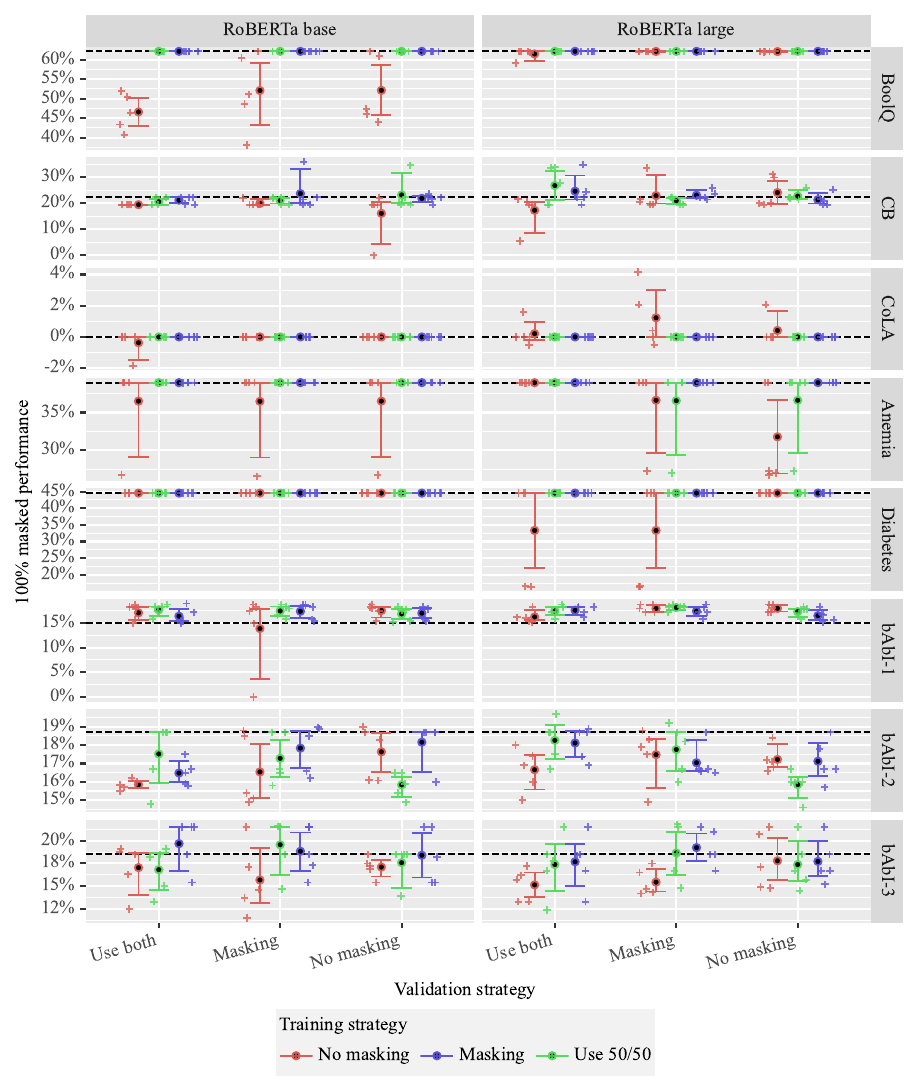}
    \caption{The 100\% masked performance, using the test dataset. The dashed line represents the class-majority classifier baseline. Results show that masking during training (\emph{``Masking''} or \emph{``Use 50/50''}) is necessary. This plot is \textbf{page-1}. Corresponding main paper results in \Cref{fig:paper:masked-100p-performance}.}
    \label{fig:appendix:masked-100p-performance:p1}
\end{figure}

\begin{figure}[H]
    \centering
    \includegraphics[width=\linewidth]{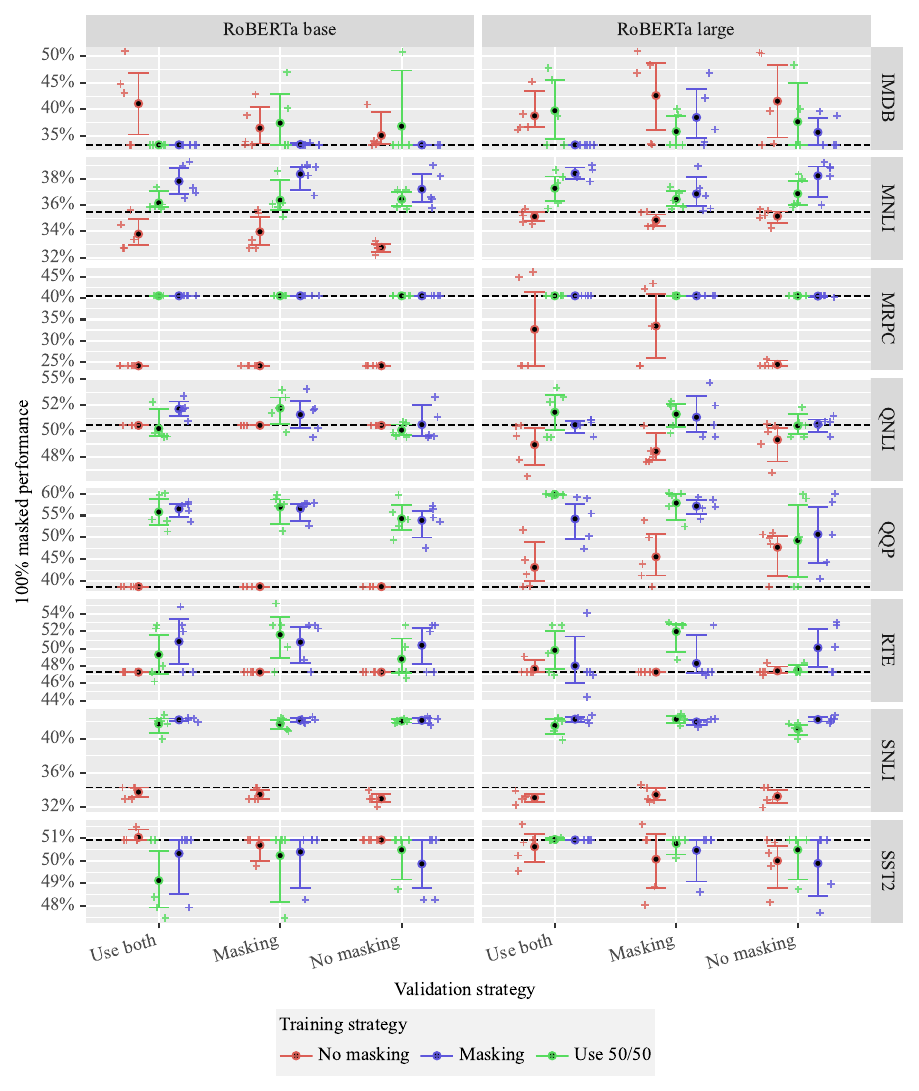}
    \caption{The 100\% masked performance, using the test dataset. The dashed line represents the class-majority classifier baseline. Results show that masking during training (\emph{``Masking''} or \emph{``Use 50/50''}) is necessary. This plot is \textbf{page-2}. Corresponding main paper results in \Cref{fig:paper:masked-100p-performance}.}
    \label{fig:appendix:masked-100p-performance:p2}
\end{figure}

\subsection{Validation dataset}

\begin{figure}[H]
    \centering
    \includegraphics[width=\linewidth]{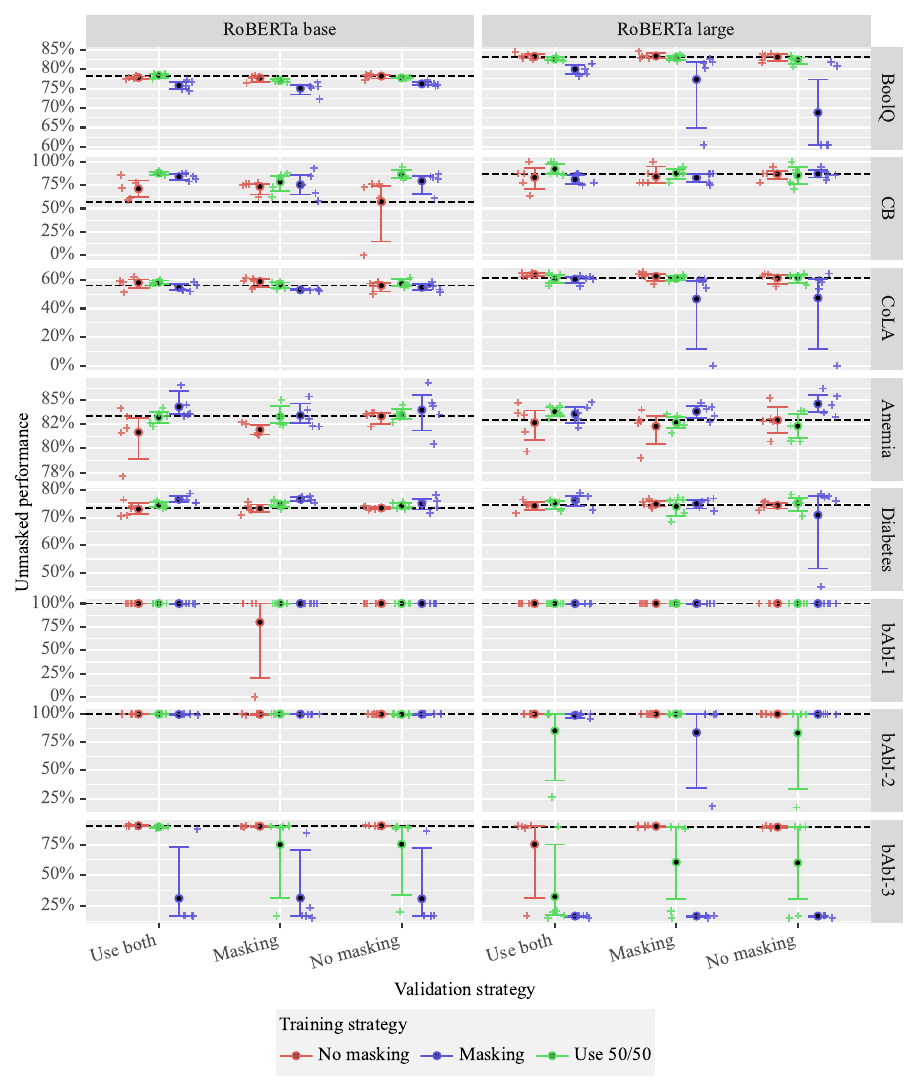}
    \caption{The unmasked performance for each validation and training strategy, using the validation dataset. Not that \emph{``No masking''} as a \emph{training strategy} is not a valid option only a baseline, as it creates OOD issues. We find that the multi-task \emph{training strategy} \emph{``Use 50/50''} works best. This plot is \textbf{page-1}.}
\end{figure}

\begin{figure}[H]
    \centering
    \includegraphics[width=\linewidth]{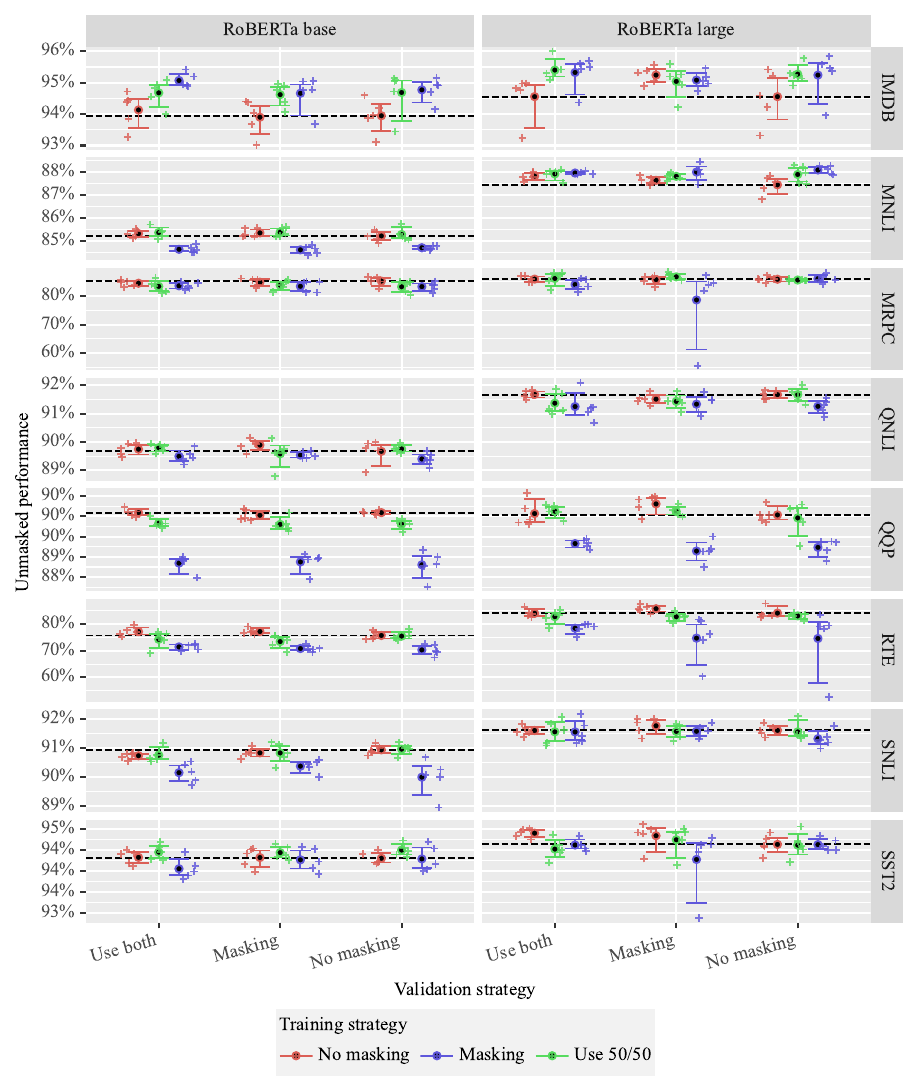}
    \caption{The unmasked performance for each validation and training strategy, using the validation dataset. Not that \emph{``No masking''} as a \emph{training strategy} is not a valid option only a baseline, as it creates OOD issues. We find that the multi-task \emph{training strategy} \emph{``Use 50/50''} works best. This plot is \textbf{page-2}.}
\end{figure}

\begin{figure}[H]
    \centering
    \includegraphics[width=\linewidth]{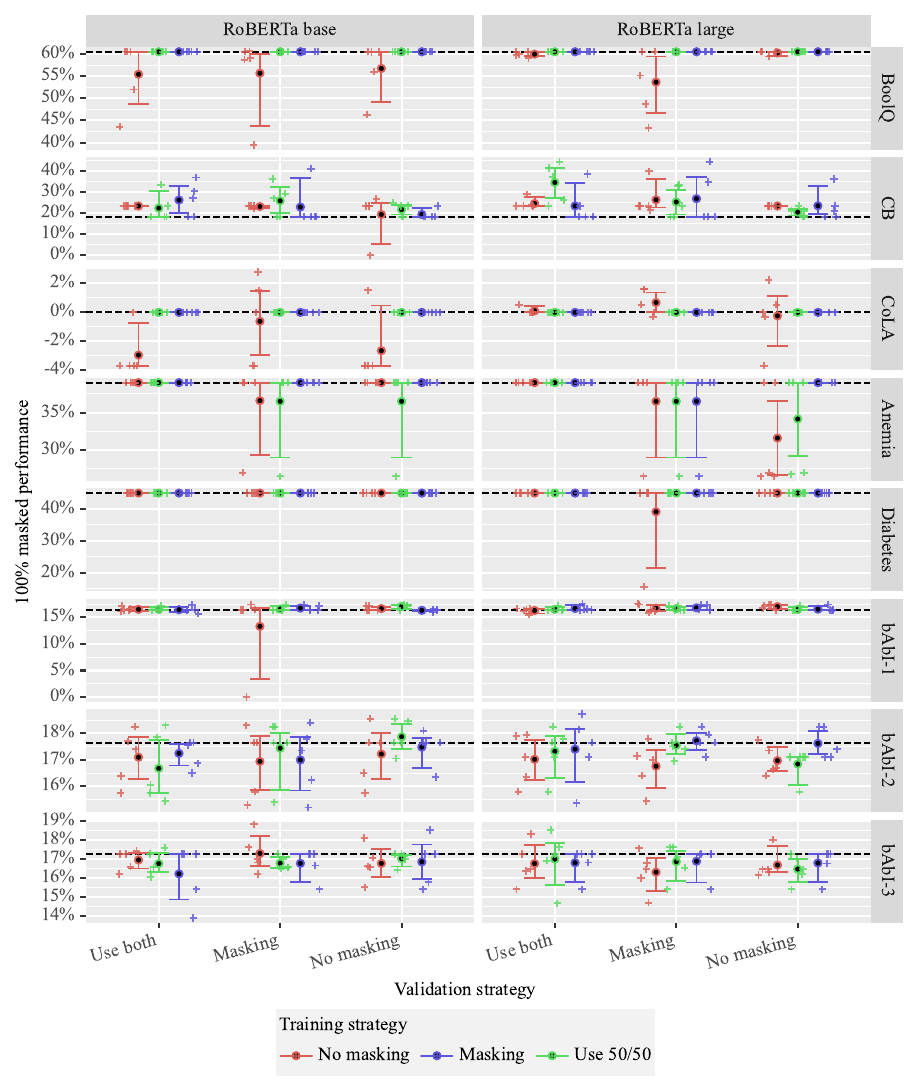}
    \caption{The 100\% masked performance, using the validation dataset. The dashed line represents the class-majority classifier baseline. Results show that masking during training (\emph{``Masking''} or \emph{``Use 50/50''}) is necessary. This plot is \textbf{page-1}.}
\end{figure}

\begin{figure}[H]
    \centering
    \includegraphics[width=\linewidth]{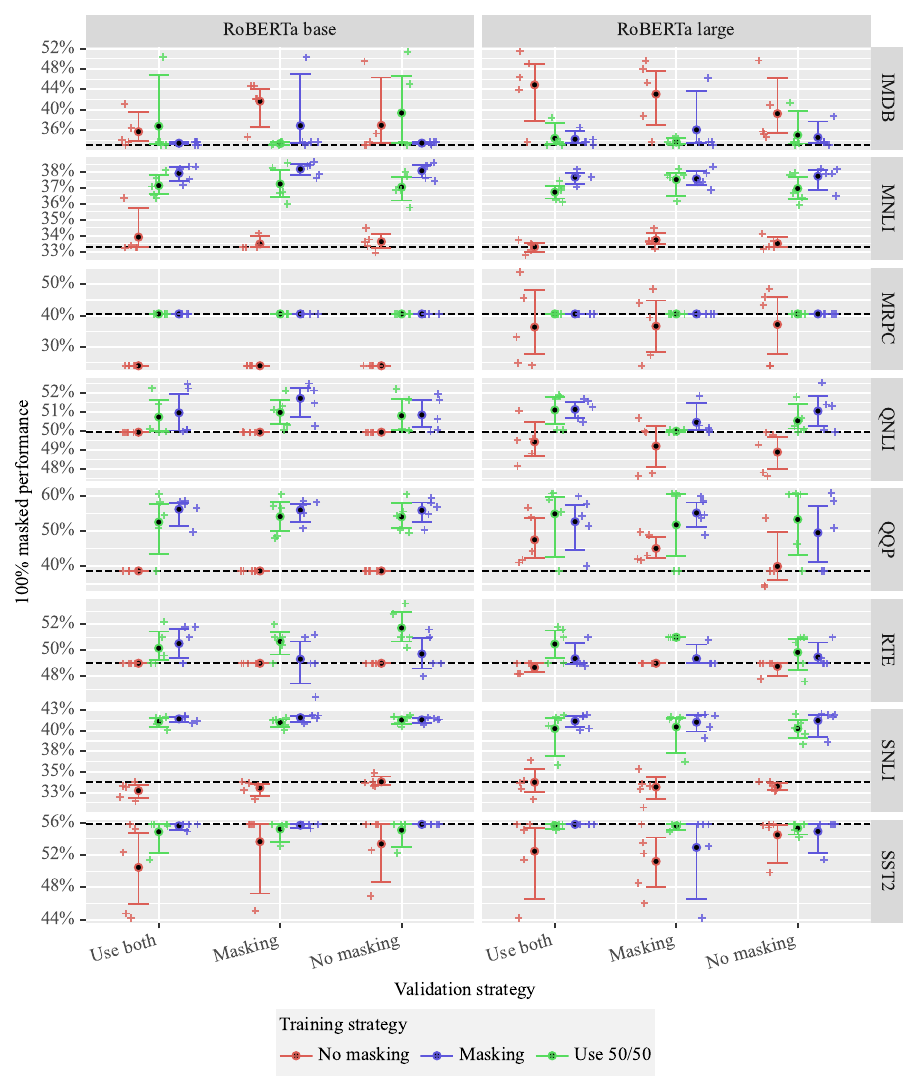}
    \caption{The 100\% masked performance, using the validation dataset. The dashed line represents the class-majority classifier baseline. Results show that masking during training (\emph{``Masking''} or \emph{``Use 50/50''}) is necessary. This plot is \textbf{page-2}.}
\end{figure}

\section{Convergence speed}
\label{appendix:epochs}
\begin{figure}[H]
    \centering
    \includegraphics[width=\linewidth]{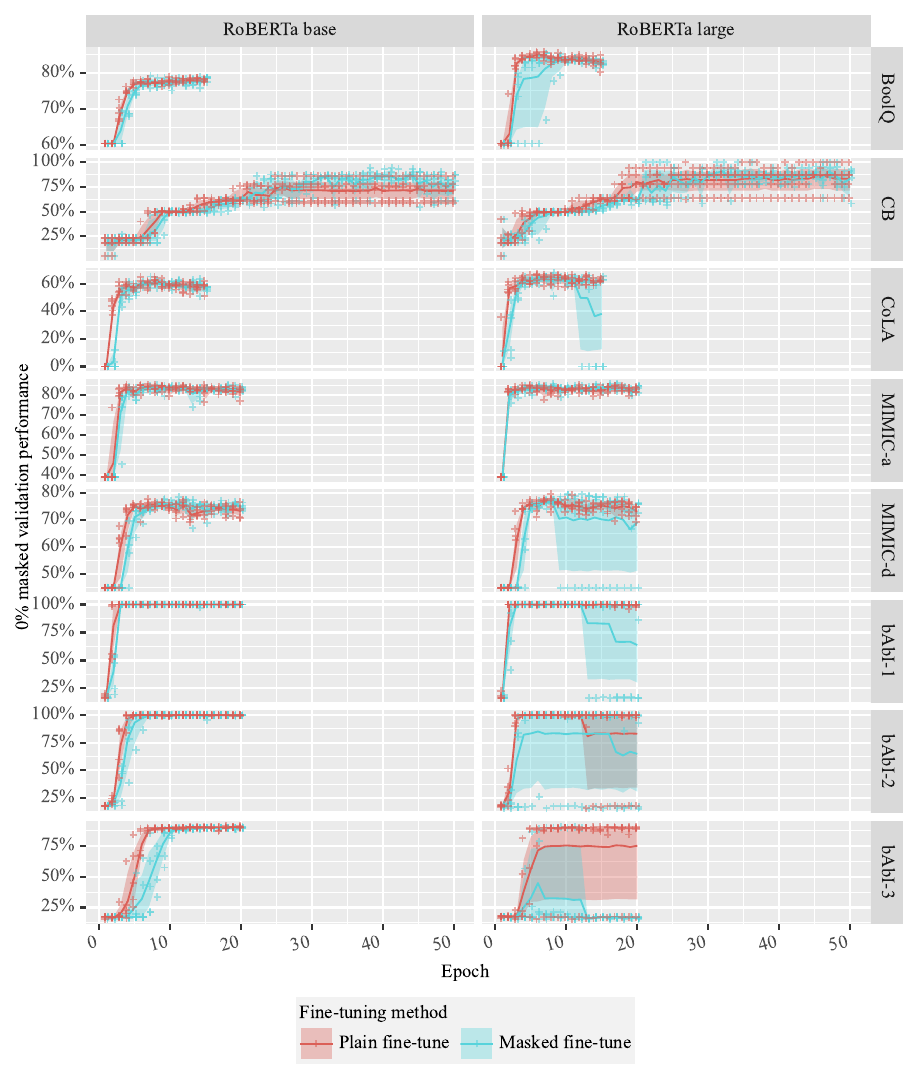}
    \caption{The validation performance for each epoch. Note that the max number of epochs vary depending on the dataset. This is only to limit the compute requirements when fine-tuning. The best epoch is selected by the ``early-stopping'' dataset, which has one copy with no masking and one copy with uniformly sampled masking ratios. This plot is \textbf{page-1}.}
    \label{fig:appendix:epochs:p1}
\end{figure}

\begin{figure}[H]
    \centering
    \includegraphics[width=\linewidth]{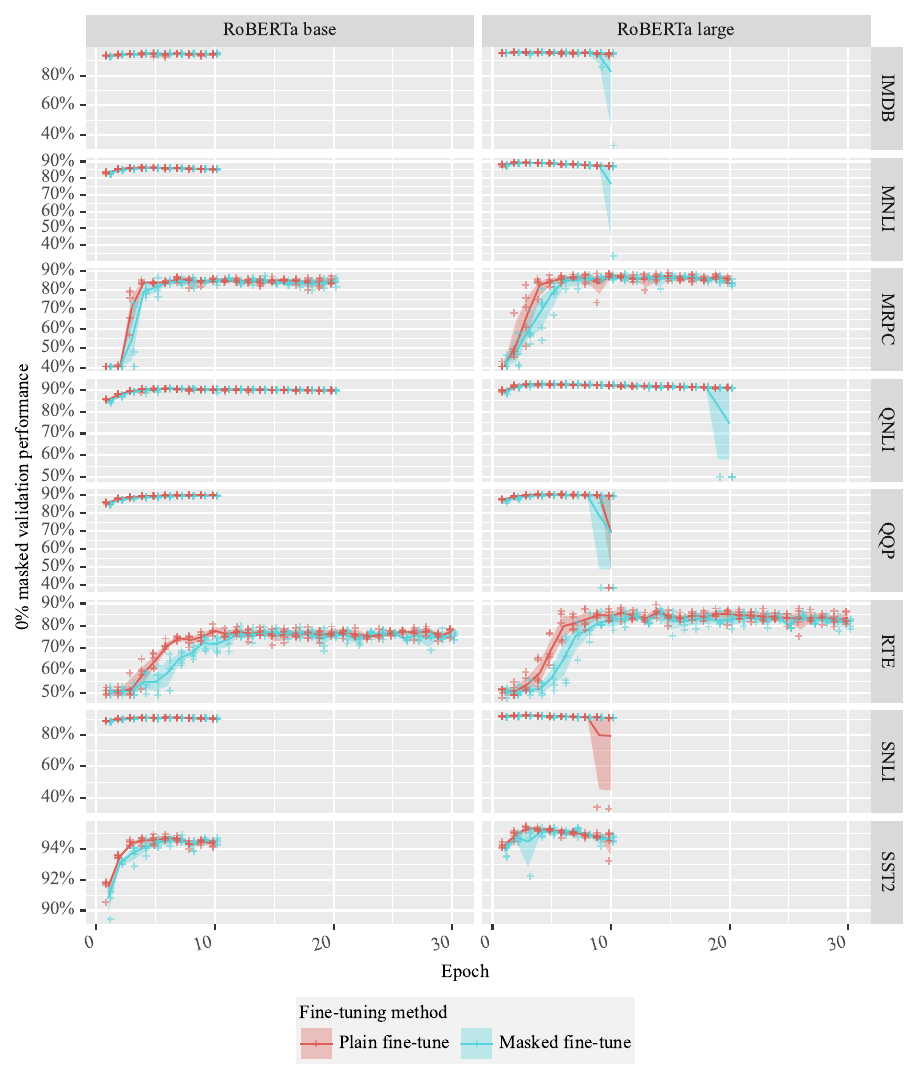}
    \caption{The validation performance for each epoch. Note that the max number of epochs vary depending on the dataset. This is only to limit the compute requirements when fine-tuning. The best epoch is selected by the ``early-stopping'' dataset, which has one copy with no masking and one copy with uniformly sampled masking ratios. This plot is \textbf{page-2}.}
    \label{fig:appendix:epochs:p2}
\end{figure}

\section{In-distribution validation}
\label{sec:appendix:ood}

\begin{figure}[H]
    \centering
    \includegraphics[width=\linewidth]{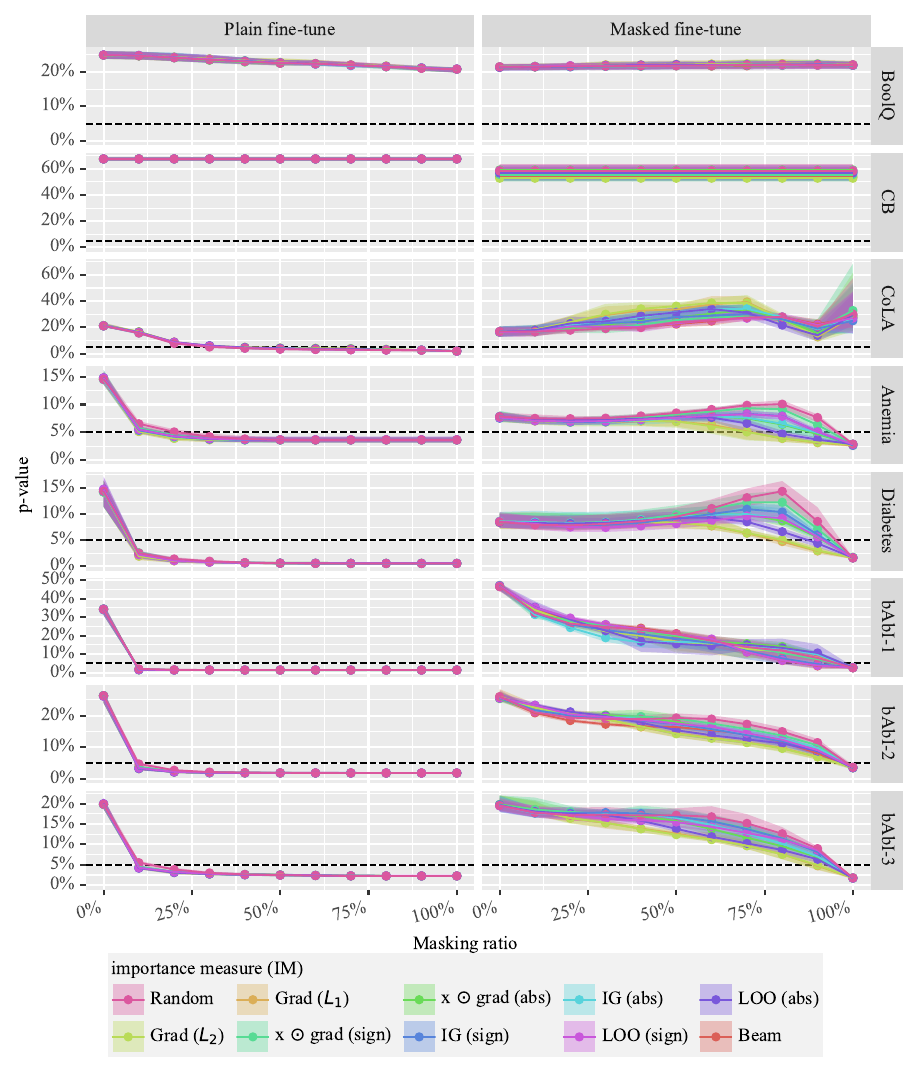}
    \caption{In-distribution p-values using MaSF, for \textbf{RoBERTa-base} with and without masked fine-tuning, \textbf{page-1}. The masked tokens are chosen according to an importance measure. P-values below the dashed line show out-of-distribution (OOD) results, given a 5\% risk of a false positive. Results show that only when using masked fine-tuning, masked data is consistently not OOD. Corresponding main paper results are in \Cref{fig:paper:ood}.}
    \label{fig:appendix:ood:roberta-sb:p1}
\end{figure}

\begin{figure}[H]
    \centering
    \includegraphics[width=\linewidth]{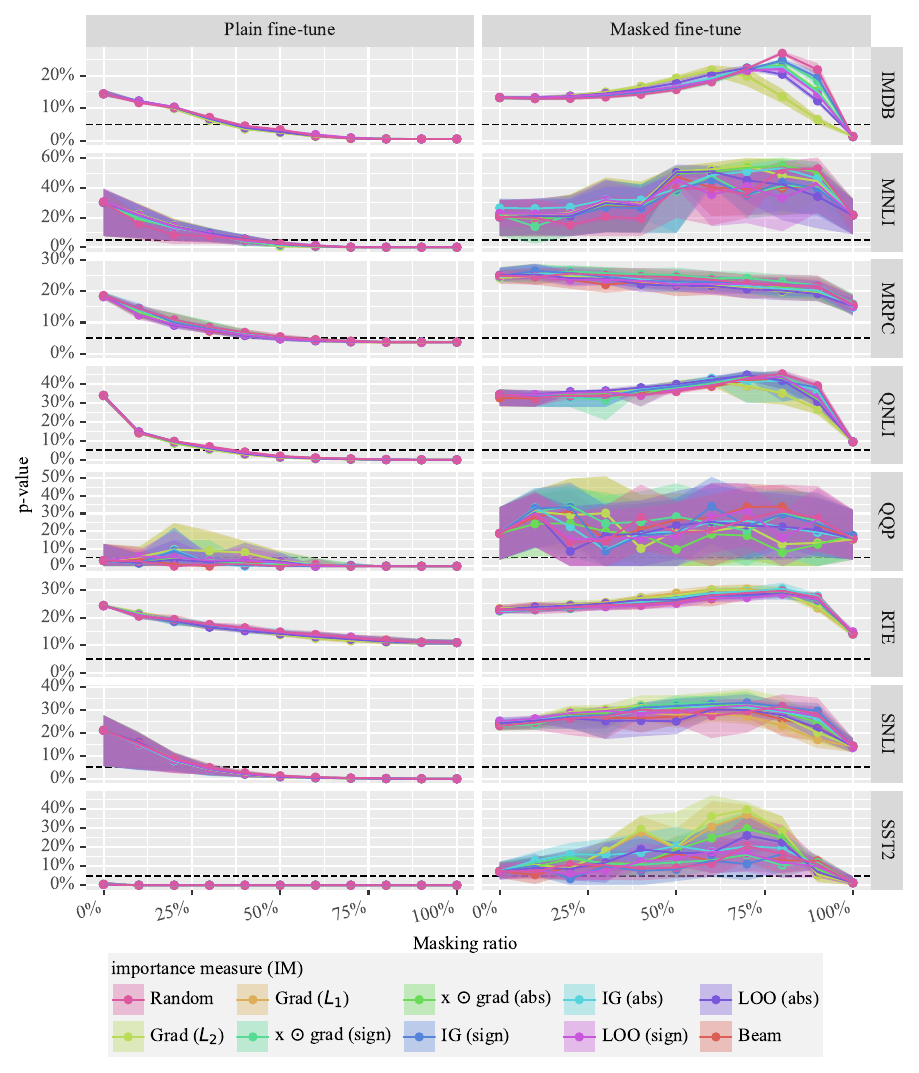}
    \caption{In-distribution p-values using MaSF, for \textbf{RoBERTa-base} with and without masked fine-tuning, \textbf{page-2}. The masked tokens are chosen according to an importance measure. P-values below the dashed line show out-of-distribution (OOD) results, given a 5\% risk of a false positive. Results show that only when using masked fine-tuning, masked data is consistently not OOD. Corresponding main paper results are in \Cref{fig:paper:ood}.}
    \label{fig:appendix:ood:roberta-sb:p2}
\end{figure}

\begin{figure}[H]
    \centering
    \includegraphics[width=\linewidth]{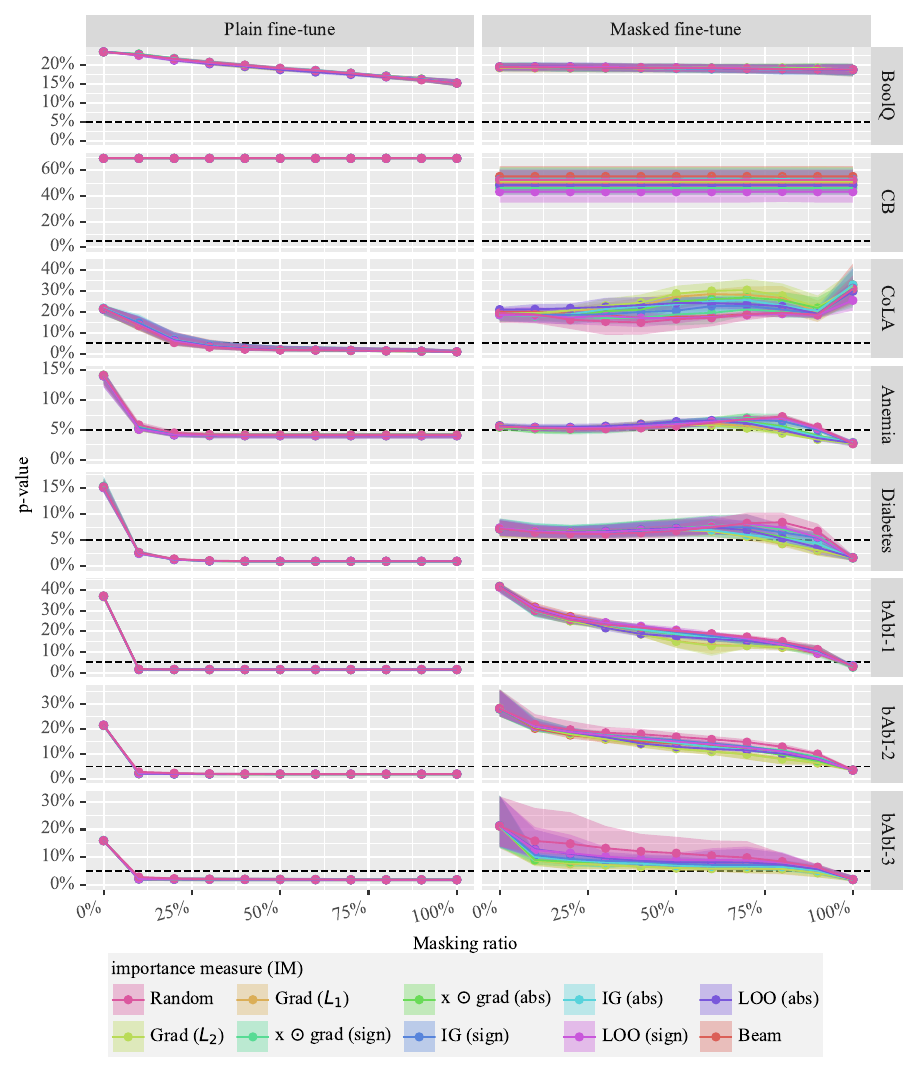}
    \caption{In-distribution p-values using MaSF, for \textbf{RoBERTa-large} with and without masked fine-tuning, \textbf{page-1}. The masked tokens are chosen according to an importance measure. P-values below the dashed line show out-of-distribution (OOD) results, given a 5\% risk of a false positive. Results show that only when using masked fine-tuning, masked data is consistently not OOD. Corresponding main paper results are in \Cref{fig:paper:ood}.}
    \label{fig:appendix:ood:roberta-sl:p1}
\end{figure}

\begin{figure}[H]
    \centering
    \includegraphics[width=\linewidth]{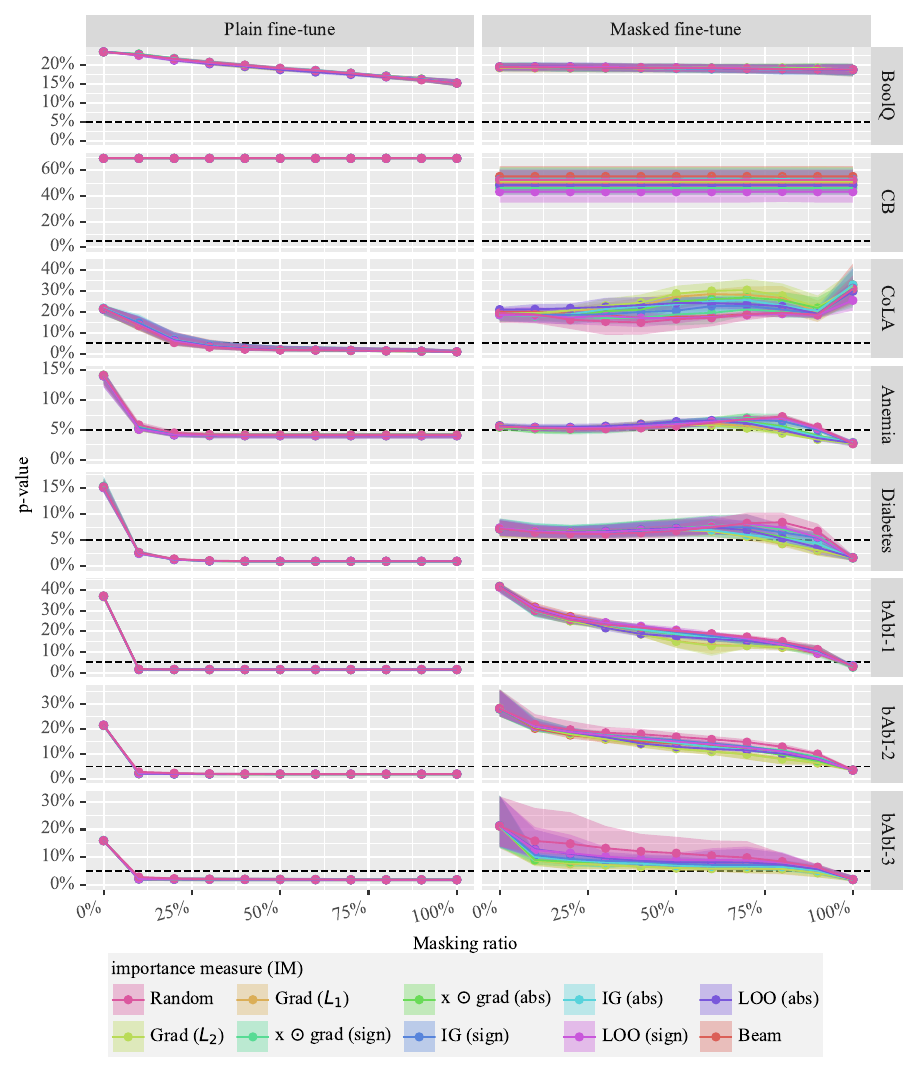}
    \caption{In-distribution p-values using MaSF, for \textbf{RoBERTa-large} with and without masked fine-tuning, \textbf{page-2}. The masked tokens are chosen according to an importance measure. P-values below the dashed line show out-of-distribution (OOD) results, given a 5\% risk of a false positive. Results show that only when using masked fine-tuning, masked data is consistently not OOD. Corresponding main paper results are in \Cref{fig:paper:ood}.}
    \label{fig:appendix:ood:roberta-sl:p2}
\end{figure}

\section{Faithfulness metrics}
\label{appendix:faithfulness}

\begin{figure}[H]
    \centering
    \includegraphics[width=\linewidth]{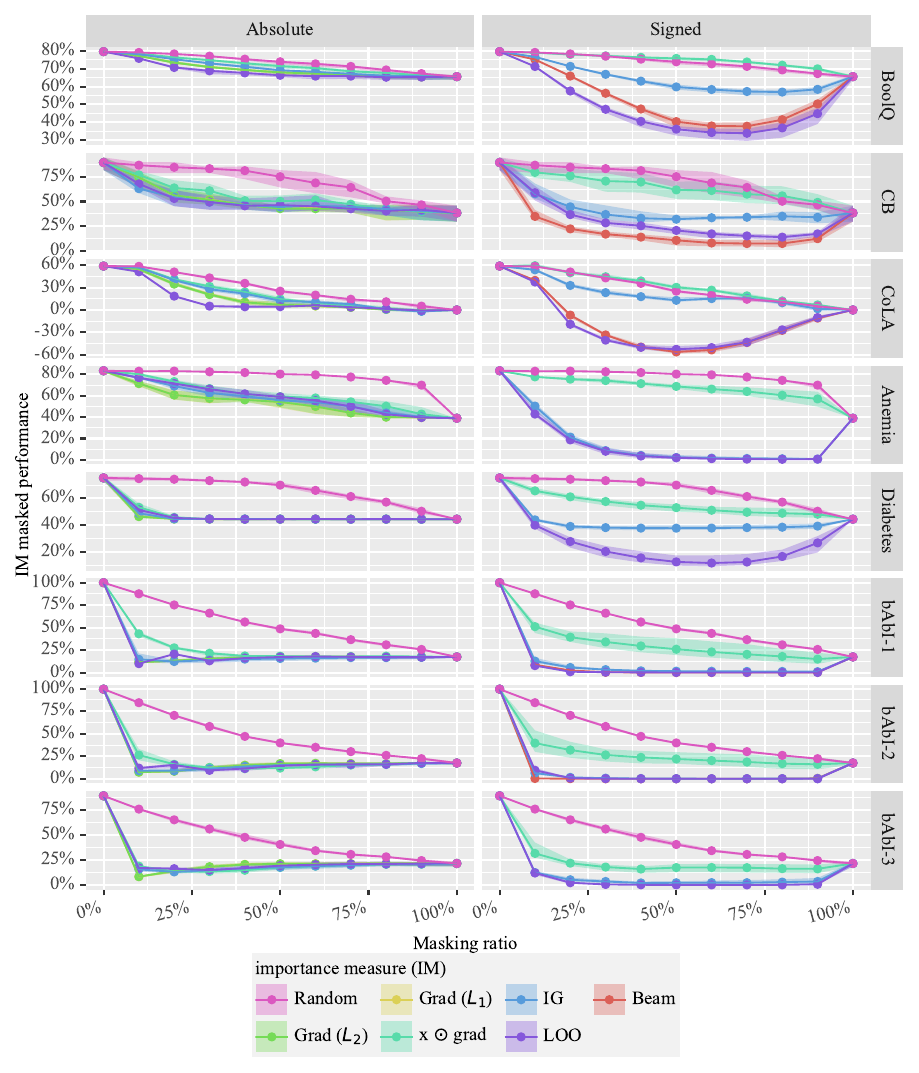}
    \caption{The performance given the masked datasets, where masking is done for the x\% allegedly most important tokens according to the importance measure. If the performance for a given explanation is below the \emph{``Random''} baseline, this shows faithfulness. Although, faithfulness is not an absolute concept, so more is better. This plot is \textbf{page-1} for \textbf{RoBERTa-base}. Corresponding main paper results in \Cref{sec:experiment:faithfulness}.}
    \label{fig:appendix:faithfulness:roberta-sb:p1}
\end{figure}

\begin{figure}[H]
    \centering
    \includegraphics[width=\linewidth]{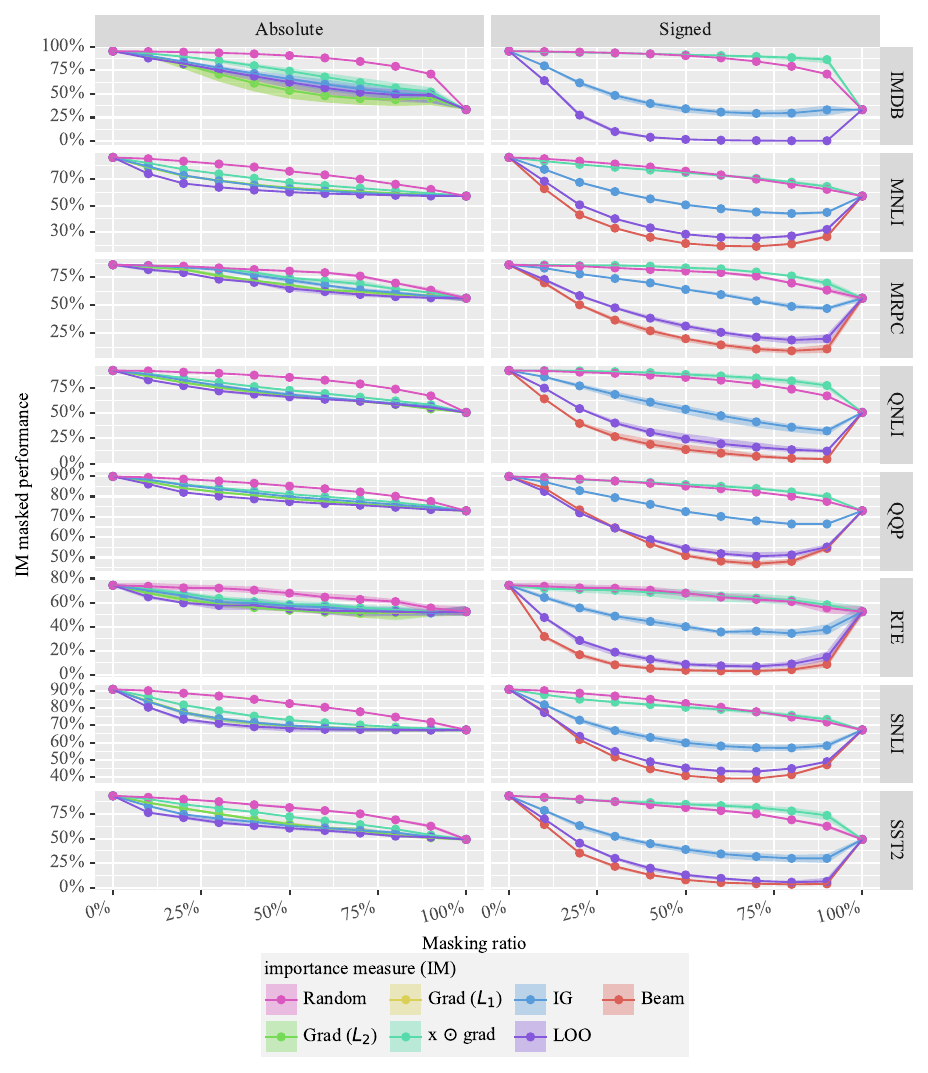}
    \caption{The performance given the masked datasets, where masking is done for the x\% allegedly most important tokens according to the importance measure. If the performance for a given explanation is below the \emph{``Random''} baseline, this shows faithfulness. Although, faithfulness is not an absolute concept, so more is better. This plot is \textbf{page-2} for \textbf{RoBERTa-base}. Corresponding main paper results in \Cref{sec:experiment:faithfulness}.}
    \label{fig:appendix:faithfulness:roberta-sb:p2}
\end{figure}

\begin{figure}[H]
    \centering
    \includegraphics[width=\linewidth]{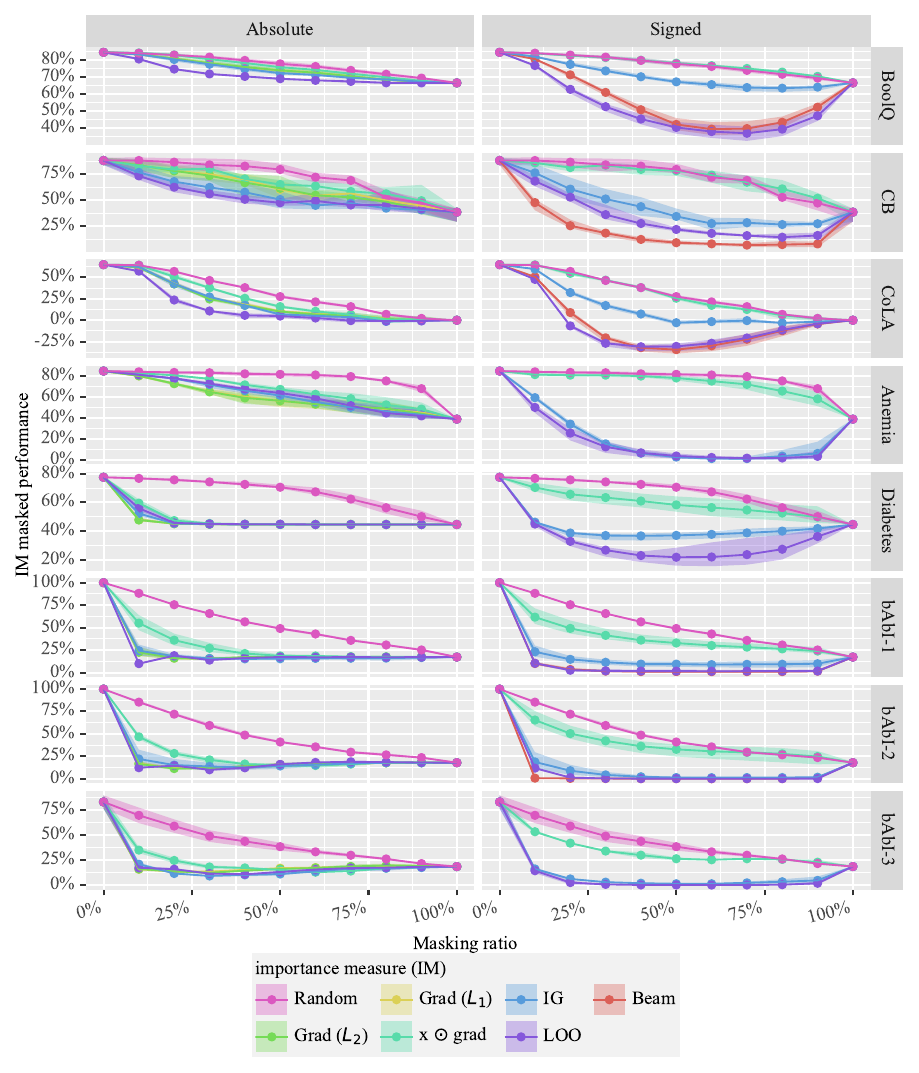}
    \caption{The performance given the masked datasets, where masking is done for the x\% allegedly most important tokens according to the importance measure. If the performance for a given explanation is below the \emph{``Random''} baseline, this shows faithfulness. Although, faithfulness is not an absolute concept, so more is better. This plot is \textbf{page-1} for \textbf{RoBERTa-large}. Corresponding main paper results in \Cref{sec:experiment:faithfulness}.}
    \label{fig:appendix:faithfulness:roberta-sl:p1}
\end{figure}

\begin{figure}[H]
    \centering
    \includegraphics[width=\linewidth]{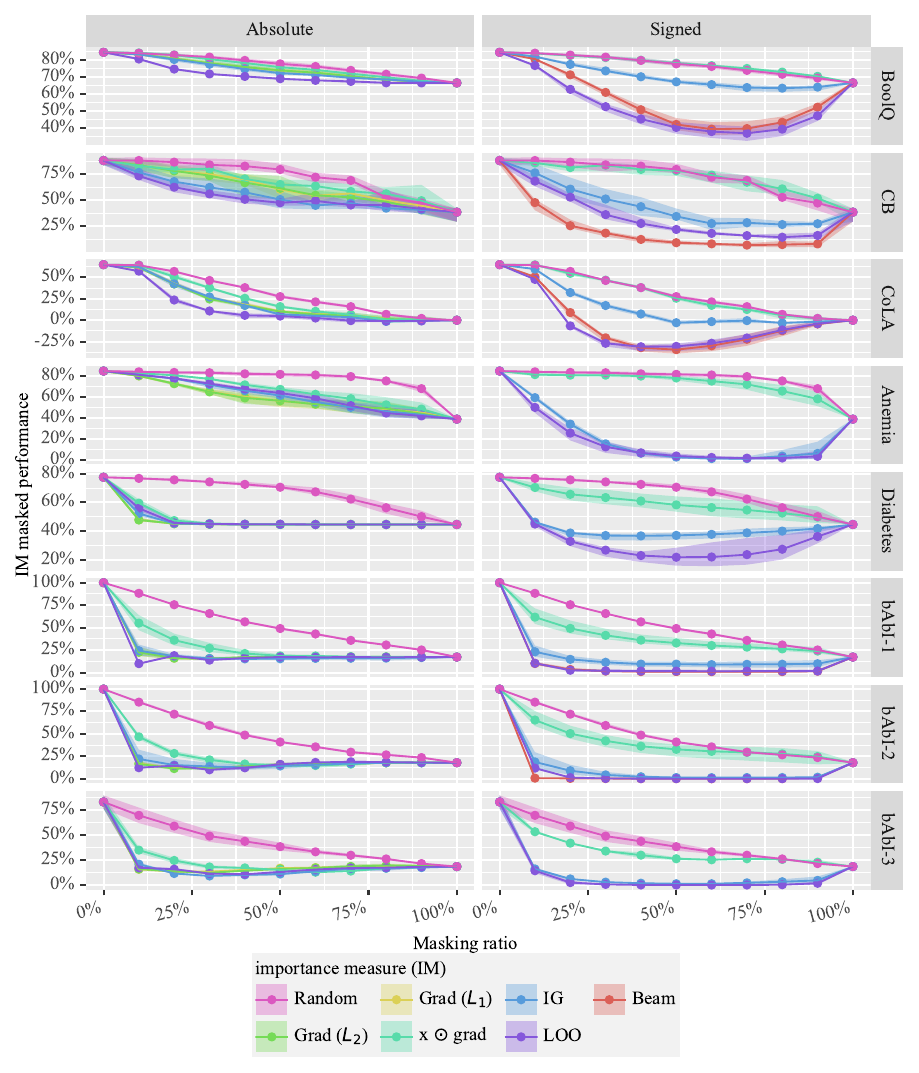}
    \caption{The performance given the masked datasets, where masking is done for the x\% allegedly most important tokens according to the importance measure. If the performance for a given explanation is below the \emph{``Random''} baseline, this shows faithfulness. Although, faithfulness is not an absolute concept, so more is better. This plot is \textbf{page-2} for \textbf{RoBERTa-large}. Corresponding main paper results in \Cref{sec:experiment:faithfulness}.}
    \label{fig:appendix:faithfulness:roberta-sl:p2}
\end{figure}

\subsection{Relative Area between Curves (RACU)}
\begin{table}[H]
    \centering
    \caption{Faithfulness scores for \textbf{RoBERTa-base}. Shows Relative Area Between Curves (RACU) and the non-relative variant (ACU), defined by \citet{Madsen2022}. Also compares with Recursive-ROAR by \citet{Madsen2022}.}
    \begin{subtable}[t]{0.35\textwidth}
    \resizebox{0.9855\linewidth}{!}{\begin{tabular}[t]{llccc}
\toprule
& & \multicolumn{3}{c}{Faithfulness [\%]}  \\
\cmidrule(r){3-5}
Dataset & IM & \multicolumn{2}{c}{Our} & R-ROAR \\
\cmidrule(r){3-4}
& & ACU & RACU & RACU \\
\midrule
\multirow[c]{9}{*}{bAbI-1} & Grad ($L_2$) & $32.6_{-1.1}^{+1.2}$ & $91.9_{-3.2}^{+2.0}$ & $64.2_{-2.6}^{+2.6}$ \\
 & Grad ($L_1$) & $32.7_{-0.9}^{+1.3}$ & $92.1_{-3.1}^{+2.0}$ & -- \\
 & x $\odot$ grad (sign) & $21.4_{-7.3}^{+4.9}$ & $60.5_{-20.7}^{+13.7}$ & -- \\
 & x $\odot$ grad (abs) & $27.0_{-1.4}^{+1.1}$ & $76.0_{-1.6}^{+3.0}$ & $52.1_{-3.7}^{+1.8}$ \\
 & IG (sign) & $44.1_{-0.8}^{+0.8}$ & $124.2_{-4.5}^{+3.3}$ & -- \\
 & IG (abs) & $33.3_{-2.4}^{+2.4}$ & $93.7_{-4.1}^{+5.5}$ & $48.2_{-5.7}^{+4.1}$ \\
 & LOO (sign) & $46.1_{-0.7}^{+0.7}$ & $129.9_{-3.5}^{+3.4}$ & -- \\
 & LOO (abs) & $32.4_{-1.2}^{+1.2}$ & $91.2_{-0.7}^{+1.2}$ & -- \\
 & Beam & $45.9_{-0.7}^{+0.7}$ & $129.2_{-3.2}^{+3.1}$ & -- \\
\cmidrule{1-5}
\multirow[c]{9}{*}{bAbI-2} & Grad ($L_2$) & $28.5_{-0.8}^{+0.8}$ & $96.3_{-2.8}^{+6.8}$ & $57.8_{-2.0}^{+2.0}$ \\
 & Grad ($L_1$) & $28.5_{-0.8}^{+0.9}$ & $96.3_{-2.7}^{+6.8}$ & -- \\
 & x $\odot$ grad (sign) & $19.7_{-8.1}^{+6.6}$ & $65.7_{-26.3}^{+24.1}$ & -- \\
 & x $\odot$ grad (abs) & $27.3_{-1.5}^{+1.7}$ & $92.0_{-3.1}^{+2.5}$ & $48.1_{-3.5}^{+3.2}$ \\
 & IG (sign) & $40.3_{-0.8}^{+0.9}$ & $136.3_{-6.4}^{+4.4}$ & -- \\
 & IG (abs) & $29.1_{-1.3}^{+1.0}$ & $98.3_{-3.9}^{+5.5}$ & $42.0_{-4.8}^{+3.8}$ \\
 & LOO (sign) & $40.2_{-0.8}^{+1.2}$ & $136.0_{-6.5}^{+4.1}$ & -- \\
 & LOO (abs) & $28.5_{-1.4}^{+0.9}$ & $96.3_{-3.6}^{+9.2}$ & -- \\
 & Beam & $41.1_{-0.7}^{+1.0}$ & $139.2_{-7.3}^{+5.0}$ & -- \\
\cmidrule{1-5}
\multirow[c]{9}{*}{bAbI-3} & Grad ($L_2$) & $23.5_{-1.5}^{+0.9}$ & $97.3_{-4.2}^{+3.5}$ & $34.0_{-15.1}^{+14.6}$ \\
 & Grad ($L_1$) & $23.5_{-1.3}^{+0.9}$ & $97.2_{-4.3}^{+3.6}$ & -- \\
 & x $\odot$ grad (sign) & $23.1_{-2.5}^{+2.4}$ & $96.7_{-14.9}^{+21.8}$ & -- \\
 & x $\odot$ grad (abs) & $24.4_{-1.2}^{+0.7}$ & $101.2_{-5.4}^{+4.3}$ & $22.4_{-12.4}^{+15.9}$ \\
 & IG (sign) & $36.6_{-1.4}^{+1.3}$ & $152.5_{-17.4}^{+14.7}$ & -- \\
 & IG (abs) & $24.4_{-1.3}^{+1.0}$ & $100.9_{-3.9}^{+4.0}$ & $-27.9_{-49.1}^{+18.0}$ \\
 & LOO (sign) & $38.7_{-0.7}^{+0.9}$ & $160.5_{-13.2}^{+13.8}$ & -- \\
 & LOO (abs) & $23.4_{-1.1}^{+0.7}$ & $97.0_{-4.6}^{+4.5}$ & -- \\
 & Beam & -- & -- & -- \\
\cmidrule{1-5}
\multirow[c]{9}{*}{BoolQ} & Grad ($L_2$) & $4.1_{-0.3}^{+0.2}$ & $50.1_{-3.2}^{+3.6}$ & -- \\
 & Grad ($L_1$) & $4.1_{-0.3}^{+0.2}$ & $50.1_{-3.1}^{+4.0}$ & -- \\
 & x $\odot$ grad (sign) & $-1.3_{-0.4}^{+0.7}$ & $-16.0_{-4.5}^{+8.2}$ & -- \\
 & x $\odot$ grad (abs) & $1.7_{-0.3}^{+0.2}$ & $21.3_{-3.3}^{+2.7}$ & -- \\
 & IG (sign) & $9.6_{-0.6}^{+0.6}$ & $118.3_{-7.9}^{+10.3}$ & -- \\
 & IG (abs) & $3.0_{-0.4}^{+0.2}$ & $37.8_{-5.3}^{+4.7}$ & -- \\
 & LOO (sign) & $26.3_{-1.4}^{+1.9}$ & $323.1_{-15.2}^{+14.4}$ & -- \\
 & LOO (abs) & $5.3_{-0.2}^{+0.3}$ & $65.2_{-2.2}^{+3.3}$ & -- \\
 & Beam & $21.2_{-1.1}^{+1.4}$ & $261.4_{-14.5}^{+10.7}$ & -- \\
\cmidrule{1-5}
\multirow[c]{9}{*}{CB} & Grad ($L_2$) & $20.3_{-1.3}^{+1.3}$ & $64.7_{-9.9}^{+9.9}$ & -- \\
 & Grad ($L_1$) & $20.2_{-1.3}^{+1.4}$ & $64.7_{-10.7}^{+10.7}$ & -- \\
 & x $\odot$ grad (sign) & $6.0_{-3.6}^{+4.0}$ & $17.8_{-11.0}^{+7.0}$ & -- \\
 & x $\odot$ grad (abs) & $15.9_{-3.2}^{+2.2}$ & $52.0_{-14.3}^{+15.7}$ & -- \\
 & IG (sign) & $30.1_{-5.7}^{+5.7}$ & $97.6_{-28.3}^{+22.7}$ & -- \\
 & IG (abs) & $21.1_{-2.3}^{+2.3}$ & $67.5_{-13.2}^{+10.8}$ & -- \\
 & LOO (sign) & $40.9_{-5.5}^{+5.8}$ & $131.4_{-27.3}^{+26.7}$ & -- \\
 & LOO (abs) & $21.0_{-2.4}^{+1.7}$ & $67.4_{-12.8}^{+12.8}$ & -- \\
 & Beam & $51.0_{-6.8}^{+6.4}$ & $163.5_{-33.8}^{+32.7}$ & -- \\
\cmidrule{1-5}
\multirow[c]{9}{*}{CoLA} & Grad ($L_2$) & $12.7_{-1.1}^{+0.9}$ & $43.4_{-3.2}^{+4.0}$ & -- \\
 & Grad ($L_1$) & $12.7_{-1.2}^{+0.8}$ & $43.2_{-3.4}^{+3.7}$ & -- \\
 & x $\odot$ grad (sign) & $-2.4_{-1.1}^{+1.9}$ & $-8.3_{-3.7}^{+6.0}$ & -- \\
 & x $\odot$ grad (abs) & $8.0_{-0.8}^{+1.0}$ & $27.1_{-2.9}^{+2.9}$ & -- \\
 & IG (sign) & $8.1_{-0.8}^{+1.1}$ & $27.5_{-2.6}^{+2.8}$ & -- \\
 & IG (abs) & $8.7_{-0.8}^{+1.7}$ & $29.8_{-3.0}^{+6.6}$ & -- \\
 & LOO (sign) & $52.0_{-2.9}^{+1.7}$ & $177.2_{-7.8}^{+7.2}$ & -- \\
 & LOO (abs) & $17.0_{-0.4}^{+0.4}$ & $57.9_{-0.9}^{+1.2}$ & -- \\
 & Beam & $50.6_{-2.3}^{+1.3}$ & $172.7_{-7.2}^{+5.9}$ & -- \\
\cmidrule{1-5}
\multirow[c]{9}{*}{Anemia} & Grad ($L_2$) & $23.8_{-0.5}^{+0.6}$ & $62.1_{-1.7}^{+1.4}$ & $18.2_{-13.8}^{+11.8}$ \\
 & Grad ($L_1$) & $23.8_{-0.6}^{+0.6}$ & $62.2_{-2.1}^{+1.4}$ & -- \\
 & x $\odot$ grad (sign) & $9.7_{-2.5}^{+2.6}$ & $25.1_{-6.2}^{+6.5}$ & -- \\
 & x $\odot$ grad (abs) & $16.6_{-1.3}^{+1.3}$ & $43.2_{-3.7}^{+3.0}$ & $8.8_{-22.8}^{+22.7}$ \\
 & IG (sign) & $62.0_{-1.5}^{+1.6}$ & $161.8_{-2.3}^{+2.7}$ & -- \\
 & IG (abs) & $20.0_{-1.6}^{+0.9}$ & $52.1_{-4.5}^{+2.5}$ & $12.5_{-7.0}^{+11.3}$ \\
 & LOO (sign) & $63.3_{-1.6}^{+1.4}$ & $165.2_{-3.2}^{+2.9}$ & -- \\
 & LOO (abs) & $18.9_{-1.3}^{+1.0}$ & $49.2_{-3.8}^{+2.5}$ & -- \\
 & Beam & -- & -- & -- \\
\cmidrule{1-5}
\multirow[c]{9}{*}{Diabetes} & Grad ($L_2$) & $19.7_{-0.7}^{+1.2}$ & $91.8_{-0.9}^{+0.6}$ & $57.9_{-19.8}^{+14.4}$ \\
 & Grad ($L_1$) & $19.6_{-0.7}^{+1.0}$ & $91.6_{-0.9}^{+0.5}$ & -- \\
 & x $\odot$ grad (sign) & $10.9_{-1.3}^{+1.9}$ & $51.1_{-7.8}^{+9.1}$ & -- \\
 & x $\odot$ grad (abs) & $18.8_{-0.7}^{+1.3}$ & $87.9_{-2.0}^{+1.4}$ & $53.4_{-29.3}^{+23.2}$ \\
 & IG (sign) & $24.8_{-2.1}^{+1.5}$ & $115.8_{-7.8}^{+2.8}$ & -- \\
 & IG (abs) & $19.4_{-0.6}^{+1.0}$ & $90.5_{-1.2}^{+0.6}$ & $26.1_{-25.1}^{+12.0}$ \\
 & LOO (sign) & $41.5_{-5.9}^{+2.3}$ & $193.4_{-17.4}^{+8.9}$ & -- \\
 & LOO (abs) & $19.1_{-0.6}^{+1.1}$ & $89.0_{-0.6}^{+0.4}$ & -- \\
 & Beam & -- & -- & -- \\
\bottomrule
\end{tabular}
}
    \end{subtable}
    \begin{subtable}[t]{0.35\textwidth}
    \resizebox{0.95\linewidth}{!}{\begin{tabular}[t]{llccc}
\toprule
& & \multicolumn{3}{c}{Faithfulness [\%]}  \\
\cmidrule(r){3-5}
Dataset & IM & \multicolumn{2}{c}{Our} & R-ROAR \\
\cmidrule(r){3-4}
& & ACU & RACU & RACU \\
\midrule
\multirow[c]{9}{*}{MRPC} & Grad ($L_2$) & $7.8_{-0.6}^{+0.5}$ & $36.8_{-6.1}^{+4.3}$ & -- \\
 & Grad ($L_1$) & $7.9_{-0.7}^{+0.5}$ & $37.4_{-6.3}^{+4.2}$ & -- \\
 & x $\odot$ grad (sign) & $-2.9_{-0.6}^{+0.5}$ & $-13.7_{-2.9}^{+2.9}$ & -- \\
 & x $\odot$ grad (abs) & $3.3_{-0.8}^{+0.6}$ & $15.6_{-4.9}^{+3.4}$ & -- \\
 & IG (sign) & $12.7_{-1.0}^{+0.8}$ & $59.9_{-10.6}^{+6.2}$ & -- \\
 & IG (abs) & $5.4_{-0.6}^{+1.0}$ & $25.4_{-3.9}^{+7.0}$ & -- \\
 & LOO (sign) & $37.1_{-1.3}^{+2.4}$ & $174.4_{-18.9}^{+12.4}$ & -- \\
 & LOO (abs) & $9.9_{-0.9}^{+0.8}$ & $47.0_{-8.9}^{+5.5}$ & -- \\
 & Beam & $45.6_{-1.5}^{+2.6}$ & $214.8_{-25.1}^{+13.3}$ & -- \\
\cmidrule{1-5}
\multirow[c]{9}{*}{RTE} & Grad ($L_2$) & $9.4_{-1.0}^{+0.8}$ & $73.9_{-19.3}^{+23.6}$ & -- \\
 & Grad ($L_1$) & $9.4_{-1.0}^{+0.8}$ & $73.5_{-19.1}^{+23.3}$ & -- \\
 & x $\odot$ grad (sign) & $0.1_{-1.7}^{+2.3}$ & $-3.5_{-13.9}^{+17.3}$ & -- \\
 & x $\odot$ grad (abs) & $5.7_{-0.5}^{+0.5}$ & $44.3_{-8.5}^{+11.8}$ & -- \\
 & IG (sign) & $20.2_{-2.2}^{+1.3}$ & $156.6_{-28.4}^{+45.5}$ & -- \\
 & IG (abs) & $7.3_{-0.9}^{+1.6}$ & $55.9_{-13.7}^{+14.8}$ & -- \\
 & LOO (sign) & $44.4_{-1.9}^{+1.6}$ & $344.8_{-75.5}^{+92.4}$ & -- \\
 & LOO (abs) & $9.4_{-0.6}^{+0.9}$ & $73.2_{-15.2}^{+21.1}$ & -- \\
 & Beam & $51.3_{-1.7}^{+2.0}$ & $399.7_{-89.9}^{+109.9}$ & -- \\
\cmidrule{1-5}
\multirow[c]{9}{*}{SST2} & Grad ($L_2$) & $12.2_{-0.7}^{+0.6}$ & $40.4_{-1.7}^{+3.0}$ & $26.1_{-2.2}^{+1.6}$ \\
 & Grad ($L_1$) & $12.1_{-0.7}^{+0.7}$ & $40.3_{-1.8}^{+3.3}$ & -- \\
 & x $\odot$ grad (sign) & $-3.7_{-1.6}^{+1.5}$ & $-12.2_{-6.0}^{+4.5}$ & -- \\
 & x $\odot$ grad (abs) & $7.1_{-0.2}^{+0.2}$ & $23.5_{-1.1}^{+1.9}$ & $18.6_{-4.6}^{+4.1}$ \\
 & IG (sign) & $31.8_{-2.2}^{+2.8}$ & $105.6_{-7.7}^{+7.7}$ & -- \\
 & IG (abs) & $13.7_{-0.8}^{+0.8}$ & $45.3_{-2.8}^{+4.1}$ & $32.9_{-1.5}^{+1.8}$ \\
 & LOO (sign) & $51.6_{-0.9}^{+1.4}$ & $171.3_{-6.2}^{+5.8}$ & -- \\
 & LOO (abs) & $16.6_{-1.0}^{+1.2}$ & $54.9_{-1.5}^{+2.1}$ & -- \\
 & Beam & $56.4_{-0.7}^{+0.5}$ & $187.3_{-7.1}^{+8.1}$ & -- \\
\cmidrule{1-5}
\multirow[c]{9}{*}{SNLI} & Grad ($L_2$) & $8.9_{-0.4}^{+1.0}$ & $62.2_{-1.7}^{+2.7}$ & $50.7_{-0.8}^{+1.1}$ \\
 & Grad ($L_1$) & $8.9_{-0.4}^{+1.0}$ & $62.2_{-1.8}^{+2.6}$ & -- \\
 & x $\odot$ grad (sign) & $1.3_{-0.8}^{+0.7}$ & $9.2_{-4.9}^{+5.2}$ & -- \\
 & x $\odot$ grad (abs) & $6.4_{-0.3}^{+0.5}$ & $44.8_{-1.6}^{+1.6}$ & $41.0_{-0.5}^{+0.4}$ \\
 & IG (sign) & $16.3_{-1.3}^{+1.3}$ & $113.5_{-6.4}^{+7.6}$ & -- \\
 & IG (abs) & $8.9_{-0.5}^{+0.8}$ & $62.3_{-2.1}^{+2.5}$ & $56.7_{-1.1}^{+1.0}$ \\
 & LOO (sign) & $26.6_{-0.4}^{+0.4}$ & $186.6_{-12.4}^{+9.1}$ & -- \\
 & LOO (abs) & $10.5_{-0.5}^{+0.8}$ & $73.6_{-1.9}^{+2.0}$ & -- \\
 & Beam & $29.3_{-0.3}^{+0.2}$ & $205.2_{-8.7}^{+8.1}$ & -- \\
\cmidrule{1-5}
\multirow[c]{9}{*}{IMDB} & Grad ($L_2$) & $24.9_{-4.3}^{+5.9}$ & $47.8_{-8.5}^{+10.9}$ & $25.4_{-2.0}^{+3.1}$ \\
 & Grad ($L_1$) & $24.9_{-4.4}^{+5.7}$ & $47.8_{-8.3}^{+10.9}$ & -- \\
 & x $\odot$ grad (sign) & $-3.2_{-1.0}^{+1.3}$ & $-6.2_{-1.9}^{+2.5}$ & -- \\
 & x $\odot$ grad (abs) & $12.9_{-2.6}^{+4.5}$ & $24.7_{-5.1}^{+8.7}$ & $16.9_{-3.0}^{+1.1}$ \\
 & IG (sign) & $40.3_{-1.9}^{+3.3}$ & $77.3_{-3.6}^{+6.3}$ & -- \\
 & IG (abs) & $18.4_{-2.8}^{+4.2}$ & $35.4_{-5.9}^{+8.0}$ & $35.1_{-1.7}^{+2.4}$ \\
 & LOO (sign) & $68.0_{-0.9}^{+0.9}$ & $130.7_{-1.5}^{+1.3}$ & -- \\
 & LOO (abs) & $20.7_{-3.2}^{+4.2}$ & $39.8_{-6.0}^{+9.5}$ & -- \\
 & Beam & -- & -- & -- \\
\cmidrule{1-5}
\multirow[c]{9}{*}{MNLI} & Grad ($L_2$) & $8.8_{-0.2}^{+0.3}$ & $49.6_{-1.2}^{+1.1}$ & -- \\
 & Grad ($L_1$) & $8.8_{-0.2}^{+0.3}$ & $49.7_{-1.2}^{+1.1}$ & -- \\
 & x $\odot$ grad (sign) & $0.6_{-0.7}^{+0.6}$ & $3.5_{-3.8}^{+3.7}$ & -- \\
 & x $\odot$ grad (abs) & $5.7_{-0.2}^{+0.2}$ & $32.0_{-0.8}^{+0.7}$ & -- \\
 & IG (sign) & $18.5_{-0.7}^{+0.6}$ & $104.3_{-5.6}^{+5.9}$ & -- \\
 & IG (abs) & $9.1_{-0.2}^{+0.2}$ & $51.3_{-2.0}^{+1.0}$ & -- \\
 & LOO (sign) & $34.7_{-0.5}^{+0.7}$ & $195.7_{-2.8}^{+3.2}$ & -- \\
 & LOO (abs) & $11.8_{-0.3}^{+0.3}$ & $66.7_{-1.1}^{+0.9}$ & -- \\
 & Beam & $40.5_{-0.4}^{+0.8}$ & $228.9_{-4.5}^{+4.5}$ & -- \\
\cmidrule{1-5}
\multirow[c]{9}{*}{QNLI} & Grad ($L_2$) & $12.7_{-0.7}^{+0.6}$ & $40.8_{-2.0}^{+2.4}$ & -- \\
 & Grad ($L_1$) & $12.7_{-0.7}^{+0.6}$ & $40.8_{-1.9}^{+2.3}$ & -- \\
 & x $\odot$ grad (sign) & $-3.7_{-1.1}^{+1.9}$ & $-11.9_{-3.4}^{+6.0}$ & -- \\
 & x $\odot$ grad (abs) & $8.8_{-0.6}^{+0.6}$ & $28.3_{-1.7}^{+2.1}$ & -- \\
 & IG (sign) & $24.4_{-2.0}^{+4.3}$ & $78.0_{-7.4}^{+10.4}$ & -- \\
 & IG (abs) & $11.4_{-0.6}^{+0.7}$ & $36.5_{-2.6}^{+2.0}$ & -- \\
 & LOO (sign) & $46.1_{-2.6}^{+1.8}$ & $147.7_{-8.7}^{+4.7}$ & -- \\
 & LOO (abs) & $14.0_{-0.4}^{+0.6}$ & $44.9_{-1.3}^{+3.2}$ & -- \\
 & Beam & $55.7_{-2.2}^{+1.5}$ & $178.4_{-8.6}^{+4.8}$ & -- \\
\cmidrule{1-5}
\multirow[c]{9}{*}{QQP} & Grad ($L_2$) & $4.5_{-0.1}^{+0.1}$ & $40.0_{-0.8}^{+0.4}$ & -- \\
 & Grad ($L_1$) & $4.5_{-0.1}^{+0.1}$ & $40.0_{-0.7}^{+0.4}$ & -- \\
 & x $\odot$ grad (sign) & $-0.8_{-0.6}^{+0.3}$ & $-7.3_{-5.5}^{+3.0}$ & -- \\
 & x $\odot$ grad (abs) & $2.8_{-0.2}^{+0.2}$ & $24.5_{-1.5}^{+1.4}$ & -- \\
 & IG (sign) & $9.1_{-0.3}^{+0.4}$ & $81.3_{-2.3}^{+2.6}$ & -- \\
 & IG (abs) & $3.5_{-0.2}^{+0.2}$ & $31.1_{-1.3}^{+2.3}$ & -- \\
 & LOO (sign) & $22.0_{-0.9}^{+0.6}$ & $195.4_{-7.7}^{+6.1}$ & -- \\
 & LOO (abs) & $5.6_{-0.2}^{+0.1}$ & $49.6_{-1.8}^{+0.8}$ & -- \\
 & Beam & $23.3_{-0.6}^{+0.5}$ & $207.3_{-5.7}^{+6.6}$ & -- \\
\bottomrule
\end{tabular}
}
    \end{subtable}
\label{tab:appendix:faithfulness:roberta-sb}
\end{table}

\begin{table}[H]
    \centering
    \caption{Faithfulness scores for \textbf{RoBERTa-large}. Shows Relative Area Between Curves (RACU) and the non-relative variant (ACU), defined by \citet{Madsen2022}. Note that \citet{Madsen2022} does not report results for Recursive-ROAR with RoBERTa-large.}
    \begin{subtable}[t]{0.35\textwidth}
    \resizebox{0.9685\linewidth}{!}{\begin{tabular}[t]{llccc}
\toprule
& & \multicolumn{3}{c}{Faithfulness [\%]}  \\
\cmidrule(r){3-5}
Dataset & IM & \multicolumn{2}{c}{Our} & R-ROAR \\
\cmidrule(r){3-4}
& & ACU & RACU & RACU \\
\midrule
\multirow[c]{9}{*}{bAbI-1} & Grad ($L_2$) & $31.1_{-1.5}^{+1.0}$ & $87.6_{-2.3}^{+1.6}$ & -- \\
 & Grad ($L_1$) & $31.1_{-1.6}^{+1.1}$ & $87.7_{-2.5}^{+1.6}$ & -- \\
 & x $\odot$ grad (sign) & $13.9_{-6.5}^{+3.1}$ & $39.1_{-16.0}^{+8.8}$ & -- \\
 & x $\odot$ grad (abs) & $24.0_{-3.2}^{+2.6}$ & $67.5_{-8.3}^{+5.7}$ & -- \\
 & IG (sign) & $36.3_{-3.1}^{+2.9}$ & $102.4_{-6.9}^{+7.7}$ & -- \\
 & IG (abs) & $31.6_{-1.5}^{+1.3}$ & $88.9_{-2.7}^{+2.5}$ & -- \\
 & LOO (sign) & $44.4_{-1.0}^{+0.9}$ & $125.2_{-1.7}^{+1.7}$ & -- \\
 & LOO (abs) & $32.4_{-0.6}^{+0.4}$ & $91.4_{-0.9}^{+0.4}$ & -- \\
 & Beam & $44.7_{-1.0}^{+0.8}$ & $126.0_{-1.1}^{+1.1}$ & -- \\
\cmidrule{1-5}
\multirow[c]{9}{*}{bAbI-2} & Grad ($L_2$) & $28.2_{-1.5}^{+0.9}$ & $94.0_{-3.4}^{+6.3}$ & -- \\
 & Grad ($L_1$) & $28.0_{-1.4}^{+0.9}$ & $93.5_{-2.8}^{+5.3}$ & -- \\
 & x $\odot$ grad (sign) & $8.2_{-4.9}^{+7.0}$ & $26.9_{-19.4}^{+20.4}$ & -- \\
 & x $\odot$ grad (abs) & $22.6_{-1.5}^{+1.5}$ & $75.5_{-5.2}^{+7.7}$ & -- \\
 & IG (sign) & $37.9_{-1.3}^{+1.3}$ & $126.6_{-9.7}^{+5.8}$ & -- \\
 & IG (abs) & $27.4_{-1.7}^{+1.7}$ & $91.7_{-9.5}^{+6.8}$ & -- \\
 & LOO (sign) & $40.6_{-0.7}^{+1.9}$ & $135.4_{-4.0}^{+4.0}$ & -- \\
 & LOO (abs) & $28.1_{-0.9}^{+0.9}$ & $93.9_{-1.8}^{+2.8}$ & -- \\
 & Beam & $41.7_{-0.7}^{+1.8}$ & $139.2_{-4.3}^{+4.3}$ & -- \\
\cmidrule{1-5}
\multirow[c]{9}{*}{bAbI-3} & Grad ($L_2$) & $22.4_{-3.8}^{+3.8}$ & $94.7_{-0.2}^{+0.2}$ & -- \\
 & Grad ($L_1$) & $22.2_{-3.6}^{+3.6}$ & $94.1_{-0.4}^{+0.4}$ & -- \\
 & x $\odot$ grad (sign) & $8.5_{-3.5}^{+3.5}$ & $34.4_{-9.0}^{+9.0}$ & -- \\
 & x $\odot$ grad (abs) & $19.9_{-4.6}^{+4.6}$ & $83.2_{-5.5}^{+5.5}$ & -- \\
 & IG (sign) & $33.0_{-3.8}^{+3.8}$ & $141.3_{-7.7}^{+7.7}$ & -- \\
 & IG (abs) & $24.3_{-3.7}^{+3.7}$ & $103.2_{-1.5}^{+1.5}$ & -- \\
 & LOO (sign) & $35.0_{-3.0}^{+3.0}$ & $150.5_{-12.4}^{+12.4}$ & -- \\
 & LOO (abs) & $23.3_{-4.2}^{+4.2}$ & $98.7_{-1.2}^{+1.2}$ & -- \\
 & Beam & -- & -- & -- \\
\cmidrule{1-5}
\multirow[c]{9}{*}{BoolQ} & Grad ($L_2$) & $2.6_{-0.3}^{+0.1}$ & $24.8_{-1.9}^{+1.7}$ & -- \\
 & Grad ($L_1$) & $2.7_{-0.3}^{+0.2}$ & $25.3_{-1.9}^{+2.3}$ & -- \\
 & x $\odot$ grad (sign) & $-0.4_{-0.1}^{+0.2}$ & $-3.6_{-1.5}^{+1.6}$ & -- \\
 & x $\odot$ grad (abs) & $1.2_{-0.3}^{+0.3}$ & $10.8_{-2.0}^{+2.0}$ & -- \\
 & IG (sign) & $6.9_{-0.8}^{+1.2}$ & $65.6_{-6.9}^{+15.7}$ & -- \\
 & IG (abs) & $3.2_{-0.8}^{+0.4}$ & $30.3_{-5.0}^{+4.6}$ & -- \\
 & LOO (sign) & $25.6_{-2.7}^{+1.7}$ & $242.9_{-26.4}^{+39.2}$ & -- \\
 & LOO (abs) & $6.2_{-0.5}^{+0.6}$ & $58.0_{-1.2}^{+3.3}$ & -- \\
 & Beam & $21.5_{-2.4}^{+1.3}$ & $204.0_{-21.7}^{+30.8}$ & -- \\
\cmidrule{1-5}
\multirow[c]{9}{*}{CB} & Grad ($L_2$) & $10.1_{-4.2}^{+4.4}$ & $30.8_{-12.9}^{+15.2}$ & -- \\
 & Grad ($L_1$) & $8.9_{-4.3}^{+3.9}$ & $27.3_{-13.2}^{+12.6}$ & -- \\
 & x $\odot$ grad (sign) & $-0.1_{-3.3}^{+2.7}$ & $0.6_{-8.1}^{+9.5}$ & -- \\
 & x $\odot$ grad (abs) & $5.6_{-2.9}^{+1.8}$ & $17.4_{-9.8}^{+6.1}$ & -- \\
 & IG (sign) & $28.5_{-3.4}^{+3.9}$ & $85.3_{-14.0}^{+19.0}$ & -- \\
 & IG (abs) & $17.1_{-2.2}^{+3.2}$ & $51.3_{-9.0}^{+10.5}$ & -- \\
 & LOO (sign) & $39.0_{-2.4}^{+2.9}$ & $116.1_{-15.8}^{+9.5}$ & -- \\
 & LOO (abs) & $19.0_{-2.4}^{+4.0}$ & $56.6_{-12.8}^{+8.8}$ & -- \\
 & Beam & $51.8_{-3.0}^{+3.0}$ & $154.6_{-22.6}^{+14.0}$ & -- \\
\cmidrule{1-5}
\multirow[c]{9}{*}{CoLA} & Grad ($L_2$) & $10.9_{-0.7}^{+0.9}$ & $35.0_{-2.5}^{+2.9}$ & -- \\
 & Grad ($L_1$) & $10.4_{-0.6}^{+0.9}$ & $33.3_{-2.2}^{+3.1}$ & -- \\
 & x $\odot$ grad (sign) & $1.2_{-0.3}^{+0.2}$ & $3.8_{-0.9}^{+0.5}$ & -- \\
 & x $\odot$ grad (abs) & $6.6_{-0.9}^{+0.9}$ & $21.3_{-2.9}^{+2.9}$ & -- \\
 & IG (sign) & $17.3_{-1.1}^{+1.1}$ & $55.4_{-3.2}^{+3.6}$ & -- \\
 & IG (abs) & $11.7_{-0.3}^{+0.4}$ & $37.5_{-0.9}^{+1.0}$ & -- \\
 & LOO (sign) & $38.9_{-2.0}^{+4.1}$ & $124.9_{-6.0}^{+11.4}$ & -- \\
 & LOO (abs) & $17.6_{-1.3}^{+0.9}$ & $56.7_{-3.4}^{+3.2}$ & -- \\
 & Beam & $37.5_{-2.7}^{+4.7}$ & $120.3_{-8.2}^{+13.4}$ & -- \\
\cmidrule{1-5}
\multirow[c]{9}{*}{Anemia} & Grad ($L_2$) & $18.6_{-1.5}^{+1.7}$ & $47.9_{-3.6}^{+3.7}$ & -- \\
 & Grad ($L_1$) & $18.4_{-1.8}^{+2.0}$ & $47.4_{-4.3}^{+4.3}$ & -- \\
 & x $\odot$ grad (sign) & $4.6_{-1.6}^{+1.8}$ & $11.9_{-4.2}^{+4.6}$ & -- \\
 & x $\odot$ grad (abs) & $11.6_{-2.1}^{+2.1}$ & $29.8_{-5.5}^{+5.5}$ & -- \\
 & IG (sign) & $58.2_{-4.0}^{+2.9}$ & $150.1_{-5.0}^{+4.2}$ & -- \\
 & IG (abs) & $16.4_{-1.9}^{+1.2}$ & $42.3_{-5.7}^{+2.4}$ & -- \\
 & LOO (sign) & $60.5_{-2.8}^{+3.3}$ & $156.0_{-5.7}^{+7.4}$ & -- \\
 & LOO (abs) & $15.6_{-1.7}^{+1.6}$ & $40.2_{-3.7}^{+3.5}$ & -- \\
 & Beam & -- & -- & -- \\
\cmidrule{1-5}
\multirow[c]{9}{*}{Diabetes} & Grad ($L_2$) & $20.1_{-1.1}^{+1.8}$ & $90.7_{-0.4}^{+0.5}$ & -- \\
 & Grad ($L_1$) & $20.1_{-1.1}^{+1.9}$ & $90.7_{-0.4}^{+0.6}$ & -- \\
 & x $\odot$ grad (sign) & $7.4_{-4.0}^{+3.1}$ & $34.2_{-19.0}^{+16.6}$ & -- \\
 & x $\odot$ grad (abs) & $18.6_{-0.9}^{+1.5}$ & $84.2_{-0.8}^{+0.6}$ & -- \\
 & IG (sign) & $25.3_{-2.3}^{+4.0}$ & $113.5_{-5.1}^{+7.9}$ & -- \\
 & IG (abs) & $19.6_{-1.0}^{+1.4}$ & $88.7_{-1.2}^{+0.5}$ & -- \\
 & LOO (sign) & $34.8_{-4.9}^{+6.7}$ & $156.0_{-18.3}^{+12.2}$ & -- \\
 & LOO (abs) & $19.2_{-1.0}^{+1.4}$ & $86.6_{-0.9}^{+0.6}$ & -- \\
 & Beam & -- & -- & -- \\
\bottomrule
\end{tabular}
}
    \end{subtable}
    \begin{subtable}[t]{0.35\textwidth}
    \resizebox{0.95\linewidth}{!}{\begin{tabular}[t]{llccc}
\toprule
& & \multicolumn{3}{c}{Faithfulness [\%]}  \\
\cmidrule(r){3-5}
Dataset & IM & \multicolumn{2}{c}{Our} & R-ROAR \\
\cmidrule(r){3-4}
& & ACU & RACU & RACU \\
\midrule
\multirow[c]{9}{*}{MRPC} & Grad ($L_2$) & $6.6_{-1.1}^{+2.2}$ & $22.9_{-4.3}^{+3.1}$ & -- \\
 & Grad ($L_1$) & $6.6_{-1.1}^{+1.9}$ & $22.7_{-3.7}^{+1.5}$ & -- \\
 & x $\odot$ grad (sign) & $-0.9_{-1.1}^{+1.0}$ & $-3.4_{-2.2}^{+3.2}$ & -- \\
 & x $\odot$ grad (abs) & $4.4_{-1.0}^{+1.2}$ & $15.7_{-4.8}^{+2.6}$ & -- \\
 & IG (sign) & $15.7_{-2.0}^{+1.6}$ & $55.9_{-8.8}^{+9.5}$ & -- \\
 & IG (abs) & $8.1_{-1.1}^{+1.3}$ & $28.8_{-3.7}^{+9.2}$ & -- \\
 & LOO (sign) & $29.9_{-1.3}^{+0.8}$ & $110.0_{-24.2}^{+30.9}$ & -- \\
 & LOO (abs) & $10.2_{-1.6}^{+1.6}$ & $36.1_{-5.2}^{+8.8}$ & -- \\
 & Beam & $40.1_{-2.3}^{+1.4}$ & $146.9_{-30.6}^{+41.5}$ & -- \\
\cmidrule{1-5}
\multirow[c]{9}{*}{RTE} & Grad ($L_2$) & $7.1_{-1.4}^{+0.8}$ & $38.8_{-5.9}^{+5.9}$ & -- \\
 & Grad ($L_1$) & $7.4_{-1.2}^{+0.9}$ & $40.4_{-7.6}^{+5.1}$ & -- \\
 & x $\odot$ grad (sign) & $-0.1_{-1.3}^{+1.0}$ & $-1.7_{-7.2}^{+4.8}$ & -- \\
 & x $\odot$ grad (abs) & $5.0_{-1.0}^{+1.3}$ & $27.1_{-3.0}^{+6.8}$ & -- \\
 & IG (sign) & $22.6_{-2.0}^{+2.3}$ & $127.1_{-31.6}^{+28.9}$ & -- \\
 & IG (abs) & $7.7_{-1.3}^{+1.2}$ & $43.0_{-10.7}^{+8.8}$ & -- \\
 & LOO (sign) & $38.5_{-3.6}^{+1.9}$ & $213.6_{-40.5}^{+37.1}$ & -- \\
 & LOO (abs) & $9.9_{-0.4}^{+0.4}$ & $55.4_{-9.9}^{+11.9}$ & -- \\
 & Beam & $50.3_{-2.9}^{+1.2}$ & $280.1_{-45.7}^{+57.9}$ & -- \\
\cmidrule{1-5}
\multirow[c]{9}{*}{SST2} & Grad ($L_2$) & $9.8_{-1.0}^{+1.1}$ & $32.4_{-3.0}^{+3.5}$ & -- \\
 & Grad ($L_1$) & $9.7_{-0.9}^{+1.1}$ & $32.0_{-2.9}^{+3.2}$ & -- \\
 & x $\odot$ grad (sign) & $-3.4_{-0.6}^{+0.7}$ & $-11.4_{-2.2}^{+2.4}$ & -- \\
 & x $\odot$ grad (abs) & $5.4_{-1.1}^{+1.6}$ & $18.0_{-3.5}^{+5.2}$ & -- \\
 & IG (sign) & $40.1_{-1.9}^{+3.2}$ & $133.1_{-6.5}^{+10.6}$ & -- \\
 & IG (abs) & $15.6_{-0.7}^{+0.9}$ & $51.6_{-1.5}^{+3.4}$ & -- \\
 & LOO (sign) & $49.4_{-1.4}^{+1.3}$ & $164.0_{-2.9}^{+1.7}$ & -- \\
 & LOO (abs) & $17.2_{-0.7}^{+1.0}$ & $57.1_{-2.2}^{+3.6}$ & -- \\
 & Beam & $55.6_{-0.5}^{+1.0}$ & $184.5_{-2.6}^{+2.1}$ & -- \\
\cmidrule{1-5}
\multirow[c]{9}{*}{SNLI} & Grad ($L_2$) & $8.2_{-0.5}^{+0.3}$ & $53.8_{-2.0}^{+1.3}$ & -- \\
 & Grad ($L_1$) & $8.2_{-0.5}^{+0.4}$ & $53.3_{-2.1}^{+1.6}$ & -- \\
 & x $\odot$ grad (sign) & $-0.3_{-0.3}^{+0.3}$ & $-2.2_{-2.2}^{+1.7}$ & -- \\
 & x $\odot$ grad (abs) & $5.6_{-0.4}^{+0.3}$ & $36.5_{-2.0}^{+1.7}$ & -- \\
 & IG (sign) & $14.0_{-0.3}^{+0.3}$ & $91.2_{-1.0}^{+1.0}$ & -- \\
 & IG (abs) & $8.0_{-0.5}^{+0.4}$ & $52.6_{-1.7}^{+1.4}$ & -- \\
 & LOO (sign) & $26.3_{-0.5}^{+0.4}$ & $172.3_{-7.3}^{+4.3}$ & -- \\
 & LOO (abs) & $11.0_{-0.3}^{+0.5}$ & $71.8_{-0.9}^{+1.7}$ & -- \\
 & Beam & $29.6_{-0.3}^{+0.3}$ & $193.7_{-4.5}^{+5.2}$ & -- \\
\cmidrule{1-5}
\multirow[c]{9}{*}{IMDB} & Grad ($L_2$) & $13.9_{-1.9}^{+3.5}$ & $29.4_{-2.4}^{+6.0}$ & -- \\
 & Grad ($L_1$) & $13.7_{-1.9}^{+4.1}$ & $28.9_{-2.5}^{+6.0}$ & -- \\
 & x $\odot$ grad (sign) & $-2.9_{-0.4}^{+0.4}$ & $-6.4_{-1.2}^{+1.3}$ & -- \\
 & x $\odot$ grad (abs) & $7.7_{-1.2}^{+2.9}$ & $16.3_{-2.8}^{+3.9}$ & -- \\
 & IG (sign) & $53.2_{-4.1}^{+3.4}$ & $114.2_{-11.5}^{+12.8}$ & -- \\
 & IG (abs) & $18.9_{-4.2}^{+3.7}$ & $40.3_{-9.8}^{+6.2}$ & -- \\
 & LOO (sign) & $60.5_{-1.0}^{+1.1}$ & $130.1_{-13.0}^{+13.0}$ & -- \\
 & LOO (abs) & $16.7_{-1.9}^{+5.0}$ & $35.5_{-3.8}^{+5.6}$ & -- \\
 & Beam & -- & -- & -- \\
\cmidrule{1-5}
\multirow[c]{9}{*}{MNLI} & Grad ($L_2$) & $7.9_{-0.2}^{+0.1}$ & $38.7_{-1.3}^{+0.9}$ & -- \\
 & Grad ($L_1$) & $7.8_{-0.2}^{+0.2}$ & $38.3_{-1.5}^{+1.0}$ & -- \\
 & x $\odot$ grad (sign) & $-0.5_{-0.2}^{+0.5}$ & $-2.3_{-1.1}^{+2.6}$ & -- \\
 & x $\odot$ grad (abs) & $5.2_{-0.1}^{+0.1}$ & $25.4_{-1.1}^{+0.8}$ & -- \\
 & IG (sign) & $18.6_{-1.1}^{+0.8}$ & $91.1_{-4.6}^{+4.3}$ & -- \\
 & IG (abs) & $9.0_{-0.2}^{+0.3}$ & $44.2_{-0.7}^{+1.5}$ & -- \\
 & LOO (sign) & $33.0_{-1.0}^{+0.9}$ & $161.9_{-8.4}^{+6.1}$ & -- \\
 & LOO (abs) & $12.4_{-0.2}^{+0.1}$ & $60.6_{-0.9}^{+1.4}$ & -- \\
 & Beam & $41.2_{-1.2}^{+0.9}$ & $201.7_{-9.1}^{+6.5}$ & -- \\
\cmidrule{1-5}
\multirow[c]{9}{*}{QNLI} & Grad ($L_2$) & $9.5_{-0.4}^{+0.5}$ & $28.3_{-0.9}^{+1.4}$ & -- \\
 & Grad ($L_1$) & $9.4_{-0.3}^{+0.3}$ & $28.0_{-0.5}^{+1.3}$ & -- \\
 & x $\odot$ grad (sign) & $-1.4_{-0.5}^{+0.4}$ & $-4.1_{-1.3}^{+1.4}$ & -- \\
 & x $\odot$ grad (abs) & $6.5_{-0.2}^{+0.3}$ & $19.1_{-0.6}^{+0.7}$ & -- \\
 & IG (sign) & $25.0_{-5.1}^{+3.3}$ & $74.0_{-13.7}^{+9.6}$ & -- \\
 & IG (abs) & $9.7_{-1.8}^{+1.2}$ & $28.7_{-5.0}^{+3.2}$ & -- \\
 & LOO (sign) & $40.4_{-2.1}^{+1.5}$ & $119.8_{-5.0}^{+4.0}$ & -- \\
 & LOO (abs) & $12.5_{-0.3}^{+0.2}$ & $37.0_{-0.4}^{+0.3}$ & -- \\
 & Beam & $53.8_{-2.1}^{+1.7}$ & $159.5_{-4.4}^{+4.6}$ & -- \\
\cmidrule{1-5}
\multirow[c]{9}{*}{QQP} & Grad ($L_2$) & $4.0_{-0.2}^{+0.3}$ & $33.5_{-1.9}^{+1.4}$ & -- \\
 & Grad ($L_1$) & $4.0_{-0.2}^{+0.3}$ & $33.0_{-1.9}^{+1.3}$ & -- \\
 & x $\odot$ grad (sign) & $-0.4_{-0.2}^{+0.3}$ & $-3.3_{-2.0}^{+2.3}$ & -- \\
 & x $\odot$ grad (abs) & $2.5_{-0.2}^{+0.3}$ & $20.7_{-2.2}^{+1.3}$ & -- \\
 & IG (sign) & $8.9_{-0.6}^{+1.0}$ & $73.7_{-3.6}^{+4.8}$ & -- \\
 & IG (abs) & $3.8_{-0.3}^{+0.7}$ & $31.6_{-1.6}^{+2.5}$ & -- \\
 & LOO (sign) & $20.4_{-0.4}^{+0.7}$ & $169.8_{-11.1}^{+8.6}$ & -- \\
 & LOO (abs) & $5.7_{-0.3}^{+0.2}$ & $47.3_{-2.8}^{+2.3}$ & -- \\
 & Beam & $22.5_{-0.8}^{+0.7}$ & $187.0_{-10.3}^{+10.4}$ & -- \\
\bottomrule
\end{tabular}
}
    \end{subtable}
\label{tab:appendix:faithfulness:roberta-sl}
\end{table}

\end{document}